%% file: main.tex
\documentclass[10pt]{article} % For LaTeX2e

\usepackage[accepted]{rlj}

\makeatletter
\@ifundefined{mathdefault}{}{}
\@ifundefined{mathit}{\def\mathit#1{\mathnormal{#1}}}{}
\@ifundefined{textminus}{}{}
\makeatother
\usepackage{lmodern}
\usepackage{algorithm}
\usepackage{algpseudocode}
\usepackage{setspace}
\usepackage[utf8]{inputenc} % allow utf-8 input
\usepackage[T1]{fontenc}    % use 8-bit T1 fonts
\usepackage{hyperref}       % hyperlinks
\usepackage{url}            % simple URL typesetting
\usepackage{booktabs}       % professional-quality tables
\usepackage{nicefrac}       % compact symbols for 1/2, etc.
\usepackage{microtype}      % microtypography
\usepackage{xcolor}         % colors
\usepackage{tcolorbox}
\usepackage{graphicx}
\usepackage{subcaption}
\usepackage{caption}
\usepackage{tabularx,booktabs}
\usepackage{comment}
\usepackage{amsthm} % Package for theorem environments
\newtheorem{assumption}{Assumption}

\usepackage{wrapfig}
\newtheorem{theorem}{Theorem}
\newtheorem{lemma}{Lemma}

\newtheorem{corollary}{Corollary}
\usepackage[nolist]{acronym}
\usepackage{float}
\usepackage{color}
\usepackage{multicol}
\usepackage{multirow} 
\usepackage{amsmath,amssymb}
\usepackage{amsfonts}
\usepackage{textcomp}
\usepackage{psfrag}
\usepackage{multimedia}
\usepackage{fancybox}
\usepackage{mathtools}
\usepackage{pgfplots}    % if your pgf file uses pgfplots
\usepackage{adjustbox}
\usepackage{tikz}
\usepackage{amsmath}

\allowdisplaybreaks
\usepackage{standalone} % Required to include external TikZ figures
\usetikzlibrary{positioning, arrows, shapes.geometric}
\usetikzlibrary{arrows.meta, positioning, shapes}

\title{Bi-Level Reinforcement Learning Pathway for Sim-to-Real Optimality}
\setrunningtitle{Bi-level RL for Sim-to-Real Optimality}
\author{Akhil S Anand\textsuperscript{1}, Shambhuraj Sawant
\textsuperscript{1}, Paavo Parmas\textsuperscript{2}, Jasper Hoffmann\textsuperscript{3}, Dirk Peter Reinhardt\textsuperscript{1}, Sebastien Gros\textsuperscript{1}}

\emails{akhil.s.anand@ntnu.no}

\affiliations{
$^{1}$\textbf{Norwegian University of Science and Technology (NTNU), Norway}\\
$^{2}$\textbf{University of Tokyo, Japan}\\
$^{3}$\textbf{University of Freiburg, Germany}

\par % If including additional comments like below, use \par to add some whitespace. 
}

\contribution{
    We derive the sensitivity of locally converged policies trained in simulation with SPG methods in an actor–critic setting with respect to simulation dynamics and reward parameters by implicitly differentiating the SPG fixed-point condition.
}
{
    Prior work on differentiating RL policies with respect to the simulation model assumes specific policy structures (e.g., softmax over Q-functions). In contrast, our analysis directly differentiates the fixed-point condition of general SPG methods and applies to parametric policies trained with actor–critic methods without imposing such structural assumptions.
}

\contribution{
     Based on this sensitivity analysis, we formulate a bi-level RL approach that adapts simulation parameters based on the performance of the policy in the real world, thereby addressing the objective mismatch in a principled way. We illustrate this approach with a bi-level PPO algorithm.
} 
{   
Current approaches address the objective mismatch problem by embedding proxy objectives into simulation model training to account for policy performance. Complementing these approaches, the proposed bi-level RL approach couples the simulation parameter learning objective and the policy performance objective by adapting simulation parameters using gradients of real-world policy performance.}

\contribution{
We provide a thorough convergence analysis of the proposed bi-level RL approach.}
 {The analysis shows that simulation and policy parameters can be updated simultaneously, provided the policy evolves on a faster timescale than the simulator parameters, thereby eliminating the need for full policy convergence at each simulator update step.
    }

\keywords{Sensitivity Analysis, Stochastic Policy Gradient, Reinforcement Learning, Bilevel Optimization, Objective Mismatch, Sim-to-Real RL}

\summary{
Training reinforcement learning (RL) policies in simulation before deploying them in real-world environments is a common strategy when real-world interaction is costly. However, the policies trained in simulation often perform poorly in the real world due to a mismatch between the simulation model and the real-world environment. This poor performance is caused by an objective mismatch: simulation models are typically designed for predictive accuracy, while policies are trained to maximize task performance. Motivated by the need to overcome this objective mismatch, we analyze the sensitivity of RL policies trained in simulation using stochastic policy gradient (SPG) methods in an actor-critic setting with respect to simulation parameters. Based on this sensitivity analysis, we formulate a bi-level RL approach that adapts simulation parameters based on the performance of the policy in the real world, thereby addressing the objective mismatch in a principled way. We provide a convergence analysis of the proposed bi-level RL approach and illustrate the concept with a proof-of-concept bi-level Proximal Policy Optimization (PPO) algorithm.
}

\makeatletter
\g@addto@macro\normalsize{%
  \setlength\abovedisplayskip{4pt plus 2pt minus 2pt}%
  \setlength\belowdisplayskip{4pt plus 2pt minus 2pt}%
  \setlength\abovedisplayshortskip{0pt plus 1pt}%
  \setlength\belowdisplayshortskip{2pt plus 1pt minus 1pt}%
}
\makeatother

\begin{document}
\input{acronyms.tex}

\makeCover  % Create the cover page
\maketitle  % Make the title section

\begin{abstract}
Training \ac{rl} policies using simulation models before deployment in real-world environments is a common strategy when real-world interaction is expensive. This approach is used in sim-to-real RL and in dyna-style model-based RL. A key limitation of this approach is that the policies trained in simulation often perform poorly in the real world due to discrepancies between the simulation model and the real-world environment, referred to as the sim-to-real gap. This gap reflects the objective mismatch: simulation models are typically constructed for predictive accuracy, whereas policies are trained to maximize task performance. Since the policy learned in simulation is implicitly defined by the simulation parameters, understanding the sensitivity of the learned policy to these parameters enables gradient-based adaptation of the simulation model to improve real-world policy performance. Motivated by this, we derive the sensitivity of locally converged policies trained with \ac{spg} methods in an actor–critic setting, which is the most widely used approach in \ac{rl}. Based on this sensitivity analysis, we formulate a bi-level \ac{rl} approach that can address the objective mismatch problem by learning simulation parameters using gradients of real-world policy performance, thereby directly coupling simulation model adaptation with policy performance. We provide a thorough convergence analysis of the proposed bi-level \ac{rl} approach and illustrate the concept through a proof-of-concept bi-level \ac{ppo} algorithm.

\end{abstract}

\acresetall

\section{Introduction}

%lift the sim2real to simulationmodelbased and specify the differenty aporaches and highliht they all do the same

%% Predictive modelling, model mismatch and accuracy vs performance
Training \ac{rl} policies in simulation models of real-world dynamics and reward functions before deployment in real-world environments is a common strategy when real-world interaction is expensive~\citep{zhao2020sim}. This approach appears in multiple flavors, including \ac{mbrl}, where simulation models are learned online~\citep{moerland2023model}, and sim-to-real \ac{rl}, where high-fidelity simulators are constructed offline with limited online adaptation~\citep{da2025survey}.  This is supported by advances in the capabilities and fidelity of simulators and world models~\citep{makoviychuk2021isaac, collins2021review}. Despite these advances, the inevitable discrepancies between the simulation model and the real world lead to a significant performance gap upon deploying the policy in the real world, referred to as the sim-to-real gap~\citep{da2025survey}. 

Recognizing that perfectly modeling the stochastic real world is infeasible even under good data coverage, a fundamental cause of the sim-to-real gap is that simulation models are typically constructed to predict real-world dynamics and rewards accurately, whereas policy training aims to maximize task performance. These objectives are not necessarily aligned, as the conditions that maximize the prediction accuracy of a simulation model differ from those that maximize the real-world performance of policies trained in that model, except in special cases~\citep{anand2025predicting, anand2024data}. This is well recognized as the \emph{objective mismatch} problem in \ac{mbrl}~\citep{wei_unified_2024}. This issue can persist even under good data coverage and low epistemic uncertainty, because simulation models trained for prediction accuracy are not informed by which aspects of the true dynamics matter for decision-making. It is particularly important in stochastic settings, especially for problems with economic or non-smooth objectives, where decision performance can depend on features of the real-world distribution that prediction-based model training need not preserve~\citep{anand2025predicting, anand2024data}. Despite theoretical and empirical evidence supporting this observation, there remains no consensus on the appropriate objective for training simulation models~\citep{wei_unified_2024}. 

In theory, for any stochastic real-world environment, it is possible to adjust the simulation model such that the optimal policy in simulation can achieve optimal performance in the real world~\citep{anand2024data, anand2024optimality, anand2025predicting}. We refer to this property as \emph{sim-to-real optimality}. Motivated by this observation, we analyze how standard \ac{rl} policies trained in simulation (in-sim policies) can achieve sim-to-real optimality. Specifically, we consider \ac{spg} methods in an actor–critic setting, which are the most widely used methods in \ac{rl}. Since the in-sim policy is implicitly defined by the simulation parameters, understanding the sensitivity of this policy with respect to these parameters enables gradient-based adaptation of the simulation model to improve real-world policy performance. This perspective is related to prior work~\citep{anand2025predicting, nikishin2022control, chen2022adaptive} on differentiating the \ac{mdp} underlying in-sim policy training. However, existing approaches are restrictive as they rely on differentiating Bellman optimality equations and therefore assume policies induced by \ac{dp}, typically represented as softmax distributions over value functions~\citep{nikishin2022control, chen2022adaptive}.

To this end, we analyze the sensitivity of locally converged policies trained with \ac{spg} methods in an actor–critic setting using the \ac{ift}, making the analysis applicable to generic \ac{rl} policies without requiring assumptions on their structure. Based on this sensitivity analysis, we frame the problem of finding simulation parameters that enable sim-to-real optimality as a policy-gradient \ac{rl} problem with real-world feedback. This forms a bi-level \ac{rl} approach, where the inner level trains an \ac{rl} policy in simulation, and the outer level adapts simulation parameters to maximize the real-world performance of the in-sim policy, thereby aligning the objectives of simulation adaptation and policy performance. We provide a convergence analysis of the proposed bi-level \ac{rl} approach using a three-timescale separation and illustrate the concept with a bi-level \ac{ppo} algorithm~\citep{schulman2017proximal}, validating it through simple simulated examples.

Although motivated by addressing objective mismatch and sim-to-real gap, \emph{the central contribution of this work is the sensitivity analysis of \ac{spg}-based policies}, which may be useful more broadly wherever the sensitivity properties of \ac{spg} methods are of interest.

The rest of the paper is organized as follows. Section~\ref{sec:related_works} discusses related work. Section~\ref{sec:preliminaries} introduces the preliminaries. Section~\ref{sec:sim_to_real_optimality} establishes the notion of sim-to-real optimality. Section~\ref{sec:bi_level_rl} presents the central results. Section~\ref{sec:examples} presents illustrative examples. Sections~\ref{sec:discussions} and~\ref{sec:conclusions} present the discussion and conclusions.

\begin{figure*}[!]
	\centering
	\def\svgwidth{1\linewidth}
	{\fontsize{8}{8}
		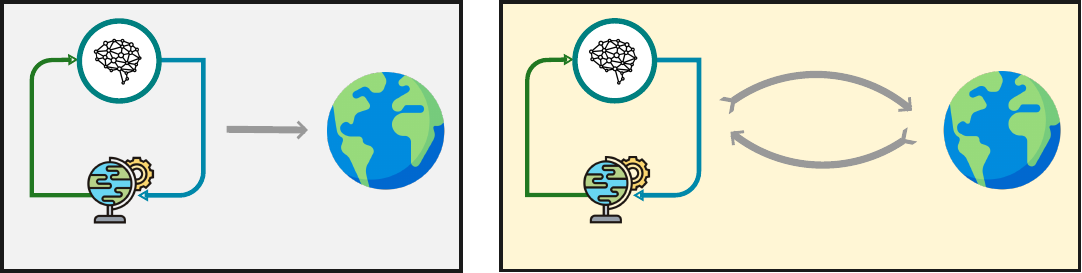}
	\caption{(Left) standard sim-to-real \ac{rl}, (right) the proposed bi-level \ac{rl} framework.}
	\label{fig:framework}
\end{figure*}

\section{Related Work}\label{sec:related_works}
Closely related works study the sensitivity of optimal policies with respect to the underlying \ac{mdp} parameters~\citep{nikishin2022control, chen2022adaptive}. These approaches differentiate Bellman optimality equations and therefore assume policies derived from value functions via \ac{dp}, typically represented as softmax (Boltzmann) distributions over Q-values~\citep{nikishin2022control, chen2022adaptive}. For example, \citet{nikishin2022control} applies implicit differentiation to the Bellman fixed-point equations of the optimal Q-function, while \citet{chen2022adaptive} derives sensitivities of entropy-regularized optimal policies using the softmax structure induced by KL regularization. Consequently, these formulations assume policies are expressed as softmax transformations of value functions and do not directly cover policies with arbitrary parameterizations, such as \ac{nn} policies optimized via direct policy search. In contrast, our work derives sensitivities of policies obtained through \ac{spg} learning by implicitly differentiating the stationary condition of the policy gradient objective. This allows the analysis to apply directly to general parametric \ac{nn} policies.

\emph{Objective mismatch} is a well-recognized problem in \ac{mbrl}~\citep{lambert2020objective,eysenbach2022mismatched, janner2019trust}, underscored by theoretical and empirical evidence~\citep{pmlr-v162-hansen22a, farahmand2017value, schrittwieser2020mastering,difftori, voelcker_value_2023a}.  Decision-aware approaches address this by incorporating value- or policy-level objectives into model learning~\citep{wei_unified_2024}.  \emph{Value-based} methods aim to learn simulation models that capture value functions instead of transition dynamics, following the \emph{Proper Value Equivalence Principle}~\citep{grimm_value_2020, grimm2021proper}. \emph{Policy-based} methods instead use policy-gradient updates to adapt the simulation model, for example, by matching real-world and simulated policy gradients or minimizing gradient discrepancies~\citep{jia_model_2024,doro_gradientaware_2020,abachi_policyaware_2021}. Similarly, most sim-to-real \ac{rl} methods are robustness-oriented rather than performance-oriented, optimizing proxy objectives such as simulator accuracy or variability instead of real-world performance directly~\citep{da2025survey,tobin2017domain,duan2024learning}; see Appendix~\ref{appendix:related_works}. We provide a complementary perspective by explicitly analyzing the sensitivity of the policy to the simulation model. Rather than heuristically modifying the model-learning objective, the sensitivity analysis is used to directly optimize model parameters to improve policy performance. Our approach falls within \emph{Differentiable Planning}, where policy optimization is embedded in a differentiable program to jointly optimize the policy and simulator toward a common objective.

\section{Preliminaries}\label{sec:preliminaries}
\subsection*{Reinforcement Learning and Optimal Decision-Making}\label{sec:background_mdp_rl}
Formally, an \ac{rl} problem is defined as a \ac{mdp}, \((\mathcal{S}, \mathcal{A}, \mathcal{P}, r, \gamma)\),
where \( \mathcal{S} \) is the state space, \( \mathcal{A} \) is the action space,  \( \mathcal{P}(s' | s, a) \) is the real-world transition model defining the probability of transitioning to state \( s' \) given \( s, a \)~\citep{sutton2018reinforcement}. \( r: \mathcal{S} \times \mathcal{A} \to \mathbb{R} \) is the scalar reward function provided by the environment, and \( \gamma \in [0, 1] \) is the discount factor. A policy \( \pi(a|s) \) defines the agent's behavior, mapping states to a distribution over actions. The objective of \ac{rl} is to find the optimal policy \( \pi^{\star} \) that maximizes the expected return {\({J(\pi) = \mathbb{E}_{\tau \sim \pi} \left[ \sum_{t=0}^{\infty} \gamma^t r(s_t, a_t) \right]}\)}. The state-value function and action-value function are defined as \( {V^\pi(s) = \mathbb{E}_{\pi} \left[ \sum_{t=0}^{\infty} \gamma^t r(s_t, a_t) \mid s_0 = s \right]} \) and \({Q^\pi(s, a) = \mathbb{E}_{\pi} \left[ \sum_{t=0}^{\infty} \gamma^t r(s_t, a_t) \mid s_0 = s, a_0 = a \right]}\), respectively. An advantage function is defined as \(A^\pi(s, a)=Q^\pi(s, a) - V^\pi(s)\). The optimal value- \( V^{\star}(s) \) and Q-functions \( Q^{\star}(s, a) \) satisfy the Bellman optimality conditions: \(V^{\star}(s) = \max_a Q^{\star}(s, a),\, 
    Q^{\star}(s, a) = r(s, a) + \gamma \mathbb{E}_{s' \sim \mathcal{P}} \left[ V^{\star}(s') \right]\). The optimal policy satisfies \(\pi^{\star}(s) = \arg\max_a Q^{\star}(s, a)\).

\subsection*{In-sim Reinforcement Learning}\label{sec:background_in-sim_rl}
We consider a simulation model as an abstract function \( {\mathcal P}_\theta \), approximating the real-world distribution $\mathcal{P}$, used to generate the next state given a current state-action pair \({{s}' \sim {\mathcal P}_\theta(\cdot \mid s, a)}\) and to generate simulation-based trajectories mimicking the real system. The parameters \(\theta\) may include the parameters of the functions defining the simulation dynamics.  We consider a reward function \( R_{\theta}(s, a) \) in the simulation as a model of the real-world reward $r$. The parameters of \(R_\theta\) and \({\mathcal P}_\theta\) together constitute what we refer to as simulation parameters \(\theta\). The in-sim policy \(\pi_\phi\) is an implicit function of simulation parameters \(\theta\), derived by solving:
\begin{equation}\label{eq:pi_phi}
    {\phi^\star} = \arg\max_{\phi} \mathbb{E}_{\tau \sim \pi_\phi, {\mathcal P}_\theta} \left[\sum_{t=0}^{\infty} \gamma^t R_{\theta}(s_t, a_t) \right]\,.
\end{equation}
The objects \(V_\theta, Q_\theta\), and \(\pi_\phi\) are implicitly defined by \({\mathcal P}_\theta\) and \(R_\theta\). The optimal policy \(\pi_{\phi^\star}\) can be derived using the Bellman equations: \(
V^{\star}_{\theta}(s) = \max_{a} Q^{\star}_{\theta}(s,a),\,
Q^{\star}_{\theta}(s,a) = R_{\theta}(s,a) + \gamma\, \mathbb{E}_{{s}' \sim {\mathcal P}_{\theta}} \!\left[ V^{\star}_{\theta}(s') \right],\, 
\pi_{\phi^\star}(s) = \arg\max_{a} Q^{\star}_{\theta}(s,a).\) 

\textbf{Notation:} Throughout the paper, quantities indexed by \(\theta\) denote in-simulation quantities, whereas quantities without a \(\theta\) subscript denote the corresponding real-world quantities.

\section{Sim-to-Real Optimality}\label{sec:sim_to_real_optimality}
In this section, we establish the notion of \emph{sim-to-real optimality}.  Assuming the real-world and in-sim \acp{mdp} defined in Section~\ref{sec:background_in-sim_rl} share the same state-space $\mathcal{S}$ and have a unique optimal policy, the in-sim \ac{mdp} achieves sim-to-real optimality if the optimal policy obtained in simulation is also optimal for the real-world environment, i.e., \(\pi_{\phi^\star} = \pi^\star\). This can be expressed as the necessary and sufficient optimality condition:
\begin{equation}
\label{eq:necessary_condition}
\mathrm{arg}\max_{{a}} \, Q_{\theta}^\star({s}, {a}) = \mathrm{arg}\max_{{a}} \, Q^\star({s}, {a}),\quad \forall\,{s} \,.
\end{equation}
Sim-to-real optimality is critical in applications where real-world performance is the priority. Conventionally, the performance objective is pursued indirectly by constructing accurate simulation models of real-world dynamics via supervised learning and then training policies on these models.
However, for \emph{stochastic} systems, predictive accuracy of the simulation model does not necessarily correlate with the real-world performance of the resulting in-sim policy~\citep{anand2025predicting}. This issue can persist even under good data coverage, because prediction-oriented model training is not informed by which aspects of the true stochastic dynamics matter for decision-making. It is particularly important in problems with economic or non-smooth objectives, where optimal decisions may depend on features of the transition distribution that prediction-based identification need not preserve~\citep{anand2025predicting}.

However, it has been shown that even an in-sim policy under an imperfect simulation model \({\mathcal P}_\theta\)  can achieve sim-to-real optimality under specific conditions on the model or on the in-sim reward function~\citep {anand2025predicting, gros2019data}. Following this observation, we propose framing the identification of the optimal simulation parameters~$\theta^\star$ satisfying the necessary condition for sim-to-real optimality in~\eqref{eq:necessary_condition} as a policy gradient \ac{rl} problem:
\begin{equation}\label{eq:bi_level_rl}
\begin{aligned}
\theta^{\star} \in \arg\max_{\theta} \quad 
& \mathbb{E}_{\tau \sim \pi_{\phi^\star}, \mathcal{P}} 
\left[ \sum_{t=0}^{\infty} \gamma^t r(s_t, a_t) \right] \\
\text{s.t.} \quad 
& \phi^\star \in \arg\max_{\phi} \quad 
\mathbb{E}_{\tau \sim \pi_\phi, \mathcal{P}_\theta} 
\left[ \sum_{t=0}^{\infty} \gamma^t R_{\theta}(s_t, a_t) \right].
\end{aligned}
\end{equation}

The \ac{rl} formulations in~\eqref{eq:pi_phi} and~\eqref{eq:bi_level_rl} together form a bi-level \ac{rl} problem. The inner level constitutes the in-sim \ac{rl} problem to derive \(\pi_{\phi^\star}\), which is implicitly defined by \({\mathcal P}_\theta\) and \(R_\theta\) \eqref{eq:pi_phi}.  Upon deployment in the real-world environment, the policy generates real transition data \((s,a,s_+,r)\). The outer-level \ac{rl} then adapts the simulator dynamics~${\mathcal P}_\theta$ and reward function~$R_{\theta}$ to optimize real-world policy performance using the real-world transition data. The outer-level \ac{rl} does not fit the simulation parameters to mimic real-world dynamics, but instead optimizes them to maximize the performance of the policy. 

Additionally, the solution \(\theta^\star\) to~\eqref{eq:bi_level_rl} is in general not unique. This allows selecting, among the simulation parameters that yield the same optimal policy performance, those that are also consistent with the real-world dynamics according to a user-specified loss \(\mathcal{L}(\cdot)\), i.e.,
\(
\theta^\star
=
\arg\min_{\theta}\, \mathcal{L}\!\left(\mathcal P_\theta,\mathcal P;\,R_\theta,r\right)
\quad
\mathrm{s.t.}
\quad
\eqref{eq:bi_level_rl}\ \text{holds for all } s,a .
\)
Here, predictive accuracy is not introduced as a competing objective, but only as a selection or regularization criterion within the set of optimal models  defined by \(\theta^\star\). The existence of such optimal models within the support of prediction-oriented models is justified in Proposition~1 of~\citep{anand2025predicting}. Common choices of \(\mathcal{L}(\cdot)\) include \ac{mle}, Bayesian estimation, or expected-value losses such as MSE. In practice, this can be implemented by regularizing the policy-gradient step induced by~\eqref{eq:bi_level_rl} with \(\nabla_\theta \mathcal{L}\).

\section{Bi-level RL}\label{sec:bi_level_rl}
In this section, we derive the mathematical tools to solve the bi-level \ac{rl} problem \eqref{eq:bi_level_rl}. We begin by stating the regularity conditions required in the following assumption.
\begin{assumption}[]
\label{ass:regularity}
The policy $\pi_\phi(a|s)$, simulation model $\mathcal{P}_\theta(s'|s,a)$, and reward function $R_\theta(s,a)$ are continuously differentiable with respect to their respective parameters $\phi$ and $\theta$.
\end{assumption}
For generality across continuous and discrete state-action domains and stochastic simulation models, we consider that the in-sim and real-world \ac{rl} agents are trained using \ac{spg}, satisfying:
\begin{align}
\label{eq:inner_pg}
\hat \varphi(\phi, \theta) &= \mathbb{E}_{s \sim \hat{\rho}_{\theta}^{\pi_{\phi}}}[\nabla_{\phi}\log\pi_{\phi}(a|s){Q}_{\theta}^{\pi_{\phi}}(s,a)]\,,
\end{align}
where $Q_{\theta}^{\pi_{\phi}}$ is the critic of policy $\pi_{\phi}$ \textit{as seen} by the simulations, and $ \hat\rho_{\theta}^{\pi_{\phi}}$ is the simulated discounted Markov chain density associated with policy $\pi_{\phi}$ under the model \({\mathcal P}_\theta\). 
\begin{assumption}
\label{ass:local_regularity}
We assume that the in-sim \ac{spg} iterates have locally converged to a stationary policy parameter \(\phi\) satisfying \(\hat{\varphi}(\phi,\theta)=0\). We further assume that \(\hat{\varphi}(\phi,\theta)\) is continuously differentiable in \((\phi,\theta)\), that the stationary solution \(\phi\) is locally isolated, and that the Jacobian \(\nabla_\phi \hat{\varphi}(\phi,\theta)\) is nonsingular.
\end{assumption}

Under Assumption~\ref{ass:local_regularity}, \ac{spg} yields a locally optimal policy \(\pi_\phi\) for the simulation model, and the bi-level \ac{rl} problem can be written as
\begin{equation}\label{eq:bi_level_rl_aprox}
    \theta^\star = \arg\max_{\theta} \mathbb{E}_{\tau \sim \pi_{\phi}, \mathcal{P}} \left[ \sum_{t=0}^{\infty} \gamma^t r(s_t, a_t) \right]
    \quad \text{s.t.} \quad \hat \varphi(\phi, \theta) = 0 \, .
\end{equation}
The local regularity assumption is reasonable in practice, since \ac{spg} methods typically converge to locally stable stationary policies with non-degenerate curvature~\citep{agarwal2021theory}. The constraint \(\hat \varphi(\phi,\theta)=0\) therefore locally defines \(\phi\) as a unique differentiable function of \(\theta\). Consequently, the sensitivity of the in-sim policy parameters \(\phi\) with respect to simulation parameters \(\theta\) can be obtained via the \acf{ift}~\citep{krantz2002implicit},
\begin{equation}\label{eq:ift}
\nabla_\theta \phi 
\;=\;
-\,\bigl(\nabla_\phi \hat{\varphi}\bigl(\phi,\theta\bigr)\bigr)^{-1}
\,
\nabla_\theta \hat{\varphi}\bigl(\phi,\theta\bigr)
\,.
\end{equation}

Note that there exist iterative methods and their approximations to invert the Jacobian \(\nabla_\phi \hat{\varphi}\bigl(\phi,\theta\bigr)\)~\citep{liesen2013krylov, yang2024tuning}. Using \(\nabla_\theta \phi\), we can then formulate a solution to \eqref{eq:bi_level_rl_aprox} as the outer-level \ac{spg}  using the real-world transition data:
\begin{subequations}
\begin{align}
\varphi\bigl(\phi,\theta\bigr)
&:=
\mathbb{E}_{s\sim\rho^{\pi_{\phi}}}\Bigl[\nabla_{\theta}\log\pi_{\phi}(a\mid s)\,Q^{\pi_{\phi}}(s,a)\Bigr], \label{eq:real_world_spg}\\
\nabla_{\theta}\log\pi_{\phi}(a\mid s)
&=
\nabla_{\theta}\phi\;\nabla_{\phi}\log\pi_{\phi}(a\mid s) \label{eq:nabla_theta_log_pi}\,,
\end{align}
\end{subequations}
where $Q^{\pi_{\phi}}$ is the critic of the in-sim policy $\pi_{\phi}$ \textit{as seen} by the real world, $\rho^{\pi_{\phi}}$ is the discounted Markov chain density resulting from using in-sim policy $\pi_{\phi}$ in the real world, and \eqref{eq:nabla_theta_log_pi} is the result of the chain-rule with the \ac{ift} condition. To update the simulation parameters via \eqref{eq:real_world_spg}, we must compute the sensitivity of the in-sim \ac{rl} solution with respect to simulation parameters, given by the gradient $\nabla_{\theta}\phi$. Computing \(\nabla_\theta \phi\) is the only additional component of \eqref{eq:real_world_spg} compared to the classic \ac{spg}. 

\subsection{Sensitivity Analysis of Stochastic Policy Gradient}
We now introduce two lemmas that will help us compute \(\nabla_{\theta}\phi\) and derive the full expression for the real-world \ac{spg} in \eqref{eq:real_world_spg}. Note that the $\partial$ in Lemma~\ref{lem:mc_sensitivity} and Lemma~\ref{lem:policy_eval_sensitivity} denotes a partial derivative with respect to either $\phi$ or $\theta$.

\begin{lemma}[Sensitivity of a Markov Chain]
\label{lem:mc_sensitivity}
% Let $\pi_\phi(a\mid s)$ be a differentiable stochastic policy parameterized by $\phi$, 
% let $\mathcal P_\theta(s'\mid s,a)$ be a differentiable transition model parameterized by $\theta$, 
% and let $\rho^0$ denote the initial state distribution. 
% The policy and transition model induce a Markov chain with discounted state distribution 
% $\hat{\rho}_{\theta,k}^{\pi_\phi}$ at time step $k$. For any bounded function $\eta(s_k,a_k)$, the sensitivity of its expected value with respect to a parameter 
% (denoted by $\partial$) can be written as
Let \(\pi_\phi(a\mid s)\) be a differentiable stochastic policy parameterized by \(\phi\), let \(\mathcal P_\theta(s'\mid s,a)\) be a differentiable transition model parameterized by \(\theta\), and let \(\rho^0\) denote the initial state distribution. For convenience, define the closed-loop one-step transition density
\(
\mathcal P^{\pi_\phi}_\theta(s_{i+1},a_i\mid s_i)
\;\triangleq\;
\pi_\phi(a_i\mid s_i)\,\mathcal P_\theta(s_{i+1}\mid s_i,a_i).
\)
The policy and transition model induce a Markov chain with discounted state distribution \(\hat{\rho}_{\theta,k}^{\pi_\phi}\) at time step \(k\). Then, for any bounded measurable function \(\eta(s_k,a_k)\), the sensitivity of its expected value with respect to a parameter (denoted by \(\partial\)) can be written as
\begin{align}\label{eq:mc_sensitivity_expected_value_form}
\partial \, \mathbb E_{s\sim \hat\rho_{\theta,k}^{\pi_{\phi}}} \bigl[\eta(s_k,a_k)\bigr] = \mathbb{E}_{\mathclap{\substack{\\[4mm]s_0\sim\rho^0,\,a_i\sim\pi_\phi,s_{i+1}\sim \mathcal{P}^{\pi_\phi}_{\theta}}}}
   \Bigl[\,
     \eta(s_k,a_k)\,\Bigl(
        \partial\log\pi_\phi(a_k\!\mid\!s_k)
      + \sum_{i=0}^{k-1}\partial\log \mathcal{P}^{\pi_\phi}_\theta(s_{i+1}\!\mid\!s_i,a_i)
     \Bigr)
   \Bigr]\,. 
\end{align}
\end{lemma}
Here, we assume that the policy and transition densities are strictly positive on the support of trajectories induced by the Markov chain, ensuring that log-likelihood derivatives in \eqref{eq:mc_sensitivity_expected_value_form} remain finite. The proof of Lemma~\ref{lem:mc_sensitivity} is provided in Appendix~\ref{appendix:proof_lemma_markov_chain}.

\begin{lemma}[Sensitivities of \(Q^{\pi}\)]
\label{lem:policy_eval_sensitivity}
Let $(s_t,a_t)_{t\ge 0}$ be samples from the in-simulation dynamics ${\mathcal P}_\theta(\cdot\mid s_t,a_t)$ and the policy
$\pi_\phi(\cdot\mid s_{t})$, with $(s_0,a_0)=(s,a)$. Then the sensitivity of the simulation-based critic with respect to the reward and model parameters \(\theta\) under a fixed policy \(\pi_\phi\)
is given by
\begin{equation}
\label{eq:critic_sensitivity_theta}
\begin{aligned}
\nabla_\theta Q^{\pi_\phi}_\theta(s,a)
&= \mathbb{E}_{\substack{s\sim\hat\rho^{\pi_\phi}_\theta,\\a\sim\pi_\phi}}\Bigg[
\sum_{t=0}^{\infty} \gamma^t\,\nabla_\theta R_\theta(s_t,a_t)
\\
&\quad \,\,
\left.+\;\sum_{t=0}^{\infty} \gamma^{t+1}\,
\nabla_\theta\log {\mathcal P}_\theta(s_{t+1}\mid s_t,a_t)\,
Q^{\pi_\phi}_\theta(s_{t+1}, a_{t+1})
\ \right|\ s_0=s,\ a_0=a \Bigg]\,,
\end{aligned}
\end{equation}
and with respect to the policy parameters \(\phi\) under fixed reward \(R_\theta\) and model \({\mathcal P}_{\theta}\) is given by
\begin{align}\label{eq:critic_sensitivity_phi}
\nabla_\phi Q^{\pi_\phi}_\theta(s,a)
=
\mathbb E_{\substack{s\sim\hat\rho^{\pi_\phi}_\theta\\a\sim\pi_\phi}}\left[
\sum_{t=0}^{\infty} \gamma^{t+1}\,
\nabla_\phi\log\pi_\phi(a_{t+1}\mid s_{t+1})\,
Q^{\pi_\phi}_\theta(s_{t+1},a_{t+1})
\ \middle|\ s_0=s,\ a_0=a
\right].
\end{align}

% Expectations there are taken with respect to the on–policy state distribution
% $s\sim\hat\rho^{\pi_\phi}_\theta$ (and $a\sim\pi_\phi(\cdot\mid s)$).
\end{lemma}
The proof of Lemma~\ref{lem:policy_eval_sensitivity} is provided in Appendix~\ref{appendix:critic_sensitivity}.
Note that score-terms in the critic sensitivities \eqref{eq:critic_sensitivity_theta} and \eqref{eq:critic_sensitivity_phi} can induce high variance, so variance reduction methods such as a baseline or the reparameterization trick should be used. A reparameterization version of the critic sensitivity terms \eqref{eq:critic_sensitivity_theta} and \eqref{eq:critic_sensitivity_phi} is provided in Appendix~\ref{appendix:critic_sensitivity}. 

There exists an alternative method to estimate the critic sensitivities, which is provided in Lemma~\ref{lem:policy_eval_sensitivity_QV_version} in Appendix~\ref{appendix:alternative_sensitivity}. This avoids unrolling entire trajectories as required in Lemma~\ref{lem:policy_eval_sensitivity} by directly differentiating the Bellman equations. However, this approach is computationally less efficient than the method in Lemma~\ref{lem:policy_eval_sensitivity}, as it requires iteratively estimating sensitivities for both the \(Q\) and \(V\) functions. For completeness, we provide a proof of equivalence between these two sensitivity estimation approaches in Appendix~\ref{appendix:sensitivities-equivalency}.

Using these two lemmas, we now present the main result on the sensitivity analysis of \ac{spg} in the following theorem.

\begin{theorem}[Sensitivity of in-sim \ac{spg}] \label{Th:bi_level_spg}
Consider a simulation model ${\mathcal P}_\theta$, reward function $R_\theta$, and the resulting in-sim policy $\pi_\phi$ obtained by solving the in-simulation \ac{rl} problem using \ac{spg}~\eqref{eq:inner_pg} in an actor-critic setting. Assume that the policy parameters $\phi$ correspond to a locally stationary solution of the \ac{spg} update, i.e., $\hat{\varphi}(\phi,\theta)=0$. Then the sensitivity of the in-sim policy parameters $\phi$ with respect to the simulation model and reward parameters $\theta$ is given by
\begin{align}\label{eq:nabla_theta_phi}
\nabla_\theta \phi 
&= -\Bigl(
   \underbrace{\mathbb E_{s\sim \hat\rho_{\theta}^{\pi_{\phi}}}
     \bigl[\nabla_{\phi\phi}\log\pi_{\phi}\,Q_{\theta}^{\pi_{\phi}}
          + \nabla_{\phi}\log\pi_{\phi}\,\nabla_{\phi}Q_{\theta}^{\pi_{\phi}}\bigr]}
   _{\substack{\mathrm{Gradient \,of\, } \hat{\varphi}\mathrm{\, w.r.t.\, }\phi}}
   \;+\;
   \underbrace{\nabla_{\phi}\mathbb E_{s\sim \hat\rho_{\theta}^{\pi_{\phi}}}
     \bigl[\nabla_{\phi}\log\pi_{\phi}\,Q_{\theta}^{\pi_{\phi}}\bigr]}
   _{\substack{\mathrm{Gradient \,of \,dist.\, }\hat\rho_{\theta}^{\pi_{\phi}}\mathrm{\,w.r.t.\, }\phi}}
\Bigr)^{-1}
\nonumber\\
&\qquad\qquad
\Bigl(
   \underbrace{\mathbb E_{s\sim \hat\rho_{\theta}^{\pi_{\phi}}}
     \bigl[\nabla_{\phi}\log\pi_{\phi}\,\nabla_{\theta}Q_{\theta}^{\pi_{\phi}}\bigr]}
   _{\substack{\mathrm{Gradient\, of\, } \hat{\varphi}\mathrm{ \,w.r.t.\, }\theta}}
   \;+\;
   \underbrace{\nabla_{\theta}\mathbb E_{s\sim \hat\rho_{\theta}^{\pi_{\phi}}}
     \bigl[\nabla_{\phi}\log\pi_{\phi}\,Q_{\theta}^{\pi_{\phi}}\bigr]}
   _{\substack{\mathrm{Gradient\, of\, dist.\, }\hat\rho_{\theta}^{\pi_{\phi}}\mathrm{\,w.r.t.\, }\theta}}
\Bigr).
\end{align}
\end{theorem}

Here, \( \nabla_\phi\mathbb E_{s\sim \hat\rho_{\theta}^{\pi_{\phi}}}\left[ \cdot \right]\) is the effect of changing $\phi$ on the simulation-based distribution $ \hat\rho_{\theta}^{\pi_{\phi}}$, and the influence of that change on the expected value \textit{under a fixed cost and model}.  \( \nabla_\theta\mathbb E_{s\sim \hat\rho_{\theta}^{\pi_{\phi}}}\left[ \cdot \right]\) is the effect of changing $\theta$ on the simulation-based distribution $ \hat\rho_{\theta}^{\pi_{\phi}}$, and the influence of that change on the expected value \textit{under a fixed policy}. The terms in the second line of~\eqref{eq:nabla_theta_phi} describe how perturbations to the simulation parameters modify both the value estimates and the distribution of simulated trajectories, which alters the policy gradient. The inverse Jacobian term in the first line of~\eqref{eq:nabla_theta_phi} maps these changes in the policy gradient to the resulting shift in the locally optimal policy parameters. A detailed and intuitive explanation of each of the terms in~\eqref{eq:nabla_theta_phi} is provided in the first part of Appendix~\ref{appendix:main_proof}.

In practice, all components in Theorem~\ref{Th:bi_level_spg} will be estimated from the in-sim transition data \((\hat{s}, a, \hat{s}_+, R)\) generated from model sampling. Let \((s_0^j,a_0^j,R_0^j,\ldots,s_N^j,a_N^j,R_N^j)\) be the sampled trajectory $j\in\left\{1,\ldots,n\right\}$ of length $N$ that should be long enough for Lemma~\ref{lem:policy_eval_sensitivity} to apply. The expected value approximation of the  Markov Chain sensitivity of the in-sim policy gradient with respect to \(\theta\) and \(\phi\) is given by:
\begin{align} \label{eq:mc_pg_phi_sensitivity_expected_value_form}
{\nabla_\phi}\mathbb E_{s\sim \hat\rho_{\theta}^{\pi_{\phi}}}\left[\nabla_{\phi}\log\pi_{\phi}\, Q_{\theta}^{\pi_{\phi}}\right] \approx
 \frac{1}{nN}\sum_{k=0}^N \sum_{j=1}^n &\nabla_{\phi}\log\pi_{\phi}\left(\left.a^j_k \right | s^j_k\right)\, Q_{\theta}^{\pi_{\phi}}\left(s_k^j,a_k^j\right) \notag\\
&\otimes \left( W_{\phi, k}^j +  \nabla_\phi \log\pi_{\phi}\left(\left.a_k^j\right|s_k^j\right)  \right) 
\end{align}
where: $\otimes$ represents the outer product and $W^j_{\phi, 0} = 0;$\,  \(W_{\phi, k}^j  = W_{\phi, k-1}^j  +  \nabla_\phi\log \pi_{\phi}\left(\left.a^j_{k-1}\right |s^j_{k-1}\right)\). Similarly,
\begin{align}\label{eq:mc_pg_theta_sensitivity_expected_value_form}
\nabla_\theta\mathbb E_{s\sim \hat\rho_{\theta}^{\pi_{\phi}}}\left[\nabla_{\phi}\log\pi_{\phi}\, Q_{\theta}^{\pi_{\phi}}\right] \approx
\frac{1}{nN}\sum_{k=0}^N \sum_{j=1}^n &\nabla_{\phi}\log\pi_{\phi}\left(\left.a^j_k \right | s^j_k\right)\, Q_{\theta}^{\pi_{\phi}}\left(s_k^j,a_k^j\right) \otimes  W_{\theta, k}^j  
\end{align}
where $W^j_{\theta, 0} = 0$ and \( W_{\theta, k}^j = W_{\theta, k-1}^j  + {\nabla_\theta}\log  {\mathcal P}_\theta\left(\left.s^j_{k}\right|s^j_{k-1},a^j_{k-1}\right)
\).

A high-level algorithmic description of the bi-level \ac{rl} approach is provided in Algorithm~\ref{alg:bi_level_rl}.

\subsection{Convergence Analysis and Efficient Implementation of Bi-level RL}\label{sec:convergence_and_implementation}

\paragraph{Convergence Analysis. }
A detailed convergence analysis for the bi-level \ac{rl} approach is provided in Appendix~\ref{appendix:convergence_analysis}. 
The analysis establishes almost-sure convergence of the proposed bi-level \ac{spg} scheme under a three-timescale stochastic-approximation scheme: the critic and its sensitivities evolve on the fastest timescale, the in-sim policy on an intermediate timescale, and the outer-level update of simulation parameters on the slowest timescale. An important practical implication of this analysis for Algorithm~\ref{alg:bi_level_rl} is that the in-sim policy does \emph{not} need to be fully optimized before each update to the simulation parameters. 
Instead, it is enough that the in-sim policy is updated on a faster timescale than the outer level, so that it remains sufficiently close to the local optimum corresponding to the current simulation parameters. 
This provides formal justification for taking outer-level steps with partially updated in-sim policies, provided that the critic and sensitivity errors remain controlled.

\paragraph{Efficient Implementation.}
The inverse of a Jacobian operator in the policy parameter space \((\nabla_\phi \hat{\varphi})\) in~\eqref{eq:nabla_theta_phi} can often be high-dimensional in deep \ac{rl} settings. Although~Theorem~\ref{Th:bi_level_spg} yields the full sensitivity matrix $\nabla_\theta \phi\in\mathbb{R}^{d_\phi\times d_\theta}$, a practical implementation of the outer update in~\eqref{eq:real_world_spg} does \emph{not} require forming this matrix explicitly. Since the outer objective (the real-world return)~\eqref{eq:bi_level_rl_aprox} is scalar, the required quantity is the vector-Jacobian product $(\nabla_\theta \phi)^\top \nabla_\phi \mathcal{L}$, where $\mathcal{L}$ denotes the outer objective in~\eqref{eq:bi_level_rl_aprox}. Using an adjoint variable, this product can be computed by (i) solving a single linear system involving $(\nabla_\phi \hat{\varphi})^\top$ and (ii) evaluating one additional vector--Jacobian product through the residual $\hat{\varphi}(\phi,\theta)$. This approach avoids constructing any dense $d_\phi\times d_\theta$ objects, and fits naturally with reverse-mode automatic differentiation. The details of this approach and a comparison of the computational efficiency are provided in Appendix~\ref{appendix:adjoint_ift}.
\begin{algorithm}[!t]
  \caption{Bi-Level SPG Framework}
  \label{alg:bi_level_rl}
  \setstretch{0.85}
  \begin{algorithmic}      % <— no [1] here, so no line numbers anywhere
    \State \textbf{Given:} initialized 
      ${\mathcal P}_\theta(s_+ \mid s, a)$, 
      $R_\theta(s, a)$, 
      $\pi_\phi(a \mid s)$, 
      $V^{\pi_\phi}(s)$, 
      and $Q^{\pi_\phi}(s,a)$.
    \For{each bi-level RL iteration}
      \State Train in-sim policy $\pi_\phi$ via SPG for \(N\) steps (Eq.~\eqref{eq:inner_pg}).
      \State Roll out $\pi_\phi$ over ${\mathcal P}_\theta,R_\theta$ to collect in-sim transitions $(s,a,s_+,R)$.
      \State Compute $\nabla_\theta\phi$ from these in-sim transitions (Eq.~\eqref{eq:nabla_theta_phi}).
      \State Roll out $\pi_\phi$ on the real world $\mathcal P$ to collect transitions $(s,a,s_+,r)$.
      \State Update $\theta$ according to Eq.~\eqref{eq:real_world_spg}.
    \EndFor
  \end{algorithmic}
\end{algorithm}

\subsection{Bi-level PPO}\label{sec:bi_level_ppo}
Using the sensitivity analysis in Theorem~\ref{Th:bi_level_spg}, we now formulate a bi-level version of the \ac{ppo} algorithm~\citep{schulman2017proximal} that updates the simulation parameters toward achieving sim-to-real optimality. Consider the in-sim \ac{ppo} objective given by: 
\begin{equation} \label{eq:ppo1}
\mathcal{L}_{\theta}(\phi) := \mathbb{E}_{\substack{s\sim\hat\rho^{\pi_\phi}_\theta,\\a\sim\pi_\phi}}\left[
\frac{\pi_\phi(a \mid s)}{\pi_{\bar\phi} (a \mid s)} 
A^{\pi_\phi}_{\theta}(s,a)
+ \beta \,\mathcal{H}(\pi_\phi(\cdot \mid s))
\right],
\end{equation}
where $\bar\phi$ and \(\phi\) denote the policy parameters before and after the current update step, and $A^{\pi_\phi}_{\theta}$ denotes the in-sim advantage, and $\mathcal{H}$ is the policy entropy with $\beta$ as the entropy coefficient. In \eqref{eq:ppo1}, the practical components in a typical \ac{ppo} implementation, such as generalized advantage estimation, clipping the importance sampling ratio, and policy divergence, are omitted for clarity. The in-sim policy is updated according to: \(\phi \leftarrow \phi - \alpha \nabla_\phi \mathcal{L}_\theta(\phi),\)
which yields the approximate fixed point $\phi(\theta)$ once converged sufficiently. The corresponding outer-level \ac{ppo} objective for updating $\theta$ is given by:
\begin{equation}
\mathcal{L}(\phi, \theta) := \mathbb{E}_{\substack{s\sim \rho^{\pi_\phi}_\theta,\\a\sim\pi_\phi}}\left[
\frac{\pi_\phi(a \mid s)}{\pi_{\bar\phi} (a \mid s)}
A^{\pi_\phi}(s,a)
+ \beta \,\mathcal{H}(\pi_\phi(\cdot\mid s))
\right]\,,
\end{equation}
where $A^{\pi_\phi}$ denotes the outer-level advantage function. 
The outer-level parameter update is $\theta \leftarrow \theta - \alpha \nabla_\theta \mathcal{L}(\phi,\theta)$,
% Based on the sensitivity analysis provided in Section~\ref{sec:sim_to_real_optimality},
with the outer-level gradient $\nabla_\theta \mathcal{L}(\phi,\theta) = (\partial \mathcal{L}(\phi,\theta)/\partial \phi)\,\nabla_\theta \phi(\theta)$, where $\nabla_\theta \phi(\theta)$ is obtained from~\eqref{eq:nabla_theta_phi} by replacing \(Q^{\pi_\phi}_\theta\) with \(A^{\pi_\phi}_\theta\). A pseudo-code of the bi-level \ac{ppo} algorithm is provided in Algorithm~\ref{alg:bi_level_ppo}.

\begin{algorithm}[h]
  \caption{Bi-Level PPO Algorithm}
  \label{alg:bi_level_ppo}
  \setstretch{0.85}
  \begin{algorithmic}
    \State Initialize simulation parameters $\theta$, 
      policy parameters $\phi$, 
      inner critic $V_{\theta}$, 
      outer critic $V$.
    \For{$k = 1, \dots, N_{\text{outer}}$} \Comment{Outer-level update of simulation parameters}
      \For{$j = 1, \dots, N_{\text{inner}}$} \Comment{In-sim PPO training}
        \State Roll out $\pi_\phi$ in the simulation 
          $\mathcal{P}_\theta, R_\theta$ to collect 
          $(s, a, s_+, R)$.
        \State Estimate returns and in-sim advantages $A_{\theta}$; 
          update $V_{\theta}$.
        \State Update $\phi$.
      \EndFor
      \State Roll out $\pi_{\phi}$ in simulator 
          $\mathcal{P}_\theta, R_\theta$ to collect 
          $(s, a, s_+, R)$.
      \State Compute $\nabla_\theta \phi(\theta)$ from in-sim (Eq.~\eqref{eq:nabla_theta_phi}). 
      \State Roll out $\pi_{\phi}$ on the real system $\mathcal{P}$ 
        to collect $(s, a, s_+, r)$.
      \State Estimate returns and real-world advantages $A$; 
        update $V$.
      \State Update $\theta \leftarrow \theta - \alpha \nabla_\theta \mathcal{L}(\phi,\theta)$.
    \EndFor
  \end{algorithmic}
\end{algorithm}
In practice, \ac{ppo} performs stochastic updates that approach a neighborhood of a local stationary point of the surrogate objective \eqref{eq:ppo1}. The derived sensitivity therefore describes how small perturbations in simulator parameters change the \ac{ppo} policy in this neighborhood. Learning can further be accelerated by performing multiple simulator updates per outer iteration, similar to standard \ac{ppo}. This would require additionally updating the policy parameters at the outer level based on real-world data, allowing the outer-level policy updates to serve as a proxy for the fully optimized in-sim policy, while also providing better policy initializations for the inner level. This strategy is effective when small simulator perturbations induce smooth, local changes in policy parameters.

\section{Illustrative Examples}\label{sec:examples}
We present three illustrative realizations of the bi-level \ac{rl} approach: (i) stochastic control in a discrete state–action space, (ii) stochastic continuous control with linear dynamics, and (iii) a quadcopter stabilization task. The first two examples validate the sensitivity analysis in \eqref{eq:nabla_theta_phi} by evaluating the outer-level \ac{spg} updates, while the third demonstrates the bi-level \ac{ppo} algorithm. In each case, we initialize five simulation models with parameters randomly sampled around the real-world environment parameters. These experiments are intended solely to illustrate the proposed framework rather than serve as a performance benchmark. Additional experimental details are provided in Appendix~\ref{appendix:experiments}.

\paragraph{Discrete MDP. }\begin{wrapfigure}{r}{0.3\textwidth}
  \centering
  % Scale down the whole picture and reduce node distance and node size
\begin{tikzpicture}[scale=0.65, every node/.style={scale=0.65}, node distance=1.5cm, auto, >=Stealth, thick, font=\Large]
  % Nodes
  \node[draw, circle, fill=blue!20,   minimum size=1cm] (s0) at (0,0)       {$s_0$};
  \node[draw, circle, fill=red!20,    minimum size=1cm] (s1) at (1.5,1.5)   {$s_1$};
  \node[draw, circle, fill=green!20,  minimum size=1cm] (s2) at (3,0)       {$s_2$};

  % Bidirectional "1" transitions
  \draw[<->, blue]           (s0) edge[bend left=20] node[above] {1} (s1);
  \draw[<->, red]            (s1) edge[bend left=20] node[right] {1} (s2);
  \draw[<->, green!70!black] (s2) edge[bend left=20] node[below] {1} (s0);

  % Loops on "0"
  \draw[->, blue]           (s0) edge[loop left]  node[left]  {0} (s0);
  \draw[->, red]            (s1) edge[loop above] node[above] {0} (s1);
  \draw[->, green!70!black] (s2) edge[loop right] node[right] {0} (s2);
\end{tikzpicture}
  \caption{Discrete MDP.}
  \label{fig:three_state_system}
\end{wrapfigure}
We consider a stochastic discrete MDP with state and action sets $\mathcal{S}=\{s_0,s_1,s_2\}$ and $\mathcal{A}=\{\mathtt{0},\mathtt{1}\}$, where each action may yield different successor states. The state-transition model includes the parameters: ${\theta^f_{s,a}=(\theta^f_{s,a,0},\theta^f_{s,a,1},\theta^f_{s,a,2})\in\mathbb{R}^3}$ via  
\({\displaystyle {\mathcal P}_\theta(s'\mid s,a)=\mathrm{softmax}(\theta^f_{s,a})_{s'}}\).  
Rewards use $\theta^R\in\mathbb{R}^{3\times2}$ via $R_\theta(s,a)=\theta^R_{s,a}$.  In this example, we use \ac{dp} to compute exact solutions of a discrete in-sim policy and its value function. This discrete policy is converted to a stochastic policy that uses action preferences (logits) \({\displaystyle h_\phi(s,a)=\phi^\pi_{s,a}}\), i.e., the learnable score for action $a$ in state $s$, with ${\theta^\pi\in\mathbb{R}^{3\times2}}$. The policy selects actions by  
\({\displaystyle \pi_\theta(a\mid s)=\mathrm{softmax}\bigl(h_\theta(s,\cdot)\bigr)_a}\). Then we estimate the critic sensitivity \eqref{eq:critic_sensitivity_theta} and \eqref{eq:critic_sensitivity_phi} in a tabular form iteratively until convergence using on-policy in-sim data. We use the same in-sim on-policy data to estimate the Markov chain sensitivities \eqref{eq:mc_pg_phi_sensitivity_expected_value_form} and \eqref{eq:mc_pg_theta_sensitivity_expected_value_form}.

\paragraph{Continuous MDP. }
We consider a linear system with additive Gaussian noise; \({s_{+} = \theta_s\, s + \theta_a\, a + \epsilon,\quad \epsilon \sim \mathcal{N}(0,\sigma^2)}\) and a parameterized exponential reward function \({R_{\theta}(s,a)\;=\;\exp\left(-\lambda\bigl(\theta_q\,s^2 \;+\;\theta_r\,a^2\bigr)\right)}\). The real system parameters are: \({\theta_s =1,\;  \theta_a=1,\; \theta_q=1,\; \theta_r=1}\). The simulation model and reward parameters are chosen randomly between $0$ and $1$. We solve the in-sim policy and value function exactly using \ac{lqr}, and then use \acp{nn} to approximate them. By approximating them with \acp{nn}, we can validate the robustness of the sensitivity estimates with regard to the noisy \ac{nn} predictions. Then we estimate the critic sensitivity \eqref{eq:critic_sensitivity_theta} and \eqref{eq:critic_sensitivity_phi} for each sample iteratively until convergence using in-sim data under the \ac{nn} policy, and the same data to estimate the Markov chain sensitivities \eqref{eq:mc_pg_phi_sensitivity_expected_value_form} and \eqref{eq:mc_pg_theta_sensitivity_expected_value_form}.

During the bi-level \ac{rl} training, the simulation parameters are iteratively updated as described in Algorithm~\ref{alg:bi_level_rl}, until convergence. Since these two problems are solved using exact methods, their resulting policies naturally satisfy the convergence condition for the \ac{spg}: \({\hat \varphi(\phi, \theta) = 0}\).   The training progress of the outer-level \ac{rl} is shown in Fig.~\ref{fig:results}. Its convergence serves as a validation of the sensitivity estimation in Theorem~\ref{Th:bi_level_spg}.

\paragraph{Quadcopter Control Task.}
We consider a 2D quadrotor stabilization task implemented in \emph{safe-control-gym}\citep{yuan2022safe}, where the goal is to stabilize the quadrotor at an arbitrary reference point from random initial positions. Both the real-world and simulation environments are realized as separate instances of the same simulator. The simulation dynamics are parameterized through the quadrotor mass and length. The ground-truth parameters are $m_0 = 0.033\,\mathrm{kg}$ and $l_0 = 0.039\,\mathrm{m}$. In the simulator, the mass is initialized as $m = \theta_m m_0$ with $\theta_m = 0.5$, while the length is initialized at $l_0$ since it has little influence on performance. During bi-level \ac{ppo} training using Algorithm~\ref{alg:bi_level_ppo}, both mass and length parameters are adapted. The policy and critic are implemented using \acp{nn}. Further experimental details are provided in Appendix~\ref{app:quad}.

Fig.~\ref{fig:quad} shows that the in-sim policy stabilizes at a different location than the desired one when deployed on the real-world environment, whereas the policy trained with bi-level \ac{ppo} closes the sim-to-real gap. We did not tune hyperparameters, as this experiment only demonstrates the convergence behavior of bi-level \ac{ppo} on a simple task (see Appendix~\ref{app:quad}). Since the objective is stabilization to a reference, a small adjustment to the mass parameter is sufficient to achieve near-optimal performance. As a result, the estimated mass parameter converges to a value that differs from the ground truth while still yielding optimal performance.

\begin{figure*}[]
  \centering
  \setlength{\fboxsep}{0pt}
\vspace{-4pt}
  \begin{subfigure}[b]{0.40\textwidth}
    \resizebox{\linewidth}{!}{\input{figures/Discrete/ds_summary.pgf}}
    \label{fig:ds}
  \end{subfigure}
  \begin{subfigure}[b]{0.40\textwidth}
    \resizebox{\linewidth}{!}{\input{figures/LinearSys/ls_summary.pgf}}
    \label{fig:ls}
  \end{subfigure}
  \vspace{-16pt}
  \caption{Normalized return of 5 randomized trials for (left) discrete and (right) continuous \ac{mdp} examples.}
  \label{fig:results}
\end{figure*}

\section{Discussions}\label{sec:discussions}
%not an algorithm 

%% WHy dont we just fine tune the policy instead
Although Theorem~\ref{Th:bi_level_spg} assumes that the in-sim policy needs to fully converge to a local optimum for applying the \ac{ift}, in practice, a partially converged in-sim policy only introduces an additional perturbation to the already noisy real-world gradient driving the outer-level \ac{rl}. The three-time-scale analysis in Appendix~\ref{appendix:convergence_analysis} formally justifies this; bi-level \ac{rl} convergence does not require the in-sim policy to fully converge for each update to the simulation parameters, but rather requires \emph{timescale separation}: the critic and its sensitivities must stabilize on the fastest timescale, the in-sim policy must evolve on a faster timescale than the simulation parameters, and the outer simulator updates must be sufficiently slow. This is empirically validated in the quadcopter control task. Importantly, the regime in which in-sim convergence matters most is when the outer-level approaches a local optimum and the true gradient norm becomes small; however, in that same regime the outer updates are naturally small, which makes the moving target $\phi^\star(\theta)$ change slowly and allows the in-sim policy to track it with relatively few inner updates (see Appendix~\ref{appendix:convergence_analysis}).

\begin{figure*}[]
\centering
\setlength{\fboxsep}{0pt}
\vspace{2pt}
\resizebox{0.85\linewidth}{!}{\input{figures/quadcopter/quadcopter.pgf}}
\vspace{-4pt}
\caption{Quadcopter example. (Left) The gap between the optimal return and the performance of the in-sim policy illustrates the sim-to-real gap. Bi-level \ac{ppo} and direct policy fine-tuning in the real world reduce this gap compared to training the policy purely in simulation. (Middle) Trajectory tracking performance of the different policies. (Right) Evolution of the simulation parameters during bi-level \ac{ppo} training.}
\label{fig:quad}
\end{figure*}

 The bi-level \ac{rl} approach is applicable to any in-sim policy optimization scheme, provided one can estimate the sensitivity of the in-sim policy with respect to simulation parameters. Notable examples include bi-level \ac{rl} schemes for \ac{mpc}~\citep{reiter2025synthesis, kordabad2023reinforcement},  and \ac{cem}~\citep{amos_differentiable_2020, difftori}.  Consequently, the key contribution of the framework presented in this paper is the sensitivity analysis of the \ac{spg}-based \ac{rl} used to train the in-sim policy. This extends the previous work of \citep{nikishin2022control} that assumed that the policy is derived from a Q-function via a softmax, allowing the derivation of a model gradient by implicit differentiation of the Bellman equations. The bi-level \ac{rl} approach is directly applicable to Dyna-style \ac{mbrl} settings, where the dynamics model and reward functions, which are typically learned for prediction accuracy, can be adapted by an outer-level \ac{rl} loop to improve the real-world performance. 
 
 As demonstrated with \ac{ppo}, the sensitivity analysis in Theorem~\ref{Th:bi_level_spg} could be incorporated into most standard \ac{rl} algorithms, such as \ac{sac}~\citep{haarnoja2018soft}, by differentiating a few extra terms in their policy gradient equations without significant additional complexity. One could also deploy separate \ac{rl} algorithms in the inner and outer levels. Additionally, both approaches for estimating the critic sensitivities, in Lemma~\ref{lem:policy_eval_sensitivity} and Lemma~\ref{lem:policy_eval_sensitivity_QV_version}, are equally applicable, as we show their equivalence in Appendix~\ref{appendix:sensitivities-equivalency}. However, we favor the unrolled formulation in Lemma~\ref{lem:policy_eval_sensitivity} due to its lower computational complexity and its better suitability for parallelization, making it the more practical choice in our setting.

In sim-to-real \ac{rl}, an alternative to adapting the simulator is to fine-tune an in-sim policy directly on real-world data. While this can be effective in some cases, fine-tuning the simulator with bi-level \ac{rl} offers a more general and flexible approach. By fine-tuning the simulation parameters for sim-to-real optimality, we can exploit a full suite of \ac{rl} techniques in simulation, including domain randomization.  Moreover, learning an optimal simulation model provides valuable insights into which scenarios drive decision-making in practice, thereby enhancing explainability. Additionally, the simulator parameter space can often be kept low-dimensional, as practitioners typically adjust only a small subset of physically meaningful parameters. Fine-tuning this small set of parameters can be computationally more efficient and stable than fine-tuning very large policy networks. Moreover, fine-tuning the in-sim reward parameters provides extra flexibility, considering that sim-to-real optimality can be achieved by adapting only the reward parameters while keeping the simulation model parameters fixed~\citep{gros2019data}. Since the simulation model parameters, typically obtained via classical model‐fitting approaches~\citep{ljung1995system}, approximate the real-world dynamics with reasonable accuracy, the optimal simulation parameters likely lie within a neighborhood of those estimates. Consequently, in practice, the outer-level \ac{rl} can be implemented as a local search on the simulation parameter space~\citep{grudic2000localizing} similar to~\citep{jia_model_2024}.

\subsection*{Limitations}\label{sec:limitations}
%% Not an algorithms
%% Computationaly expensice due to critic sensitivities
 The practical deployment of the bi-level \ac{rl} framework requires the development of scalable bi-level \ac{rl} algorithms based on the results of Section~\ref{sec:bi_level_rl}. Given the challenges inherent in real-world online learning, offline-\ac{rl} is a suitable framework for the outer‐level \ac{rl}, which is not addressed in this work. The proposed bi-level \ac{rl} framework only provides local, not global, sim-to-real optimality, as is typical of gradient-based methods in contrast to global optimization approaches such as Bayesian optimization or evolutionary algorithms. Additionally, sim-to-real optimality is limited by model specification: if the simulator class is not sufficiently expressive, the desired optimality may not be achievable.

Moreover, the proposed bi-level \ac{rl} approach can be computationally heavy, due to the estimation of the critic sensitivities~\eqref{eq:critic_sensitivity_theta}. While the VJP formulation of the outer-level gradient described in 
Section~\ref{sec:convergence_and_implementation} avoids the computational 
complexity of explicitly forming $\nabla_\theta \phi$, parallelization is still required to make the framework scalable to larger problems.
 % Since, in practice, these sensitivities need to be approximated using \acp{dnn}, the computational burden increases significantly with the size of in-sim policy parameters.

While bi-level \ac{rl} accounts for the aleatoric uncertainties in the real-world environment, it does not account for epistemic uncertainties.  However, in real-world decision-making problems, epistemic uncertainty can significantly impact the value function estimates, leading to issues such as overestimation. To address this, epistemic uncertainties must be managed using probabilistic models of dynamics, or by penalizing the in-sim selection of out-of-distribution state-action pairs~\citep{kidambi2020morel}. 

%notion of state
One key assumption in our framework is that the simulator and the real system share the same state representation. While this assumption simplifies the analysis, real-world environments often provide only partial or noisy observations (e.g., images or sensor readings). Therefore, extending the bi‐level \ac{rl} to work directly with the observation space is an interesting direction.

%% differentiablilty
The bi‐level \ac{rl} framework requires regularity conditions on the dynamics model and reward functions (Assumption~\ref{ass:regularity}). Although this may seem limiting, differentiable simulators are rapidly becoming standard, including domains with discrete state or action spaces and applications such as contact robotics, where dynamics are inherently discontinuous~\citep{newbury2024review}. In the case of \ac{mbrl}, when both dynamics and reward functions are parameterized by \acp{nn}, they are inherently differentiable.

A limitation of the proposed sensitivity analysis is the local regularity assumption (Assumption~\ref{ass:local_regularity}) that \(\nabla_\phi \hat{\varphi}(\phi,\theta)\) is nonsingular. In deep \ac{rl}, this Jacobian may become nearly singular due to overparameterization, redundant parameterizations, saturated policies, flat local objectives, or inaccurate critic estimates, even when the in-sim policy has locally converged. In such cases, the sensitivity \(\nabla_\theta \phi\) can become poorly conditioned, leading to large and numerically unstable outer-level updates. Practical implementations should therefore regularize or damp the linear solve, use conservative step sizes or trust-region updates, and monitor the conditioning of \(\nabla_\phi \hat{\varphi}(\phi,\theta)\).

\section{Conclusions}\label{sec:conclusions}

We derived the sensitivity of locally converged \ac{rl} policies trained with \ac{spg} methods in an actor–critic setting with respect to simulation parameters. Based on this sensitivity analysis, we formulated a bi-level \ac{rl} approach that addresses the objective mismatch problem by adapting simulation parameters using gradients of real-world policy performance, thereby directly coupling the simulation model adaptation objective with the policy performance objective. We further provided a thorough convergence analysis of the proposed bi-level \ac{rl} approach, along with a proof-of-concept bi-level \ac{ppo} algorithm. We also discussed practical considerations for implementing this approach, including jointly updating simulation and policy parameters and using VJP for computational efficiency. While this paper establishes the theoretical foundations and mathematical tools for bi-level \ac{rl}, further algorithmic developments are necessary to scale the approach to large-scale \ac{rl} problems. Although motivated by achieving sim-to-real optimality in \ac{rl}, this work may be useful wherever sensitivity properties of \ac{spg} methods are of interest.

\bibliographystyle{plainnat}
\bibliography{mbrl} 

\newpage
\appendix
\section{Proof of Theorem 1, Lemma 1, and Lemma 2}\label{appendix:main_proof}
Here we provide the proof for lemma \ref{lem:mc_sensitivity}, lemma \ref{lem:policy_eval_sensitivity}, and theorem \ref{Th:bi_level_spg}.
We consider the stochastic policy gradient:
\begin{align}
\label{eq:sim:gradient}
\hat \varphi\left(\phi, \theta\right) = \mathbb E_{s\sim \hat\rho_{\theta}^{\pi_{\phi}}}\left[\nabla_{\phi}\log\pi_{\phi} Q_{\theta}^{\pi_{\phi}}\right] 
\end{align}
where $\phi$ are the policy parameters, $\theta$ are the parameters attached to the model and cost in the simulation-based training environment, and $Q_{\theta}^{\pi_{\phi}}$ is the critic of policy $\pi_{\phi}$ \textit{as seen} by the simulations. Distribution $ \hat\rho_{\theta}^{\pi_{\phi}}$ is the simulated ``discounted Markov chain density" associated to policy $\pi_{\phi}$. Assuming that the in-sim policy training sets the policy parameters $\phi$ such that \eqref{eq:sim:gradient} is zero, $\phi$ becomes an implicit function of $\theta$. 

The real-world (outer-level) policy gradient reads as:
\begin{align}
\varphi\left(\phi, \theta\right) &= \mathbb E_{s\sim \rho^{\pi_{\phi}}}\left[\nabla_{\theta}\log\pi_{\phi} Q^{\pi_{\phi}}\right] 
\end{align}
where $Q^{\pi_{\phi}}$ is the critic of policy $\pi_{\phi}$ \textit{as seen} by the real world, and $\rho^{\pi_{\phi}}$ is the ``discounted Markov chain density'' resulting from using policy $\pi_{\phi}$ in the real world. Here, $\theta$ implicitly affects $\log\pi_\phi$, as any changes in $\theta$ would change the local optima for the inner-level resulting in a different policy $\pi_\phi$, hence, we have,
\begin{align}
    \varphi\left(\phi, \theta\right) &= \mathbb E_{s\sim \rho^{\pi_{\phi}}}\left[\nabla_{\theta}\phi\nabla_{\phi}\log\pi_{\phi} Q^{\pi_{\phi}}\right]
\end{align}

The key element to compute the world policy gradient is to find the sensitivity $\nabla_{\theta}\phi$ using $\hat \varphi\left(\phi, \theta\right) =0$. The IFT readily applies (under some conditions), and
\begin{align}\label{eq:ift_spg}
\nabla_\phi \hat \varphi \nabla_\theta \phi + \nabla_\theta \hat \varphi &= 0 \\
\nabla_\theta \phi &= - \nabla_\phi \hat \varphi^{-1} \nabla_\theta \hat \varphi
\end{align}
Sensitivities $\nabla_\theta \hat \varphi,\,\nabla_\phi \hat \varphi$ can be computed as:
% \begin{subequations}
% \begin{align}
% \nabla_\theta \hat \varphi &= \mathbb E_{s\sim \hat\rho_{\theta}^{\pi_{\phi}}}\left[\nabla_{\phi}\log\pi_{\phi}  \nabla_\theta Q_{\theta}^{\pi_{\phi}}\right] \label{eq:sens:theta} \\
% \nabla_\phi \hat \varphi &=\mathbb E_{s\sim \hat\rho_{\theta}^{\pi_{\phi}}}\left[\nabla_{\phi\phi}\log\pi_{\phi} Q_{\theta}^{\pi_{\phi}}+ \nabla_{\phi}\log\pi_{\phi} \nabla_\phi Q_{\theta}^{\pi_{\phi}} \right] \label{eq:sens:phi} 
% \end{align}
% \end{subequations}
\begin{subequations}
\begin{align}
\nabla_\theta \hat \varphi &= \mathbb E_{s\sim \hat\rho_{\theta}^{\pi_{\phi}}}\left[\nabla_{\phi}\log\pi_{\phi} \nabla_\theta Q_{\theta}^{\pi_{\phi}}\right] + \nabla_\theta\mathbb E_{s\sim \hat\rho_{\theta}^{\pi_{\phi}}}\left[\nabla_{\phi}\log\pi_{\phi}\, Q_{\theta}^{\pi_{\phi}}\right] \label{eq:sens:theta} \\
\nabla_\phi \hat \varphi &=\mathbb E_{s\sim \hat\rho_{\theta}^{\pi_{\phi}}}\left[\nabla_{\phi\phi}\log\pi_{\phi} Q_{\theta}^{\pi_{\phi}}+ \nabla_{\phi}\log\pi_{\phi} \nabla_\phi Q_{\theta}^{\pi_{\phi}} \right] + \nabla_\phi\mathbb E_{s\sim \hat\rho_{\theta}^{\pi_{\phi}}}\left[\nabla_{\phi}\log\pi_{\phi}\, Q_{\theta}^{\pi_{\phi}}\right] \label{eq:sens:phi} 
\end{align}
\end{subequations}
where:
\begin{itemize}
\item $\nabla_{\phi\phi}\log\pi_{\phi}$ is an explicit expression from the policy function
\item $\nabla_\theta Q_{\theta}^{\pi_{\phi}}$ is the influence of changing the cost and model on the simulation-based critic \textit{under a fixed policy}.
\item $\nabla_\phi Q_{\theta}^{\pi_{\phi}}$ is the influence of changing the policy parameters on the simulation-based critic \textit{under a fixed cost and model}.
\item \( \nabla_\phi\mathbb E_{s\sim \hat\rho_{\theta}^{\pi_{\phi}}}\left[ \cdot \right]\) is the effect of changing $\phi$ on the simulation-based distribution $ \hat\rho_{\theta}^{\pi_{\phi}}$, and the influence of that change on the expected value \textit{under a fixed cost and model}.
\item \( \nabla_\theta\mathbb E_{s\sim \hat\rho_{\theta}^{\pi_{\phi}}}\left[ \cdot \right]\) is the effect of changing $\theta$ on the simulation-based distribution $ \hat\rho_{\theta}^{\pi_{\phi}}$, and the influence of that change on the expected value \textit{under a fixed policy}.
\end{itemize}

From \eqref{eq:ift_spg}:
% \begin{equation}
% \frac{\partial \phi}{\partial \theta}
% \;=\;
% -\,\biggl(\frac{\partial \hat{\varphi}\bigl(\phi,\theta\bigr)}{\partial \phi}\biggr)^{-1}
% \;\cdot\;
% \frac{\partial \hat{\varphi}\bigl(\phi,\theta\bigr)}{\partial \theta}
% \,.
% \end{equation}
\begin{align}
\nabla_\theta \phi 
&= -\Bigl(
   \underbrace{\mathbb E_{s\sim \hat\rho_{\theta}^{\pi_{\phi}}}
     \bigl[\nabla_{\phi\phi}\log\pi_{\phi}\,Q_{\theta}^{\pi_{\phi}}
          + \nabla_{\phi}\log\pi_{\phi}\,\nabla_{\phi}Q_{\theta}^{\pi_{\phi}}\bigr]}
   _{\substack{\mathrm{Gradient \,of\, } \hat{\varphi}\mathrm{\, w.r.t.\, }\phi}}
   \;+\;
   \underbrace{\nabla_{\phi}\mathbb E_{s\sim \hat\rho_{\theta}^{\pi_{\phi}}}
     \bigl[\nabla_{\phi}\log\pi_{\phi}\,Q_{\theta}^{\pi_{\phi}}\bigr]}
   _{\substack{\mathrm{Gradient \,of \,dist.\, }\hat\rho_{\theta}^{\pi_{\phi}}\mathrm{\,w.r.t.\, }\phi}}
\Bigr)^{-1}
\nonumber\\
&\qquad\qquad
\Bigl(
   \underbrace{\mathbb E_{s\sim \hat\rho_{\theta}^{\pi_{\phi}}}
     \bigl[\nabla_{\phi}\log\pi_{\phi}\,\nabla_{\theta}Q_{\theta}^{\pi_{\phi}}\bigr]}
   _{\substack{\mathrm{Gradient\, of\, } \hat{\varphi}\mathrm{ \,w.r.t.\, }\theta}}
   \;+\;
   \underbrace{\nabla_{\theta}\mathbb E_{s\sim \hat\rho_{\theta}^{\pi_{\phi}}}
     \bigl[\nabla_{\phi}\log\pi_{\phi}\,Q_{\theta}^{\pi_{\phi}}\bigr]}
   _{\substack{\mathrm{Gradient\, of\, dist.\, }\hat\rho_{\theta}^{\pi_{\phi}}\mathrm{\,w.r.t.\, }\theta}}
\Bigr).
\end{align}

\subsection*{Explanation of Theorem 1}
The Theorem~\ref{Th:bi_level_spg} defines how small perturbations in the simulator parameters $\theta$ propagate through the \ac{rl} process to affect the resulting policy parameters $\phi$. The inverse term in the first line of~\eqref{eq:nabla_theta_phi} captures how changes in the policy gradient translate into changes in the stationary policy parameters. This term corresponds to the Jacobian of the policy gradient with respect to the policy parameters. The first expectation inside this term represents the local curvature of the policy objective: the term $\nabla_{\phi\phi}\log\pi_\phi\,Q_\theta^{\pi_\phi}$ captures the direct curvature of the policy log-likelihood weighted by the critic, while the term $\nabla_\phi\log\pi_\phi\,\nabla_\phi Q_\theta^{\pi_\phi}$ accounts for how changes in the policy parameters modify the critic estimates. The second component of this line,
$\nabla_{\phi}\mathbb{E}_{s\sim\hat{\rho}_\theta^{\pi_\phi}}[\cdot]$, captures the additional effect that changing the policy parameters alters the state distribution induced by the policy in the simulator, which in turn modifies the policy gradient.

The second line of~\eqref{eq:nabla_theta_phi} describes how perturbations in the simulator parameters change the policy gradient itself. The first expectation term captures the effect of simulator parameters on the critic $Q_\theta^{\pi_\phi}$ through the simulator dynamics and reward model. The second term,
$\nabla_{\theta}\mathbb{E}_{s\sim\hat{\rho}_\theta^{\pi_\phi}}[\cdot]$, captures how changes in simulator parameters modify the state distribution generated by the simulator, which indirectly alters the policy gradient.

The four components together describe how perturbations in simulator parameters influence the learned policy. Perturbations in the simulator modify both the value estimates and the distribution of simulated trajectories, which alters the policy gradient. The inverse Jacobian then maps these changes in the policy gradient to the resulting shift in the locally optimal policy parameters.

With this, we next unpack the effect of changing $\phi, \theta$ on the simulation-based distribution $ \hat\rho_{\theta}^{\pi_{\phi}}$, and the influence of that change on the expected value. In order to develop that aspect, let us develop the notion of sensitivity over Markov Chains.

\subsection{Proof of Lemma 1: Sensitivity over Markov Chains}\label{appendix:proof_lemma_markov_chain}
For compactness, let us consider a generic function $\eta\left(s, a\right)$ of the state and action, and its expected value at a specific time stage $k$:
\begin{align}
&\mathbb E_{s\sim \hat\rho_{\theta,k}^{\pi_{\phi}}}\left[\eta\left(s_k,a_k\right)\right] = \int   \eta\left(s_k,a_k\right)  \int \rho^0\left(s_0\right) \,\pi_\phi(a_k|s_k) \prod_{i=0}^{k-1} \mathcal{P}^{\pi_\phi}_{\theta}\left(s_{i+1},s_{i},a_{i}\right)   \mathrm ds_i\mathrm da_i \mathrm ds_k 
\end{align}

where
\begin{align}
\mathcal{P}^{\pi_\phi}_{\theta}\left(s_{i+1},s_{i},a_{i}\right) = {\mathcal P}_\theta\left(s_{i+1}|s_{i},a_{i}\right)\pi_{\phi}\left(a_{i}|s_{i}\right)
\end{align}
where ${\mathcal P}_\theta$ is the simulation model (assumed stochastic here), and $\rho^0$ the initial condition distribution, assumed fixed here. Differentiating this expected value w.r.t. $\phi, \theta$ (assuming Leibniz rule) then reads as (arguments omitted for simplicity):

\begin{align}
\label{eq:MC:sens}
\partial\mathbb E_{s\sim \hat\rho_{\theta,k}^{\pi_{\phi}}} &\left[\eta\left(s_k,a_k\right)\right] \\
&=\int  \eta   \int \rho^0 \partial\left(\pi_{\phi}\left(a_{k}|s_{k}\right) \prod_{i=0}^{k-1}  \mathcal{P}^{\pi_\phi}_{\theta}\right)\mathrm ds_i\mathrm da_i \mathrm ds_k \nonumber \\
&= \int  \eta   \int \rho^0 \partial \log\left(\pi_{\phi}\left(a_{k}|s_{k}\right) \prod_{i=0}^{k-1} \mathcal{P}^{\pi_\phi}_{\theta}\right) \pi_{\phi}\left(a_{k}|s_{k}\right)  \prod_{i=0}^{k-1} \mathcal{P}^{\pi_\phi}_{\theta}\mathrm ds_i\mathrm da_i \mathrm ds_k \nonumber\\
&= \int  \eta   \int \rho^0 \left( \partial \log\pi_{\phi}\left(a_{k}|s_{k}\right)  + \sum_{i=0}^{k-1} \partial \log \mathcal{P}^{\pi_\phi}_{\theta}\right) \pi_{\phi}\left(a_{k}|s_{k}\right)  \prod_{i=0}^{k-1} \mathcal{P}^{\pi_\phi}_{\theta}\mathrm ds_i\mathrm da_i \mathrm ds_k \nonumber
\end{align}
Which has the expected value form:
\begin{align}\label{eq:MC:sens2}
\partial \, \mathbb E_{s\sim \hat\rho_{\theta,k}^{\pi_{\phi}}} \bigl[\eta(s_k,a_k)\bigr] = \mathbb{E}_{\substack{s_0\sim\rho^0,\,a_i\sim\pi_\phi,\\s_{i+1}\sim \mathcal{P}^{\pi_\phi}_{\theta}}}
   \Bigl[\,
     \eta(s_k,a_k)\,\Bigl(
        \partial\log\pi_\phi(a_k\!\mid\!s_k)
      + \sum_{i=0}^{k-1}\partial\log \mathcal{P}^{\pi_\phi}_\theta
     \Bigr)
   \Bigr]\,.
\end{align}

\subsection*{Computational aspect}\label{sec:senscomputation}
In practice, \eqref{eq:MC:sens2} will be evaluated from a model sampling. Let us label 
\begin{align}
s_0^j,a_0^j,\ldots,s_N^j,a_N^j
\end{align}
the sampled trajectory $j\in\left\{1,\ldots,n\right\}$ of length $N$. The expected value approximation would then read:
\begin{align}
\mathbb E_{s\sim \hat\rho_{\theta,k}^{\pi_{\phi}}}\left[\eta\left(s_k,a_k\right)\right] \approx \frac{1}{n}\sum_{j=1}^n \eta\left(s_k^j,a_k^j\right)
\end{align}
Its sensitivities would then read as:
\begin{align}
\partial \mathbb E_{s\sim \hat\rho_{\theta,k}^{\pi_{\phi}}}\left[\eta\left(s_k,a_k\right)\right]  \approx \frac{1}{n}\sum_{j=1}^n \eta\left(s_k^j,a_k^j\right) \otimes \left( W^j_k +  \partial \log\pi_{\phi}\left(\left.a_k^j\right|s_k^j\right)  \right) 
\end{align}
where
\begin{align}
W^j_k = \sum_{i=0}^{k-1} \partial \log \mathcal{P}^{\pi_\phi}_{\theta}\left(\left.s^j_{i+1}\right|s^j_{i},a^j_{i}\right)
\end{align}
Note that $W^j_k$ can be carried forward as a dynamic alongside the model trajectory, i.e.
\begin{align}
W^j_k = W^j_{k-1} +  \partial \log \mathcal{P}^{\pi_\phi}_{\theta}\left(\left.s^j_{k}\right | s^j_{k-1},a^j_{k-1}\right),\quad 
W^j_0 = 0  
\end{align}

\subsection*{Expected value over full Markov Chain} 
For policy gradient methods, we need to compute objects in the form
\begin{align}
\mathbb E\left[\sum_{k=0}^\infty \eta\left(s_k,a_k\right)\right] \approx \frac{1}{nN}\sum_{k=0}^N \sum_{j=1}^n \eta\left(s_k^j,a_k^j\right)
\end{align}
and process their sensitivities. There is no significant difference from Sec. \ref{sec:senscomputation}, i.e. the sensitivities read as:
\begin{align}
\label{eq:MCSens}
\partial \mathbb E\left[\sum_{k=0}^\infty \eta\left(s_k,a_k\right)\right] \approx \frac{1}{nN}\sum_{k=0}^N \sum_{j=1}^n \eta\left(s_k^j,a_k^j\right) \otimes \left( W^j_k +  \partial \log\pi_{\phi}\left(\left.a_k^j\right|s_k^j\right)  \right)  
\end{align}
Efforts can be put in computing \eqref{eq:MCSens} in the most efficient way in terms of flops and memory requirements, but that may be best left for other people to figure out.

\subsection*{Sensitivity for the policy gradient} 
We can then provide computational expressions for the policy gradient. We observe that
\begin{align}
\nabla_\phi\mathbb E_{s\sim \hat\rho_{\theta}^{\pi_{\phi}}}\left[\nabla_{\phi}\log\pi_{\phi}\, Q_{\theta}^{\pi_{\phi}}\right] \approx
 \frac{1}{nN}\sum_{k=0}^N \sum_{j=1}^n &\nabla_{\phi}\log\pi_{\phi}\left(s_k^j\right)\, Q_{\theta}^{\pi_{\phi}}\left(s_k^j,a_k^j\right) \notag\\
&\otimes \left( W_{\phi, k}^j +  \partial \log\pi_{\phi}\left(\left.a_k^j\right|s_k^j\right)  \right) 
\end{align}
where $W^j_{\phi, 0} = 0$ and 
\begin{align}
W_{\phi, k}^j  &= W_{\phi, k-1}^j  +  \nabla_\phi\log \mathcal{P}^{\pi_\phi}_{\theta}\left(s^j_{k},s^j_{k-1},a^j_{k-1}\right)  \\
&= W_{\phi, k-1}^j + \nabla_\phi\log \pi_{\phi}\left(\left.a^j_{k-1}\right |s^j_{k-1}\right)\nonumber
\end{align}
Similarly
\begin{align}
\nabla_\theta\mathbb E_{s\sim \hat\rho_{\theta}^{\pi_{\phi}}}\left[\nabla_{\phi}\log\pi_{\phi}\, Q_{\theta}^{\pi_{\phi}}\right] \approx
\frac{1}{nN}\sum_{k=0}^N \sum_{j=1}^n &\nabla_{\phi}\log\pi_{\phi}\left(s_k^j\right)\, Q_{\theta}^{\pi_{\phi}}\left(s_k^j,a_k^j\right) \notag\\
&\otimes \left( W_{\theta, k}^j +  \partial \log\pi_{\phi}\left(\left.a_k^j\right|s_k^j\right)  \right) 
\end{align}
where $W^j_{\theta, 0} = 0$ and 
\begin{align}
W_{\theta, k}^j  &= W_{\theta, k-1}^j  +  \nabla_\theta\log \mathcal{P}^{\pi_\phi}_{\theta}\left(s^j_{k},s^j_{k-1},a^j_{k-1}\right)  \\
&= W_{\theta, k-1}^j  +  \nabla_\theta\log  {\mathcal P}_\theta\left(\left.s^j_{k}\right|s^j_{k-1},a^j_{k-1}\right)\nonumber
\end{align}
These terms ought to be added to \eqref{eq:sens:phi}, \eqref{eq:sens:theta} respectively. In implementations, these terms would simply weigh the contributions in forming the policy gradient sensitivities.

\subsection{Proof of Lemma 2: Sensitivities of Policy Evaluation}\label{appendix:critic_sensitivity}

\subsubsection{Operator-theoretic preliminaries}
\label{sec:prelim-operators}

We work on the state–action space \(\mathcal S\times\mathcal A\).
For the \emph{in-sim} fast layer, write the transition kernel and reward as
\({\mathcal P}_\theta(\mathrm d s'\mid s,a)\) and \(R_\theta(s,a)\).
Let \(\pi_\phi(\mathrm d a\mid s)\) be the policy. The discount factor is given by \(\gamma\in(0,1)\). Let \(\mathcal B_\infty\) be the Banach space of bounded measurable
\(f:\mathcal S\times\mathcal A\to\mathbb R\) with sup-norm
\(\|f\|_\infty=\sup_{(s,a)}|f(s,a)|\).

To interpret likelihood-ratio / score terms such as \(\nabla_\phi\log \pi_\phi(a\mid s)\) and
\(\nabla_\theta\log {\mathcal P}_\theta(s'\mid s,a)\), we assume there exist fixed reference measures
\(\lambda_{\mathcal A}\) on \(\mathcal A\) and \(\lambda_{\mathcal S}\) on \(\mathcal S\) such that
\(\pi_\phi(\cdot\mid s)\ll \lambda_{\mathcal A}\) for all \(s\) and
\({\mathcal P}_\theta(\cdot\mid s,a)\ll \lambda_{\mathcal S}\) for all \((s,a)\).
We denote the corresponding densities by \(\pi_\phi(a\mid s)\) and
\({\mathcal P}_\theta(s'\mid s,a)\), so the above log/score expressions are well-defined.

% For critic sensitivity estimation, in addition to the standard boundedness and differentiability assumptions, we assume \( Q^{\pi_\phi}_\theta\) has a fixed point. 

\paragraph{Markov and Bellman operators.}
Given any bounded function \(f:\mathcal S\times\mathcal A\to\mathbb R\), the
policy–induced Markov operator \(P^{\pi_\phi}_\theta:\mathcal B_\infty\to\mathcal B_\infty\) (is the “one-step look-ahead” operator for the Markov chain over state–action pairs), maps \(f\) to a new function as
\begin{equation}
\label{eq:Ppi-def}
(P^{\pi_\phi}_\theta f)(s,a)
\;:=\;
\int_{\mathcal S}\!\!\int_{\mathcal A}
f(s',a')\,\pi_\phi(\mathrm da'\mid s')\,{\mathcal P}_\theta(\mathrm ds'\mid s,a)
\;=\;
\mathbb E\!\big[f(s',a')\mid s_0\!=\!s,\, a_0\!=\!a\big].
\end{equation}
Then \(\|P^{\pi_\phi}_\theta f\|_\infty\le \|f\|_\infty\).
The (discounted) Bellman operator for action values is
\[
(\hat T_{\theta,\phi} Q_\theta)\;=\;R_\theta+\gamma\,(P^{\pi_\phi}_\theta Q_\theta).
\]
\(\hat T_{\theta,\phi}\) is a \(\gamma\)-contraction on \((\mathcal B_\infty,\|\cdot\|_\infty)\),
hence the inverse of $(I-\gamma P^{\pi_\phi}_\theta)$ exists and admits a unique fixed point
\(Q^{\pi_\phi}_\theta\in\mathcal B_\infty\) solving
\((I-\gamma P^{\pi_\phi}_\theta)Q^{\pi_\phi}_\theta=R_\theta\).~\citep{puterman2014markov}.

\paragraph{Poisson equations for sensitivities }
Under the policy $\pi=\pi_{\phi(\theta)}$, the critic $Q^{\pi_{\phi(\theta)}}_\theta$ solves the Bellman fixed point equation:
\begin{equation}\label{eq:fixed_point_cirtic}
    Q^{\pi_\phi}_\theta \;=\; R_\theta \;+\; \gamma\,P^{\pi_{\phi(\theta)}}_\theta\,Q^{\pi_\phi}_\theta.
\end{equation}

For a parameter $\eta\in\{\phi,\theta\}$, $\nabla_\eta Q^{\pi_\phi}_\theta$ can be derived by differentiating the Bellman fixed point~\eqref{eq:fixed_point_cirtic}  with respect to $\eta$
\begin{equation}
\nabla_\eta Q^{\pi_\phi}_\theta
=
\nabla_\eta R_\theta
+
\gamma\, (\nabla_\eta P^{\pi_\phi}_\theta)\, Q^{\pi_\phi}_\theta
+
\gamma\, P^{\pi_\phi}_\theta \nabla_\eta Q^{\pi_\phi}_\theta .
\end{equation}

Rearranging terms gives the linear system
\begin{equation}
\big(I - \gamma P^{\pi_\phi}_\theta \big)
\nabla_\eta Q^{\pi_\phi}_\theta
=
b_\eta,
\label{eq:critic_sensitivity_linear}
\end{equation}
where
\begin{equation}
b_\eta
:=
\nabla_\eta R_\theta
+
\gamma\, (\nabla_\eta P^{\pi_\phi}_\theta)\, Q^{\pi_\phi}_\theta .
\end{equation}

With Neumann series $(I-\gamma P^{\pi_\phi}_\theta)^{-1}=\sum_{t\ge0}(\gamma P^{\pi_\phi}_\theta)^t$ 
which converges because $\gamma<1$, applying it to~\eqref{eq:critic_sensitivity_linear} yields
\begin{equation}
\nabla_\eta Q^{\pi_\phi}_\theta
=
\big(I - \gamma P^{\pi_\phi}_\theta\big)^{-1} b_\eta
=
\sum_{t=0}^{\infty}(\gamma P^{\pi_\phi}_\theta)^t\, b_\eta.
\end{equation}
Now, the following Lemma provides the sensitivity of the Markov operator.
\begin{lemma}\label{lem:markov_operator_sensitivity}
     For any bounded $f$, the sensitivity of the Markov operator w.r.t. \(\theta\) and \(\phi\) are given by:
\begin{equation}
   \begin{aligned}
\nabla_\phi{P^{\pi_\phi}_\theta}\,f(s,a)
&= \mathbb{E}_{\substack{s'\sim {\mathcal P}_\theta(\cdot\,|\,s,a)\,\,a'\sim \pi_\phi(\cdot\,|\,s')}}\!\left[\,\nabla_\phi\log\pi_\phi(a'|s')\, f(s',a') \,\middle|\, s,a\right],\\
\nabla_\theta{P^{\pi_\phi}_\theta}\,f(s,a)
&= \mathbb{E}_{\substack{s'\sim {\mathcal P}_\theta(\cdot\,|\,s,a)\,\,a'\sim \pi_\phi(\cdot\,|\,s')}}\!\left[\,\nabla_\theta\log {\mathcal P}_\theta(s'|s,a)\,  f(s', a') \,\middle|\, s,a\right]\,,
\end{aligned} 
\end{equation}
respectively.
\end{lemma}
\begin{proof}
For any bounded $f$ and fixed $(s,a)$,
\begin{align}
(P^{\pi_\phi}_\theta f)(s,a)
&= \int_{\mathcal S}\!\!\int_{\mathcal A}
    f(s',a')\,\pi_\phi(a'\mid s')\,\mathcal P_\theta(s' \mid s,a)\, da'\, ds' \nonumber\\
&=: \int_{\mathcal S} g_\phi(s')\,\mathcal P_\theta(s' \mid s,a)\, ds',
\qquad
g_\phi(s') := \int_{\mathcal A} f(s',a')\,\pi_\phi(a'\mid s')\, da'.
\label{eq:iterated}
\end{align}

Only $g_\phi(s')$ depends on $\phi$ in \eqref{eq:iterated}.
Differentiate under the inner integral (over $a'$):
\begin{align}
\nabla_\phi\,g_\phi(s')
&= \int_{\mathcal A} f(s',a')\,{\nabla_\phi}\,\pi_\phi(a'\mid s')\,
\mathrm da'\nonumber\\
&= \int_{\mathcal A} f(s',a')\,\pi_\phi(a'\mid s')\,
\nabla_\phi\log \pi_\phi(a'\mid s')\,
\mathrm da'\nonumber\\
&= \mathbb E_{a'\sim \pi_\phi(\cdot\mid s')}
\big[f(s',a')\,\nabla_\phi\log \pi_\phi(a'\mid s')\big].
\label{eq:d-gphi}
\end{align}
Now differentiate \eqref{eq:iterated} and use that ${\mathcal P}_\theta$ does not depend on $\phi$:
\begin{align}\label{eq:markov_oepraor_sensitivity_phi}
\nabla_\phi\,(P^{\pi_\phi}_\theta f)(s,a)
&= \int_{\mathcal S} \nabla_\phi\,g_\phi(s')\,
{\mathcal P}_\theta(s'\mid s,a)\,\mathrm ds'\nonumber\\
&= \int_{\mathcal S}
\mathbb E_{a'\sim \pi_\phi(\cdot\mid s')}
\big[f(s',a')\,\nabla_\phi\log \pi_\phi(a'\mid s')\big]\,
{\mathcal P}_\theta(s'\mid s,a)\,\mathrm ds'\nonumber\\
&= \mathbb E\!\left[\nabla_\phi\log \pi_\phi(a'\mid s')\,f(s',a')\ \middle|\ s,a \right].
\end{align}

In \eqref{eq:iterated}, only ${\mathcal P}_\theta(\cdot\mid s,a)$ depends on $\theta$ as we keep the policy fixed:
\begin{align}
{\nabla_\theta}\,(P^{\pi_\phi}_\theta f)(s,a)
&= \int_{\mathcal S} g_\phi(s')\,\nabla_{\theta}\,
{\mathcal P}_\theta(s'\mid s,a)\,\mathrm ds'.
\end{align}
Assuming ${\mathcal P}_\theta(s'\mid s,a)>0$ (a.e.) and differentiable in $\theta$,
\[
{\nabla_\theta}\,{\mathcal P}_\theta(s'\mid s,a)
= {\mathcal P}_\theta(s'\mid s,a)\,\nabla_\theta \log {\mathcal P}_\theta(s'\mid s,a).
\]
Therefore
\begin{align}\label{eq:markov_oepraor_sensitivity_theta}
{\nabla_\theta}\,(P^{\pi_\phi}_\theta f)(s,a)
&= \int_{\mathcal S} g_\phi(s')\,{\mathcal P}_\theta(s'\mid s,a)\,
\nabla_\theta \log {\mathcal P}_\theta(s'\mid s,a)\,\mathrm ds'\nonumber\\
&= \mathbb E_{s'\sim {\mathcal P}_\theta(\cdot\mid s,a)}
\big[\,g_\phi(s')\,\nabla_\theta \log {\mathcal P}_\theta(s'\mid s,a)\,\big]\nonumber\\
&= \mathbb E\!\left[\nabla_\theta\log {\mathcal P}_\theta(s'|s,a)\,f(s',a') \ \middle|\ s,a \right],
\end{align}
\end{proof}

\subsubsection{Critic Sensitivities based on fixed point}
\paragraph{(i) Derivative of Q w.r.t.\ $\phi$:}
Combining~\eqref{eq:critic_sensitivity_linear} and~\eqref{eq:markov_oepraor_sensitivity_phi} yields,
\begin{align}
\label{eq:Uphi-operator}
(I-\gamma P^{\pi_\phi}_\theta)\,\nabla_\phi  Q^{\pi_\phi}_\theta
&\;=\; \nabla_\phi{R_\theta}
\;+\; \gamma\,\mathbb{E}_{\substack{s'\sim {\mathcal P}_\theta(\cdot\,|\,s,a)\,\,a'\sim \pi_\phi(\cdot\,|\,s')}}\!\left[\nabla_\phi\log\pi_\phi(a'|s')\, Q^{\pi_\phi}_\theta(s',a') \mid s,a \right],\\
\label{eq:Uphi-series}
\nabla_\phi  Q^{\pi_\phi}_\theta
&\;=\; \sum_{t=0}^{\infty} (\gamma P^{\pi_\phi}_\theta)^t
\Big(\nabla_\phi{R_\theta}
+ \gamma\,\nabla_\phi{ P^{\pi_\phi}_\theta}\,Q^{\pi_\phi}_\theta\Big),\\
\label{eq:Uphi-trajectory}
\nabla_\phi  Q^{\pi_\phi}_\theta(s,a)
&\;=\; \mathbb E\!\left[\sum_{t=0}^{\infty} \gamma^{t+1}\,\nabla_\phi\log\pi_\phi(a_{t+1}\!\mid s_{t+1})\, Q^{\pi_\phi}_\theta(s_{t+1},a_{t+1})
\ \middle|\ s_0\!=\!s,a_0\!=\!a
\right].
\end{align}

\medskip\noindent
\paragraph{(ii) Derivative w.r.t.\ $\theta$:} Combining~\eqref{eq:critic_sensitivity_linear} and~\eqref{eq:markov_oepraor_sensitivity_theta}
\begin{align}
\label{eq:Utheta-operator}
(I-\gamma P^{\pi_\phi}_\theta)\,\nabla_\theta  Q^{\pi_\phi}_\theta
&\;=\; \nabla_\theta{R_\theta}
\;+\; \gamma\,\mathbb E\!\left[\nabla_\theta\log {\mathcal P}_\theta(s'|S,A)\, {V}^{\pi_\phi}_\theta(s') \mid s_0\!=\!s,\,a_0\!=\!a \right],\\
\label{eq:Utheta-series}
\nabla_\theta  Q^{\pi_\phi}_\theta
&\;=\; \sum_{t=0}^{\infty} (\gamma P^{\pi_\phi}_\theta)^t
\Big(\nabla_\theta{R_\theta}
+ \gamma\,\nabla_\theta{P^{\pi_\phi}_\theta}\,Q^{\pi_\phi}_\theta\Big),\\
\label{eq:Utheta-trajectory}
\nabla_\theta Q^{\pi_\phi}_\theta(s,a)
&= \mathbb{E}\Bigg[
\sum_{t=0}^{\infty} \gamma^t\,\nabla_\theta R_\theta(s_t,a_t) \,\,+ 
\\
&\qquad\quad\,\,
\sum_{t=0}^{\infty} \gamma^{t+1}\,
\nabla_\theta \log{\mathcal P}_\theta(s_{t+1}\mid s_t,a_t)\,
Q^{\pi_\phi}_\theta(s_{t+1},a_{t+1})
\;\Bigg|\; s_0=s, a_0=a
\Bigg].
\end{align}

If the simulator is differentiable, we can use the corresponding \ac{rp} term instead of the \ac{lr} terms. Assume independent noise sources 
$\epsilon_t \sim p(\epsilon)$ and $\xi_{t+1} \sim p(\xi)$, 
which do not depend on $(\theta,\phi)$. 
Then the simulator dynamics ${\mathcal P}_\theta$ 
and the policy $\pi_\phi$ admit the following 
reparameterized representations:
\begin{align}
\label{eq:rp_dynamics_mapping}
s_{t+1} &= {\mathcal P}_\theta(s_t,a_t,\epsilon_t), \\
\label{eq:rp_policy_mapping}
a_{t+1} &= \pi_\phi(s_{t+1},\xi_{t+1}),
\end{align}
where ${\mathcal P}_\theta$ and $g_\phi$ are deterministic, differentiable mappings. 
In this form, sampling from ${\mathcal P}_\theta$ or $\pi_\phi$ 
is expressed as pushing forward a parameter-free noise distribution 
through a parameterized transformation.
\begin{equation}
\label{eq:critic_sensitivity_theta_rp}
\begin{aligned}
\nabla_\theta Q^{\pi_\phi}_\theta(s,a)
&= \mathbb{E}\Bigg[
\sum_{t=0}^{\infty} \gamma^t \,\nabla_\theta R_\theta(s_t,a_t)
\\ &\qquad\qquad
+\;\sum_{t=0}^{\infty} \gamma^{t+1}\,
\nabla_{s_{t+1}} Q^{\pi_\phi}_\theta\!\big(s_{t+1},a_{t+1}\big)\,
\nabla_\theta {\mathcal P}_\theta(s_t,a_t,\epsilon_t)
\ \Bigm|\ s_0=s,\ a_0=a \Bigg],
\end{aligned}
\end{equation}

Where,
\begin{equation}
\label{eq:rp_state_chain}
\nabla_{s_{t+1}}Q^{\pi_\phi}_\theta(s_{t+1},a_{t+1})
=
\nabla_s Q^{\pi_\phi}_\theta(s_{t+1},a_{t+1})
\;+\;
\nabla_a Q^{\pi_\phi}_\theta(s_{t+1},a_{t+1})\,
\nabla_{s_{t+1}} \pi_\phi(s_{t+1},\xi_{t+1}).
\end{equation}

\begin{equation}
\label{eq:critic_sensitivity_phi_rp}
\nabla_\phi Q^{\pi_\phi}_\theta(s,a)
= \mathbb E\!\left[
\sum_{t=0}^{\infty} \gamma^{t+1}\,
\nabla_{a_{t+1}} Q^{\pi_\phi}_\theta\!\big(s_{t+1},a_{t+1}\big)\,
\nabla_\phi \pi_\phi(s_{t+1},\xi_{t+1})
\ \middle|\ s_0=s,\ a_0=a
\right].
\end{equation}

\subsection{Alternative Sensitivity estimation for the policy evaluation}\label{appendix:alternative_sensitivity}
\begin{lemma}[Sensitivities of \(Q^{\pi}\) and \(V^{\pi}\)]
\label{lem:policy_eval_sensitivity_QV_version}
Given the Bellman conditions,
\begin{align*}
Q^{\pi}(s,a) &= R(s,a) + \gamma\mathbb E_{s_+ \sim f}[V^{\pi}(s_+)], \quad \text{and}\quad 
V^{\pi}(s) = \mathbb E_{a \sim \pi}[Q^{\pi}(s,a)],
\end{align*}
the sensitivity of \(Q^{\pi}\) and \(V^{\pi}\) are given by:
\begin{subequations}
\begin{align}
\partial Q^{\pi}(s,a)&= \partial R(s,a)+\gamma\mathbb E_{s_+ \sim f}\left[\partial V^{\pi}(s_+) + V^{\pi}(s_+)\partial\log f(s_+|s,a)\right],\\
\partial V^{\pi}(s)&=\mathbb E_{a\sim\pi}\left[^{\pi}(s,a)+Q^{\pi}(s,a)\partial\log\pi(a|s)\right].
\end{align}
\end{subequations}
%The proof is provided in Appendix \ref{}.
\end{lemma}

\begin{proof}
The simulation-based critic ideally satisfies the following equations:
\begin{subequations}
\label{eq:policy_evalution}
\begin{align}
Q_{\theta}^{\pi_{\phi}}\left(s,a\right) &= R_{\theta}\left(s,a\right) + \gamma\mathbb E_{s_+\sim {\mathcal P}_\theta}\left[\left.V_{\theta}^{\pi_{\phi}}\left(s_+\right)\,\right|\,s,a \right] \\
V_{\theta}^{\pi_{\phi}}\left(s\right) &= \mathbb E_{a\sim \pi_{\phi}}\left[Q_{\theta}^{\pi_{\phi}}\left(s,a\right) \right]
\end{align}
\end{subequations}
To compute the sensitivities ``safely", we observe that \eqref{eq:policy_evalution} in full reads as:
\begin{subequations}
\label{eq:policy_evalution:full}
\begin{align}
Q_{\theta}^{\pi_{\phi}}\left(s,a\right) &= R_{\theta}\left(s,a\right) + \gamma\int V_{\theta}^{\pi_{\phi}}\left(s_+\right) {\mathcal P}_\theta\left(s_{+}|s,a\right)\mathrm d s_+\\
V_{\theta}^{\pi_{\phi}}\left(s\right) &= \int Q_{\theta}^{\pi_{\phi}}\left(s,a\right) \pi_{\phi}\left(a|s\right)\mathrm d a
\end{align}
\end{subequations}
Their sensitivities are then given by:
\begin{subequations}
\label{eq:policy_evalution:sens:full}
\begin{align}
\partial Q_{\theta}^{\pi_{\phi}}\left(s,a\right) =&\, \partial R_{\theta}\left(s,a\right) + \gamma\int \partial V_{\theta}^{\pi_{\phi}}\left(s_+\right) {\mathcal P}_\theta\left(s_{+}|s,a\right)\mathrm d s_+\nonumber\\
&\, + \gamma\int V_{\theta}^{\pi_{\phi}}\left(s_+\right) {\mathcal P}_\theta\left(s_{+}|s,a\right) \partial \log {\mathcal P}_\theta\left(s_{+}|s,a\right)\mathrm d s_+  \\
\partial V_{\theta}^{\pi_{\phi}}\left(s\right) =&\, \int  \partial Q_{\theta}^{\pi_{\phi}}\left(s,a\right) \pi_{\phi}\left(a|s\right)\mathrm d a + \int  Q_{\theta}^{\pi_{\phi}}\left(s,a\right) \pi_{\phi}\left(a|s\right)\partial\log\pi_{\phi}\left(a|s\right)\mathrm d a
\end{align}
\end{subequations}
In the compact ``expected value" format, they then read as:
\begin{subequations}
\label{eq:policy_evalution:sens:compact}
\begin{align}
\partial Q_{\theta}^{\pi_{\phi}}\left(s,a\right) =&\; \partial R_{\theta}\left(s,a\right) + \gamma\mathbb E_{s_+\sim {\mathcal P}_\theta\left(s,a\right) }\left[ \partial V_{\theta}^{\pi_{\phi}}\left(s_+\right) + V_{\theta}^{\pi_{\phi}}\left(s_+\right) \partial \log {\mathcal P}_\theta\left(s_{+}|s,a\right) \right] \\
\partial V_{\theta}^{\pi_{\phi}}\left(s\right) =&\; \mathbb E_{a\sim \pi_{\phi}\left(a|s\right)}\left[  \partial Q_{\theta}^{\pi_{\phi}}\left(s,a\right) + Q_{\theta}^{\pi_{\phi}}\left(s,a\right)\partial\log\pi_{\phi}\left(a|s\right)\right] 
\end{align}
\end{subequations}%
\end{proof}
Note that this ``format" of sensitivity ($Q$ and $V$ split) is possibly best for a critic that has function approximators dedicated to $V$ and $Q$ separately (or rather $V$ and $A$).

\subsection*{Sensitivity of the Policy Evaluation w.r.t.\ \(\theta\) and \(\phi\)}

We want to derive separately the two sensitivities
\[
\nabla_\theta {Q}_{\theta}^{\pi_{\phi}}
\quad\text{and}\quad
\nabla_\phi {Q}_{\theta}^{\pi_{\phi}},
\]
starting from the general (``compact'') policy‐evaluation equations \eqref{eq:policy_evalution:sens:compact}:
% Here, \(\partial\) can stand for \(\frac{\partial}{\partial \theta}\) or \(\frac{\partial}{\partial \phi}\), depending on which parameters we are differentiating.  
Here, \emph{each} of the two sensitivity questions below assumes that the \emph{other} parameter is held fixed.  

\subsection*{1.\quad Sensitivity w.r.t.\ \(\theta\), keeping \(\phi\) fixed}

We want
\[
\nabla_\theta {Q}_{\theta}^{\pi_{\phi}}(s,a)
\quad\text{and}\quad
\nabla_\theta {V}_{\theta}^{\pi_{\phi}}(s).
\]
Starting from \eqref{eq:policy_evalution:sens:compact}:    
\begin{align}
      \nabla_\theta {Q}_{\theta}^{\pi_{\phi}}(s,a)
\;=\;
\nabla_\theta R_{\theta}(s,a)+\gamma
\mathbb{E}_{s_+\sim {\mathcal P}_\theta(\cdot\mid s,a)}
\Bigl[\nabla_\theta {V}_{\theta}^{\pi_{\phi}}(s_+) + {V}_{\theta}^{\pi_{\phi}}(s_+)\,\nabla_\theta \log {\mathcal P}_\theta(s_+\mid s,a)
\Bigr].  
\end{align}

Since, \(\pi_{\phi}\) is fixed w.r.t.\ \(\theta\), hence: \(\nabla_\theta\log \pi_{\phi}(a\mid s)
\;=\;0.\) We get

\begin{equation}
   \nabla_\theta {V}_{\theta}^{\pi_{\phi}}(s) = \mathbb{E}_{a\sim \pi_{\phi}(\cdot\mid s)}
\bigl[
\nabla_\theta {Q}_{\theta}^{\pi_{\phi}}(s,a)
\bigr]. 
\end{equation}

These two coupled equations show how to backpropagate the effect of changing \(\theta\) (the model or cost parameters) \emph{under a fixed policy} \(\pi_{\phi}\).

\subsection*{2.\quad Sensitivity w.r.t.\ \(\phi\), keeping \(\theta\) fixed}

Now we want
\[
\nabla_\phi {Q}_{\theta}^{\pi_{\phi}}(s,a)
\quad\text{and}\quad
\nabla_\phi {V}_{\theta}^{\pi_{\phi}}(s),
\]
with \(\theta\) fixed.  That means
\begin{equation}
   \nabla_\phi \log {\mathcal P}_\theta(s_+\mid s,a)
\;=\;0\,,
\end{equation}
and the reward sensitivity is
\begin{equation}
    \nabla_\phi R_{\theta}(s,a)
\;=\; 0\,.
\end{equation}

From \eqref{eq:policy_evalution:sens:compact} again:
\begin{align}
    \nabla_\phi {Q}_{\theta}^{\pi_{\phi}}(s,a) \;=&\; \gamma\,\mathbb{E}_{s_+\sim {\mathcal P}_\theta(\cdot\mid s,a)}
\Bigl[ \nabla_\phi {V}_{\theta}^{\pi_{\phi}}(s_+) \;\Bigr]\,, \\
    \nabla_\phi {V}_{\theta}^{\pi_{\phi}}(s) \;=&\; \mathbb{E}_{a\sim \pi_{\phi}(\cdot\mid s)}
\Bigl[
\nabla_\phi {Q}_{\theta}^{\pi_{\phi}}(s,a)
\;+\;
{Q}_{\theta}^{\pi_{\phi}}(s,a)\,\nabla_\phi \log \pi_{\phi}(a\mid s)
\Bigr].
\end{align}

Here, we iteratively compute the required gradients, $\nabla_\theta Q^{\pi_\phi}_\theta, \nabla_\theta V^{\pi_\phi}_\theta, \nabla_\phi Q^{\pi_\phi}_\theta, \nabla_\phi Q^{\pi_\phi}_\theta$, similar to temporal difference learning in value-based methods. 
For the illustrative examples, we calculate these values till convergence, but they can be approximated using a generic function approximator for bigger problems.

\subsection{Equivalency between the two critic sensitivity approximations}
\label{appendix:sensitivities-equivalency}

We first treat the two partial derivatives with the \emph{other} parameter held fixed.

\paragraph{\quad Sensitivity w.r.t.\ $\theta$ (policy $\pi_\phi$ fixed).}

Differentiating the “compact” policy-evaluation equations yields
\begin{align}
\label{eq:A1-Utheta}
\nabla_\theta Q^{\pi_\phi}_\theta(s,a)
&=
\nabla_\theta R_\theta(s,a)
+\gamma\,\mathbb E_{s'\sim {\mathcal P}_\theta(\cdot|s,a)}
\Big[\,\nabla_\theta V^{\pi_\phi}_\theta(s')
+V^{\pi_\phi}_\theta(s')\,\nabla_\theta\log {\mathcal P}_\theta(s'\mid s,a)\Big],
\\
\label{eq:A1-Wtheta}
\nabla_\theta V^{\pi_\phi}_\theta(s)
&=\mathbb E_{a\sim\pi_\phi(\cdot|s)}\big[\nabla_\theta Q^{\pi_\phi}_\theta(s,a)\big].
\end{align}
Substitute \eqref{eq:A1-Wtheta} into \eqref{eq:A1-Utheta} and recognize
the operator $P^{\pi_\phi}_\theta$ and the score identity:
\begin{align}
\nabla_\theta Q^{\pi_\phi}_\theta
&= \nabla_\theta R_\theta
+\gamma\,P^{\pi_\phi}_\theta \nabla_\theta V^{\pi_\phi}_\theta
+\gamma\Big(\nabla_\theta P^{\pi_\phi}_\theta\Big)Q^{\pi_\phi}_\theta \\
& = \nabla_\theta R_\theta
+\gamma\,P^{\pi_\phi}_\theta \nabla_\theta Q^{\pi_\phi}_\theta
+\gamma\Big(\nabla_\theta P^{\pi_\phi}_\theta\Big)Q^{\pi_\phi}_\theta.
\end{align}
The equality $P^{\pi_\phi}_\theta \nabla_\theta V^{\pi_\phi}_\theta=P^{\pi_\phi}_\theta \nabla_\theta Q^{\pi_\phi}_\theta$
is proved in Lemma~\ref{lem:PpiW-eq-PpiU} below. By rearranging, we get the Bellman fixed point equation,
\begin{equation}
\label{eq:A1-poisson}
(I-\gamma P^{\pi_\phi}_\theta)\,\nabla_\theta Q^{\pi_\phi}_\theta
\;=\; \nabla_\theta R_\theta
\;+\; \gamma\Big(\nabla_\theta P^{\pi_\phi}_\theta\Big)Q^{\pi_\phi}_\theta.
\end{equation}

\paragraph{\quad Sensitivity w.r.t.\ $\phi$ (dynamics ${\mathcal P}_\theta$ fixed).}
Similarly, with $\nabla_\phi R_\theta\equiv 0$ and $\nabla_\phi\log {\mathcal P}_\theta\equiv 0$,
\begin{align}
\label{eq:A2-Uphi}
\nabla_\phi Q^{\pi_\phi}_\theta(s,a)
&=\gamma\,\mathbb E_{s'\sim {\mathcal P}_\theta(\cdot|s,a)}\Big[\nabla_\phi V^{\pi_\phi}_\theta(s')\Big],
\\
\label{eq:A2-Wphi}
\nabla_\phi V^{\pi_\phi}_\theta(s)
&=\mathbb E_{a\sim\pi_\phi(\cdot|s)}\Big[\nabla_\phi Q^{\pi_\phi}_\theta(s,a)+Q^{\pi_\phi}_\theta(s,a)\,\nabla_\phi\log\pi_\phi(a\mid s)\Big].
\end{align}
Substitute \eqref{eq:A2-Wphi} into \eqref{eq:A2-Uphi} we get the Bellman fixed point equation:
\begin{align}
\nabla_\phi Q^{\pi_\phi}_\theta
&=\gamma\,P^{\pi_\phi}_\theta \nabla_\phi V^{\pi_\phi}_\theta
+\gamma\Big(\nabla_\phi P^{\pi_\phi}_\theta\Big)Q^{\pi_\phi}_\theta,
\quad\Longleftrightarrow\quad
(I-\gamma P^{\pi_\phi}_\theta)\,\nabla_\phi Q^{\pi_\phi}_\theta
\;=\; \gamma\Big(\nabla_\phi P^{\pi_\phi}_\theta\Big)Q^{\pi_\phi}_\theta.
\label{eq:A2-poisson}
\end{align}

\paragraph{ Proving $\,P^{\pi_\phi}_\theta \nabla_\theta V^{\pi_\phi}_\theta=P^{\pi_\phi}_\theta \nabla_\theta Q^{\pi_\phi}_\theta\,$}
\label{sec:why-PW-eq-PU}

\begin{lemma}
\label{lem:PpiW-eq-PpiU}
Let $Q^{\pi_\phi}_\theta:\mathcal S\times\mathcal A\to\mathbb R$ be bounded and define
$V^{\pi_\phi}_\theta(s):=\mathbb E_{a\sim\pi_\phi(\cdot|s)}[Q^{\pi_\phi}_\theta(s,a)]$. Then
\[
(P^{\pi_\phi}_\theta V^{\pi_\phi}_\theta)(s,a)
\;=\;
(P^{\pi_\phi}_\theta Q^{\pi_\phi}_\theta)(s,a)\qquad\forall(s,a)\in\mathcal S\times\mathcal A.
\]
\end{lemma}
\begin{proof}
By the definition \eqref{eq:Ppi-def} of $P^{\pi_\phi}_\theta$,
\begin{align*}
(P^{\pi_\phi}_\theta V^{\pi_\phi}_\theta)(s,a)
&=\int_{\mathcal S} V^{\pi_\phi}_\theta(s')\,{\mathcal P}_\theta(\mathrm ds'\mid s,a)
=\int_{\mathcal S}\Big(\int_{\mathcal A} Q^{\pi_\phi}_\theta(s',a')\,\pi_\phi(\mathrm da'\mid s')\Big)\,{\mathcal P}_\theta(\mathrm ds'\mid s,a)
\\
&=\int_{\mathcal S}\!\!\int_{\mathcal A} Q^{\pi_\phi}_\theta(s',a')\,\pi_\phi(\mathrm da'\mid s')\,{\mathcal P}_\theta(\mathrm ds'\mid s,a)
=(P^{\pi_\phi}_\theta Q^{\pi_\phi}_\theta)(s,a).
\end{align*}

$P^{\pi_\phi}_\theta$ already does \emph{both}
``go to the next state'' and ``sample the next action under $\pi_\phi$''; irrespective of if the average is taken over $a'$ inside $V^{\pi_\phi}_\theta$ first, or let $P^{\pi_\phi}_\theta$ average over $a'$ directly,
the result is the same.
\end{proof}

Coupled ($Q/V$) critic sensitivity recursions and the Poisson/operator sensitivity equation
describe the \emph{same} linear system.
Lemma~\ref{lem:PpiW-eq-PpiU} explains why we can eliminate $\nabla V$
and obtain the compact operator form
$(I-\gamma P^{\pi_\phi}_\theta)\nabla_\eta Q^{\pi_\phi}_\theta=b_\eta$.

%%%%%%%%%%%%%%%%%%%%%%%%%%%%%%%%%%%%%%%%%%%%%%%%%%%%%%%%%%%%%%%%%%%%%%%%%%%%%%%%%%%%%%%%%%%%%%%%%%%%%%%
\section{Convergence Analysis of Bi-Level SPG via Three Timescales}
\label{appendix:convergence_analysis}

In this section, we establish almost-sure convergence of the full three-timescale scheme where (i) the critic and its sensitivities are estimated on the \emph{fastest} timescale,
(ii) the inner (in-sim) policy parameter $\phi$ on a \emph{medium} timescale,
and (iii) the outer simulation parameter $\theta$ on the \emph{slowest} timescale.
The two-timescale and nested/single-level cases are immediate corollaries.

\begin{lemma}[Affine contraction for sensitivities]
\label{lem:affine-contraction}
Fix $(\phi,\theta)$ and define, on the Banach space
$\bigl(\mathcal B_\infty,\|\cdot\|_\infty\bigr)$ of bounded measurable
functions on $\mathcal S\times\mathcal A$, the affine map
\[
\mathcal C_\eta:\ \nabla_{\eta} Q_\theta^{\pi_\phi}\ \mapsto\
b_\eta\;+\;\gamma\,P^{\pi_\phi}_\theta\,\nabla_{\eta} Q_\theta^{\pi_\phi},
\qquad
b_\eta\ :=\ \nabla_{\eta}R_\theta\;+\;\gamma\,(\nabla_{\eta}P^{\pi_\phi}_\theta)\,Q^{\pi_\phi}_\theta,
\]
where $\eta\in\{\phi,\theta\}$ and $\gamma\in(0,1)$. Then $\mathcal C_\eta$ is a
$\gamma$-contraction in $\|\cdot\|_\infty$, hence admits a unique fixed point
$\nabla_{\eta}Q^{\pi_\phi}_\theta\in\mathcal B_\infty$ satisfying
\[
\nabla_{\eta}Q^{\pi_\phi}_\theta
\;=\; (I-\gamma P^{\pi_\phi}_\theta)^{-1} b_\eta
\;=\;\sum_{t=0}^{\infty}(\gamma P^{\pi_\phi}_\theta)^t\,b_\eta.
\]

\end{lemma}

\begin{proof}
\textit{Contraction in $\|\cdot\|_\infty$.}
Let $G_1,G_2\in\mathcal B_\infty$ be two candidate sensitivity functions for 
$\nabla_{\eta} Q_\theta^{\pi_\phi}$. Since $b_\eta$ does not depend on
the input function, we have
\[
\mathcal C_\eta(G_1)-\mathcal C_\eta(G_2)
=\gamma\,P^{\pi_\phi}_\theta\,(G_1-G_2).
\]
The policy–induced Markov operator $P^{\pi_\phi}_\theta$ is an averaging
operator, hence non-expansive in the sup–norm, i.e. 
$\|P^{\pi_\phi}_\theta f\|_\infty\le \|f\|_\infty$ for all
$f\in\mathcal B_\infty$. Therefore
\[
\|\mathcal C_\eta(G_1)-\mathcal C_\eta(G_2)\|_\infty
=\gamma\,\|P^{\pi_\phi}_\theta(G_1-G_2)\|_\infty
\ \le\ \gamma\,\|G_1-G_2\|_\infty.
\]
Since $\gamma\in(0,1)$, $\mathcal C_\eta$ is a contraction with modulus $\gamma$.

\smallskip
% \paragraph{ Existence and uniqueness (Banach fixed point).}
Because $(\mathcal B_\infty,\|\cdot\|_\infty)$ is complete, Banach’s fixed–point
theorem implies that $\mathcal C_\eta$ has a unique fixed point
$G^\star\in\mathcal B_\infty$, and 
$G^{(k+1)}=\mathcal C_\eta(G^{(k)})$ converges to $G^\star$ for any start
$G^{(0)}$~\citep{granas2003fixed}. By
construction, $G^\star=\nabla_{\eta}Q^{\pi_\phi}_\theta$, the fixed–point identity
is:
\[
\nabla_{\eta}Q^{\pi_\phi}_\theta=(I-\gamma P^{\pi_\phi}_\theta)^{-1} b_\eta
=\sum_{t=0}^{\infty}(\gamma P^{\pi_\phi}_\theta)^t b_\eta,
\]
where the Neumann series converges because
$\|\gamma P^{\pi_\phi}_\theta\|\le \gamma<1$. 
\end{proof}

\paragraph{Fast-layer mean field from three Bellman residuals.}
Let us represent the variables in the fast layer estimating critic and its sensitivities by \(z=(w,u,v)\), where
\(w\) estimates \(Q^{\pi_\phi}_\theta\) and
\(u\), \(v\) estimate \(\nabla_\phi Q^{\pi_\phi}_\theta\),
\(\nabla_\theta Q^{\pi_\phi}_\theta\).
Their residuals with respect to the fixed points can be written as:
\[
\begin{aligned}
H_w(w;\phi,\theta) &:= R_\theta + \gamma P^{\pi_\phi}_\theta Q^{\pi_\phi}_\theta - Q^{\pi_\phi}_\theta,\\
H_u(u,w;\phi,\theta) &:= b_\phi(w;\phi,\theta) + \gamma P^{\pi_\phi}_\theta u - u,\\
H_v(v,w;\phi,\theta) &:= b_\theta(w;\phi,\theta) + \gamma P^{\pi_\phi}_\theta v - v,
\end{aligned}
\]
and the stacked mean field
\[
H(z;\phi,\theta)\;:=\;\big(H_w(w;\phi,\theta),\; H_u(u,w;\phi,\theta),\; H_v(v,w;\phi,\theta)\big).
\]

\begin{lemma}[Existence, stability, and Lipschitz dependence of $z^\star$]
\label{lem:fast-existence-stability}
For each $(\phi,\theta)$, consider the fast-layer ODE
\[
\dot z \;=\; H(z;\phi,\theta),
\qquad
z=(w,u,v),
\]
with the stacked drift
\[
H(z;\phi,\theta)
:=\bigl(\underbrace{R_\theta+\gamma P^{\pi_\phi}_\theta w - w}_{=:H_w(w)},\;
\underbrace{b_\phi(w;\phi,\theta)+\gamma P^{\pi_\phi}_\theta u - u}_{=:H_u(u,w)},\;
\underbrace{b_\theta(w;\phi,\theta)+\gamma P^{\pi_\phi}_\theta v - v}_{=:H_{U_\theta}(v,w)}\bigr).
\]
Here $\gamma\in(0,1)$, $P^{\pi_\phi}_\theta$ is the policy-induced Markov operator,
$R_\theta$ the reward function, and
\[
b_\eta(w;\phi,\theta)
:= \nabla_{\eta}R_\theta \;+\; \gamma\,(\nabla_{\eta}P^{\pi_\phi}_\theta)\,w,
\qquad \eta\in\{\phi,\theta\}.
\]
Assume: (i) $P^{\pi_\phi}_\theta$ is a contraction in the chosen norm
($\|P^{\pi_\phi}_\theta f\|\le \|f\|$), (ii) $\nabla_{\eta}P^{\pi_\phi}_\theta$ and
$\nabla_{\eta}R_\theta$ are locally Lipschitz in $(\phi,\theta)$, and (iii) all
iterates/limits are restricted to a compact set. Then:

\begin{enumerate}
\item Existence and uniqueness of $z^\star$: for each $(\phi,\theta)$ there is a unique equilibrium $z^\star(\phi,\theta)$ with
$H\bigl(z^\star(\phi,\theta);\phi,\theta\bigr)=0$.
\item Dissipativity and stability: there exists $\mu>0$ (independent of $z$ on compacts) such that for all $z$,
\begin{equation}
\label{eq:dissipativity}
\big\langle H(z;\phi,\theta)-H\!\left(z^\star(\phi,\theta);\phi,\theta\right),\; z-z^\star(\phi,\theta)\big\rangle
\;\le\; -\,\mu\,\|z-z^\star(\phi,\theta)\|^2,
\end{equation}
where $\langle\cdot,\cdot\rangle$ and $\|\cdot\|$ are the inner product and norm induced by
the chosen (Hilbert) metric.
\item Lipschitz dependence: there exists a constant
$L_{z^\star}<\infty$ such that for all $(\phi,\theta),(\tilde\phi,\tilde\theta)$ in a compact set,
\[
\big\|z^\star(\phi,\theta)-z^\star(\tilde\phi,\tilde\theta)\big\|
\;\le\; L_{z^\star}\big(\|\phi-\tilde\phi\|+\|\theta-\tilde\theta\|\big).
\]
\end{enumerate}
\end{lemma}

\begin{proof}
For fixed $(\phi,\theta)$, the maps
\[
T(w):=R_\theta+\gamma P^{\pi_\phi}_\theta w,
\qquad
\mathcal C_\eta(u;w):=b_\eta(w;\phi,\theta)+\gamma P^{\pi_\phi}_\theta u
\]
are $\gamma$-contractions in $w$ and $u$ respectively (Lemma~\ref{lem:affine-contraction}),
with contraction modulus $\gamma\in(0,1)$ and bounded affine terms on compacts.

\smallskip
\noindent\textbf{(1) Existence and uniqueness of $z^\star$.}
The stacked system is \emph{upper triangular} in $(w,u,v)$:

\begin{itemize}
\item Since $T$ is a $\gamma$-contraction, the Banach fixed-point theorem gives a unique
$w^\star(\phi,\theta)$ such that $w^\star=T(w^\star)$.

\item Given $w^\star$, the $u$-block satisfies
$H_u(u,w^\star)=0 \iff u=\mathcal C_\phi(u;w^\star)$, and
$\mathcal C_\phi(\cdot;w^\star)$ is a $\gamma$-contraction in $u$. Thus there is a unique
$u^\star(\phi,\theta)=(I-\gamma P^{\pi_\phi}_\theta)^{-1} b_\phi(w^\star;\phi,\theta)$.

\item Similarly, $v^\star(\phi,\theta)=(I-\gamma P^{\pi_\phi}_\theta)^{-1} b_\theta(w^\star;\phi,\theta)$
is unique.
\end{itemize}

Hence $z^\star(\phi,\theta):=(w^\star,u^\star,v^\star)$ is the unique solution of
$H\bigl(z;\phi,\theta\bigr)=0$.

\smallskip
\noindent\textbf{(2) Dissipativity and stability.}
Let $e_w:=w-w^\star$, $e_u:=u-u^\star$, $e_v:=v-v^\star$, and $e:=(e_w,e_u,e_v)$.
We work in a Hilbert norm so that
$\langle\cdot,\cdot\rangle$ is the associated inner product and Cauchy--Schwarz holds.

Since $T$ is a $\gamma$-contraction,
\begin{align*}
\big\langle H_w(w)-H_w(w^\star),\, e_w\big\rangle
& = \big\langle T(w)-T(w^\star) - (w-w^\star),\, e_w\big\rangle \\
& \qquad\qquad\qquad \le \|T(w)-T(w^\star)\|\,\|e_w\| - \|e_w\|^2
\le (\gamma-1)\|e_w\|^2.
\end{align*}
Thus
$\langle H_w(w)-H_w(w^\star), e_w\rangle \le -(1-\gamma)\|e_w\|^2$.

Using $H_u(u,w)=b_\phi(w)+\gamma P^{\pi_\phi}_\theta u - u$ and
$H_u(u^\star,w^\star)=0$,
\[
H_u(u,w)-H_u(u^\star,w^\star)
= \underbrace{\bigl(b_\phi(w)-b_\phi(w^\star)\bigr)}_{\text{depends on }e_w}
+ \gamma P^{\pi_\phi}_\theta e_u - e_u.
\]
Now taking the inner product with $e_u$ and using the non-expansiveness of $P^{\pi_\phi}_\theta$:
\[
\langle H_u(u,w)-H_u(u^\star,w^\star),\, e_u\rangle
\le (\gamma-1)\|e_u\|^2 + \|b_\phi(w)-b_\phi(w^\star)\|\,\|e_u\|.
\]
By local Lipschitzness of $b_\phi(\cdot;\phi,\theta)$ in $w$ on compacts:
$\|b_\phi(w)-b_\phi(w^\star)\|\le L_\phi \|e_w\|$.
Applying Young’s inequality with any $\varepsilon_\phi>0$ yields:
\[
\|b_\phi(w)-b_\phi(w^\star)\|\,\|e_u\|
\le \frac{\varepsilon_\phi}{2}\|e_u\|^2 + \frac{L_\phi^2}{2\varepsilon_\phi}\|e_w\|^2.
\]
Hence
\[
\langle H_u(u,w)-H_u(u^\star,w^\star),\, e_u\rangle
\le -\bigl(1-\gamma-\tfrac{\varepsilon_\phi}{2}\bigr)\|e_u\|^2
\;+\; \frac{L_\phi^2}{2\varepsilon_\phi}\,\|e_w\|^2.
\]

The same argument gives, for some $L_\theta$ and $\varepsilon_\theta>0$,
\[
\langle H_{U_\theta}(v,w)-H_{U_\theta}(v^\star,w^\star),\, e_v\rangle
\le -\bigl(1-\gamma-\tfrac{\varepsilon_\theta}{2}\bigr)\|e_v\|^2
\;+\; \frac{L_\theta^2}{2\varepsilon_\theta}\,\|e_w\|^2.
\]

 Summing the bounds for \(H_w, H_u\) and \(H_v\) gives:
\[
\big\langle H(z;\phi,\theta)-H(z^\star;\phi,\theta),\, e\big\rangle
\le - (1-\gamma)\|e_w\|^2
-\Bigl(1-\gamma-\tfrac{\varepsilon_\phi}{2}\Bigr)\|e_u\|^2
-\Bigl(1-\gamma-\tfrac{\varepsilon_\theta}{2}\Bigr)\|e_v\|^2
+ C_w\,\|e_w\|^2,
\]
with $C_w:=\tfrac{L_\phi^2}{2\varepsilon_\phi}+\tfrac{L_\theta^2}{2\varepsilon_\theta}$.
Choose $\varepsilon_\phi,\varepsilon_\theta>0$ small enough and use compactness to ensure
$C_w < (1-\gamma)/2$. Then there exists $\mu\in(0,1-\gamma)$ such that
\[
\big\langle H(z;\phi,\theta)-H(z^\star;\phi,\theta),\, z-z^\star\big\rangle
\le -\,\mu\,\bigl(\|e_w\|^2+\|e_u\|^2+\|e_v\|^2\bigr)
= -\,\mu\,\|z-z^\star\|^2.
\]
This is \eqref{eq:dissipativity}. In particular, the ODE
$\dot z = H(z;\phi,\theta)$ is globally (on the compact set) asymptotically stable at $z^\star$.

\smallskip
\noindent\textbf{(3) Lipschitz dependence $(\phi,\theta)\mapsto z^\star(\phi,\theta)$.}
We use a fixed-point perturbation argument.

\emph{Critic $w^\star$.} For $(\phi,\theta)$ and $(\tilde\phi,\tilde\theta)$, let
$T:=R_\theta+\gamma P^{\pi_\phi}_\theta(\cdot)$ and
$\tilde T:=R_{\tilde\theta}+\gamma P^{\pi_{\tilde\phi}}_{\tilde\theta}(\cdot)$.
Both are $\gamma$-contractions. A standard bound for fixed points of contractions gives
\[
\|w^\star-\tilde w^\star\|
\;\le\; \frac{1}{1-\gamma}\,\big\|T(\tilde w^\star)-\tilde T(\tilde w^\star)\big\|
\;\le\; \frac{1}{1-\gamma}\,\Big(\|R_\theta-  R_{\tilde\theta}\|
+ \gamma\,\|P^{\pi_\phi}_\theta-P^{\pi_{\tilde\phi}}_{\tilde\theta}\|\,\|\tilde w^\star\|\Big).
\]
On a compact parameter set, $\|\tilde w^\star\|$ is bounded and
$(\phi,\theta)\mapsto (R_\theta,P^{\pi_\phi}_\theta)$ is locally Lipschitz; hence
$\|w^\star-\tilde w^\star\|\le L_w(\|\phi-\tilde\phi\|+\|\theta-\tilde\theta\|)$.

\emph{Sensitivities $(u^\star,v^\star)$.} With
$u^\star=(I-\gamma P^{\pi_\phi}_\theta)^{-1} b_\phi(w^\star;\phi,\theta)$, we have
\[
\|u^\star-\tilde u^\star\|
\le \big\|(I-\gamma P^{\pi_\phi}_\theta)^{-1}\big\|\,\big\|b_\phi(w^\star;\phi,\theta)-b_\phi(\tilde w^\star;\tilde\phi,\tilde\theta)\big\|
+ \big\|(I-\gamma P^{\pi_\phi}_\theta)^{-1}-(I-\gamma P^{\pi_{\tilde\phi}}_{\tilde\theta})^{-1}\big\|\;\|b_\phi(\tilde w^\star;\tilde\phi,\tilde\theta)\|.
\]
The resolvent is uniformly bounded by $(1-\gamma)^{-1}$, and
$(\phi,\theta,w)\mapsto b_\phi(w;\phi,\theta)$ is locally Lipschitz. The resolvent difference
is locally Lipschitz in $P$ on compacts. Using the bound for $\|w^\star-\tilde w^\star\|$
and Lipschitzness of $(\phi,\theta)\mapsto P^{\pi_\phi}_\theta$, we conclude
$\|u^\star-\tilde u^\star\|\le L_u(\|\phi-\tilde\phi\|+\|\theta-\tilde\theta\|)$.
An identical argument applies to $v^\star$.

Combining the three blocks yields the stated Lipschitz bound for $z^\star$.
\end{proof}

\paragraph{Smoothness and projected smoothness.}
On a compact $\Theta$, if $F$ is $C^1$ with $L$-Lipschitz gradient,
then for any $x,y\in \Theta$,
$F(y)\ge F(x)+\nabla F(x)^\top (y-x)-\tfrac{L}{2}\|y-x\|^2$~\citep{nesterov2013introductory}.
If $\Pi_\Theta$ is the metric projection, then
$F(\Pi_\Theta(x+u))\ge F(x)+\nabla F(x)^\top u-\tfrac{L}{2}\|u\|^2$.

\paragraph{Robbins--Siegmund.}
We use the standard almost-supermartingale theorem: if
$V_{k+1}\le (1-a_k)V_k+b_k-\xi_k$, with $a_k\in[0,1]$, $\sum a_k=\infty$,
$\sum b_k<\infty$, $V_k\ge 0$, then $V_k$ converges and $\sum a_k V_k<\infty$~\citep{borkar2008stochastic}.

\medskip

% ------------------------------------------------------------------------------
\subsection{ODE Approach}
\paragraph{Stochastic-approximation (SA) viewpoint}
Each update in Algorithm~\ref{alg:bi_level_rl} is a noisy Euler step for a mean-field drift; with projections keeping iterates bounded. Concretely, with stepsizes $(\gamma_k)$ for the fast layer $z=(w,u,v)$, $(\beta_k)$ for the medium layer $\phi$, and $(\alpha_k)$ for the slow layer $\theta$, the projected recursions implement
\begin{align}
\text{(fast)}\quad &z_{k+1}=\Pi_{\mathcal Z}\!\big(z_k+\gamma_k[\,H(z_k;\phi_k,\theta_k)+M^{(z)}_{k+1}\,]\big), \label{eq:fast_layer}\\
\text{(medium)}\quad &\phi_{k+1}=\Pi_{\Phi}\!\big(\phi_k+\beta_k[\,\hat\varphi(\phi_k,\theta_k;z_k)+M^{(\phi)}_{k+1}\,]\big),\label{eq:medium_layer}\\
\text{(slow)}\quad &\theta_{k+1}=\Pi_{\Theta}\!\big(\theta_k+\alpha_k[\,\varphi(\phi_k,\theta_k;z_k)+M^{(\theta)}_{k+1}\,]\big).\label{eq:slow_layer}
\end{align}

\paragraph{Outer-level gradient.}
Define
\begin{equation}
\label{eq:outer-potential}
F(\theta):=J_{\text{real}}\!\big(\pi_{\phi^\star(\theta)}\big),
\qquad
\nabla_\theta F(\theta)=\varphi\!\big(\phi^\star(\theta),\theta;\,z^{\!}(\phi^\star(\theta),\theta)\big).
\end{equation}

\noindent \textbf{Assumptions (Three Timescales).}
Let $\mathcal Z,\Phi,\Theta$ be compact convex; all norms are Euclidean/operator.

\begin{enumerate}
\item {Regularity/IFT for inner best response.}
$\phi^\star(\theta)$ is locally unique with nonsingular
$\nabla_\phi \hat\varphi(\phi^\star(\theta),\theta)$; thus $\phi^\star\in C^1$.

\item {Smoothness / finite moments.}
Rewards are bounded; $\log\pi_\phi,{\mathcal P}_{\theta},R_\theta$ are $C^1$ (or $C^2$ as used),
locally Lipschitz on $\Phi\times\Theta$. On-policy Markov chains (sim and real)
are geometrically ergodic; all MC estimators have bounded second moments.

\item {Fast-layer contraction \& Lipschitz.}
Lemma~\ref{lem:affine-contraction} and Lemma~\ref{lem:fast-existence-stability} hold
(for critic \emph{and} sensitivities, with projection if using linear FA).

\item {Inner-layer stability.}
For each fixed $\theta$, there exists $m>0$ s.t.\ for $\phi$ near $\phi^\star(\theta)$,
$\langle \hat\varphi(\phi,\theta)-\hat\varphi(\phi^\star,\theta),\,\phi-\phi^\star\rangle
\le -m\|\phi-\phi^\star\|^2$.

\item {Stepsizes and separation.}
Let \(\gamma, \beta\) and \(\alpha)\) represents the step sizes for fast, medium and slow ODEs.
$\sum_k \gamma_k=\sum_k \beta_k=\sum_k \alpha_k=\infty$,
$\sum_k \gamma_k^2,\sum_k \beta_k^2,\sum_k \alpha_k^2<\infty$,
and $\beta_k/\gamma_k\to 0$, $\alpha_k/\beta_k\to 0$.

\item {Noise.}
$M^{(z)}_{k},M^{(\phi)}_{k},M^{(\theta)}_{k}$ are square-integrable martingale differences with uniformly bounded conditional second moments.

\item {Projection.}
Updates are projected onto $\mathcal Z,\Phi,\Theta$; projections are nonexpansive.
\end{enumerate}

Under \textbf{A1}--\textbf{A7} (regularity, bounded moments, contraction/stability, stepsize separation, projections), the corresponding mean-field ODEs are
\[
\dot z = H(z;\phi,\theta), \qquad
\dot \phi = \hat\varphi(\phi,\theta;z), \qquad
\dot \theta = \varphi(\phi,\theta;z).
\]

\paragraph{Steps that follows}
To establish this three timestep convergence, we need to establish:
\begin{enumerate}
\item \textit{Fast equilibrium is well-posed, stable, and smooth in $(\phi,\theta)$.}
This is provided by Lemma~\ref{lem:fast-existence-stability} using affine $\gamma$-contraction of Bellman/Poisson operators for the critic and its sensitivities.
\item \textit{Fast tracking with a moving target.}
Despite the slow motion of $(\phi_k,\theta_k)$, the error $e_k^z$ contracts up to a small \emph{moving-target} term $\Delta_{k+1}^z:=z^\star(\phi_{k+1},\theta_{k+1})-z^\star(\phi_k,\theta_k)$, which is square-summable because $\|\Delta_{k+1}^z\|=O(\beta_k)+O(\alpha_k)$ and $\sum_k (\beta_k^2+\alpha_k^2)<\infty$. This is provided by Lemma~\ref{lem:fast-tracking-3ts}.
\item \textit{Medium tracking with a moving target and a fast-layer perturbation.}
Define the reduced inner drift $h^\star$ and note
$\hat\varphi(\phi_k,\theta_k;z_k) = h^\star(\phi_k,\theta_k) + \delta_k^z$ with
$\delta_k^z:=\hat\varphi(\phi_k,\theta_k;z_k)-\hat\varphi(\phi_k,\theta_k;z^\star(\phi_k,\theta_k))$.
The perturbation obeys $\|\delta_k^z\|\le L\|e_k^z\|$ by Lipschitzness, while the moving target
$\Delta_{k+1}^\phi:=\phi^\star(\theta_{k+1})-\phi^\star(\theta_k)$ is $O(\alpha_k)$.
A Lyapunov (quadratic) drift argument with Young’s inequality and \textbf{A4} (local strong stability) proves Lemma~\ref{lem:medium-tracking-3ts}.
\item \textit{Slow layer as SGD with summable bias.}
Write the slow estimator as
\[
\widehat g_k=\varphi(\phi_k,\theta_k;z_k)+M^{(\theta)}_{k+1}
=g(\theta_k) + b_k + M^{(\theta)}_{k+1},
\]
where \(g(\theta)\;:=\;\varphi\big(\phi^\star(\theta),\theta;\,z^{\!}(\phi^\star(\theta),\theta)\big)\)  and the bias $b_k$ satisfies $\|b_k\|\le C(\|e_k^\phi\|+\|e_k^z\|)$. Timescale separation gives
$\sum_k \alpha_k\|b_k\|^2<\infty$ (Lemma~\ref{lem:slow-3ts}). The standard smooth-SGD inequality with projection and Robbins--Siegmund then yield convergence to the stationary set $\{\nabla_\theta F=0\}$.
\end{enumerate}

\paragraph{Error notations used in the lemmas.}
We introduce the tracking errors and the ``moving-target'' increments for the fast and medium layers:
\begin{align}
    &e_k^z:=z_k-z_k^\star,\qquad \Delta_{k+1}^z:=z_{k+1}^\star-z_k^\star,\\
&e_k^\phi:=\phi_k-\phi^\star(\theta_k),\qquad
\Delta_{k+1}^\phi:=\phi^\star(\theta_{k+1})-\phi^\star(\theta_k).
\end{align}
The \(^\star\) notation represents the target at an iterate \(k\).
The lemmas quantify how $e_k^z$ and $e_k^\phi$ contract up to the small (square-summable)
perturbations $\Delta_{k+1}^z,\Delta_{k+1}^\phi$ and the martingale noises, and how this yields a summable bias for the slow layer.

% ------------------------------------------------------------------------------
\subsection{Lemmas and Main Theorem}

\begin{lemma}[Fast tracking of critic and sensitivities]
\label{lem:fast-tracking-3ts}
Work in the $D_\mu$--Hilbert norm $\|f\|_{D_\mu}^2=\sum_{s,a}\mu(s,a)f(s,a)^2$ (or its block product for $z=(w,u,v)$),
with inner product $\langle \cdot,\cdot\rangle_{D_\mu}$. 
Under \textbf{A2}--\textbf{A7} and Lemma~\ref{lem:fast-existence-stability}, for the fast recursion
\[
z_{k+1} = \Pi_{\mathcal Z}\!\Big(z_k + \gamma_k\big[\,H(z_k;\phi_k,\theta_k) + M^{(z)}_{k+1}\big]\Big),
\quad e_k^z := z_k - z_k^\star,
\]
we have $e_k^z\to 0$ almost surely. In particular,
$\sum_{k=0}^\infty \gamma_k\,\mathbb E\|e_k^z\|_{D_\mu}^2<\infty$ and $\|e_k^z\|_{D_\mu}\to 0$ a.s.
\end{lemma}

\begin{proof}

Fix $k$ and abbreviate $z_k^\star:=z^{\!}(\phi_k,\theta_k)$, $H_k:=H(z_k;\phi_k,\theta_k)$, $D_k:=H_k-H_k^\star=H(z_k;\phi_k,\theta_k)-H(z_k^\star;\phi_k,\theta_k)$, and
$\Delta_{k+1}^z:=z_{k+1}^\star-z_k^\star$.

From Lemma~\ref{lem:fast-existence-stability} (proved via Bellman/Poisson contractions):
\begin{align}
\text{(dissipativity)}\qquad
&\langle D_k,\,e_k^z\rangle_{D_\mu}\ \le\ -\,\mu\,\|e_k^z\|_{D_\mu}^2,
\label{eq:dissi}\\
\text{(Lipschitz in $z$)}\qquad
&\|D_k\|_{D_\mu}\ \le\ L_H\,\|e_k^z\|_{D_\mu},
\label{eq:LipH}
\end{align}
for some $\mu>0$ and $L_H<\infty$, uniform on the compact set where iterates live.

\medskip
By nonexpansiveness of $\Pi_{\mathcal Z}$, adding and subtracting \(H_k^\star\) and using \(H_k^\star = 0\) we can write,
\begin{align}
\|e_{k+1}^z\|_{D_\mu}
&= \big\|\Pi_{\mathcal Z}\big(z_k+\gamma_k(H_k+M^{(z)}_{k+1})\big)-z_{k+1}^\star \big\|_{D_\mu}\nonumber\\
&\le \big\| z_k+\gamma_k(H_k+M^{(z)}_{k+1})-z_{k+1}^\star \big\|_{D_\mu}\nonumber\\
&= \big\| e_k^z + \gamma_k D_k + \gamma_k M^{(z)}_{k+1} - \Delta_{k+1}^z \big\|_{D_\mu}.
\label{eq:one-step}
\end{align}

Now by the martingale-difference property (Assumption~\textbf{A6}),
\[
\mathbb E\langle e_k^z+\gamma_k D_k-\Delta_{k+1}^z,\ M^{(z)}_{k+1}\mid\mathcal F_k\rangle_{D_\mu}=0,
\quad
\mathbb E\|M^{(z)}_{k+1}\|_{D_\mu}^2\mid\mathcal F_k\le \sigma_z^2.
\]
Using this to take the conditional expectation $\mathbb E[\cdot\mid\mathcal F_k]$ of the square of \eqref{eq:one-step} gives:
\begin{align}
\mathbb E\!\left[\|e_{k+1}^z\|_{D_\mu}^2\mid\mathcal F_k\right]
&\le \|e_k^z+\gamma_k D_k-\Delta_{k+1}^z\|_{D_\mu}^2 + \gamma_k^2 \sigma_z^2.
\label{eq:cond-master}
\end{align}

According to Young's inequality, for any $\eta>0$ and any vectors $x,y$, $\|x-y\|^2\le (1+\eta)\|x\|^2+(1+1/\eta)\|y\|^2$.
Apply with $x=e_k^z+\gamma_k D_k$ and $y=\Delta_{k+1}^z$, then take
$\mathbb E[\cdot\mid\mathcal F_k]$:
\begin{align}
\mathbb E\!\left[\|e_{k+1}^z\|_{D_\mu}^2\mid\mathcal F_k\right]
&\le (1+\eta)\,\|e_k^z+\gamma_k D_k\|_{D_\mu}^2
\;+\; \Big(1+\tfrac{1}{\eta}\Big)\,\mathbb E\!\left[\|\Delta_{k+1}^z\|_{D_\mu}^2\mid\mathcal F_k\right]
\;+\; \gamma_k^2 \sigma_z^2 .
\label{eq:s1}
\end{align}

Now let us consider only the term $\|e_k^z+\gamma_k D_k\|^2$. By dissipativity  and Lipschitzness of $H$ in $z$ as shown in \eqref{eq:dissi} and \eqref{eq:LipH}, we get:
\begin{align}
\|e_k^z+\gamma_k D_k\|_{D_\mu}^2
&= \|e_k^z\|_{D_\mu}^2 + 2\gamma_k \langle e_k^z, D_k\rangle_{D_\mu}
   + \gamma_k^2 \|D_k\|_{D_\mu}^2 \nonumber\\
&\le \big(1 - 2\mu \gamma_k + L_H^2 \gamma_k^2\big)\,\|e_k^z\|_{D_\mu}^2
\;\le\; \big(1 - 2\mu \gamma_k + C \gamma_k^2\big)\,\|e_k^z\|_{D_\mu}^2 .
\label{eq:s2}
\end{align}

By the Lipschitz dependence of $z^{\!}(\cdot,\cdot)$ (Lemma~\ref{lem:fast-existence-stability})
and projected bounded steps (Assumptions \textbf{A2}, \textbf{A7}),
\[
\|\Delta_{k+1}^z\|_{D_\mu}
=\|z^\star(\phi_{k+1},\theta_{k+1})-z^\star(\phi_k,\theta_k)\|_{D_\mu}
\le L_{z^\star}\big(\|\phi_{k+1}-\phi_k\|+\|\theta_{k+1}-\theta_k\|\big).
\]
Taking $\mathbb E[\cdot\mid\mathcal F_k]$ and using
$\mathbb E\|\phi_{k+1}-\phi_k\|^2\le C\beta_k^2$ and
$\mathbb E\|\theta_{k+1}-\theta_k\|^2\le C\alpha_k^2$,
\begin{equation}
\label{eq:target-motion}
\mathbb E\!\left[\|\Delta_{k+1}^z\|_{D_\mu}^2\mid\mathcal F_k\right]\;\le\; C\big(\beta_k^2+\alpha_k^2\big).
\end{equation}

Plug \eqref{eq:s2} and \eqref{eq:target-motion} into \eqref{eq:s1}, and choose
$\eta=\mu \gamma_k$ (small for large $k$):
\begin{align*}
\mathbb E\!\left[\|e_{k+1}^z\|_{D_\mu}^2\mid\mathcal F_k\right]
&\le (1+\mu\gamma_k)\big(1 - 2\mu \gamma_k + C \gamma_k^2\big)\,\|e_k^z\|_{D_\mu}^2
\;+\; \Big(1+\tfrac{1}{\mu\gamma_k}\Big)\,C(\beta_k^2+\alpha_k^2)
\;+\; \gamma_k^2 \sigma_z^2\\
&= \big(1 - \mu \gamma_k + C \gamma_k^2\big)\,\|e_k^z\|_{D_\mu}^2
\;+\; C(\beta_k^2+\alpha_k^2) + \tfrac{C}{\mu}\tfrac{\beta_k^2+\alpha_k^2}{\gamma_k}
\;+\; \gamma_k^2 \sigma_z^2,
\end{align*}
where we expanded $(1+\mu\gamma_k)(1 - 2\mu \gamma_k + C \gamma_k^2)$ and absorbed higher-order
terms into $C\gamma_k^2$.
\[
(1+\mu\gamma_k)\big(1 - 2\mu \gamma_k + C \gamma_k^2\big)
= 1 \;-\; \mu\gamma_k \;+\; (C-2\mu^2)\gamma_k^2 \;+\; C\mu\,\gamma_k^3
\ \le\ 1 \;-\; \mu\gamma_k \;+\; C\gamma_k^2,
\]
for a (possibly larger) constant $C<\infty$ (since $\gamma_k^3\le \gamma_k^2$ for all $k$ large
enough and we can absorb finitely many initial indices into $C$). Thus there exists $c\in(0,\mu]$
such that
\begin{equation}
\label{eq:coef-simplified}
(1+\mu\gamma_k)\big(1 - 2\mu \gamma_k + C \gamma_k^2\big)\ \le\ 1 - c\gamma_k + C\gamma_k^2.
\end{equation}

Keep the squared step-size term as is and bound the factor:
\[
\Big(1+\tfrac{1}{\mu\gamma_k}\Big)\,C(\beta_k^2+\alpha_k^2)
\ \le\ C(\beta_k^2+\alpha_k^2)\;+\;\frac{C}{\mu}\,\frac{\beta_k^2+\alpha_k^2}{\gamma_k}.
\]
Under the timescale separation $\beta_k/\gamma_k\to 0$ and $\alpha_k/\gamma_k\to 0$, we
have $(\beta_k^2+\alpha_k^2)/\gamma_k = o(\gamma_k)$; hence, for all sufficiently large $k$,
\[
\frac{\beta_k^2+\alpha_k^2}{\gamma_k}\ \le\ C_1\,\gamma_k
\ \le\ C_2\,\gamma_k^2 \ +\ C_3\,(\beta_k^2+\alpha_k^2),
\]
where the last inequality holds after (i) possibly enlarging constants to account for a finite
prefix of indices, and (ii) using that $\beta_k^2+\alpha_k^2=O(\gamma_k^2)$ under separation.
Absorbing constants, we obtain the per–step bound
\begin{equation}
\label{eq:target-motion-final}
\Big(1+\tfrac{1}{\mu\gamma_k}\Big)\,C(\beta_k^2+\alpha_k^2)
\ \le\ C\,\gamma_k^2 \;+\; C\,(\beta_k^2+\alpha_k^2).
\end{equation}

Now combining \eqref{eq:cond-master}, \eqref{eq:target-motion}, and the Young bound, and taking $\varepsilon>0$ small so that
$1-2\mu\gamma_k\mapsto 1-c\gamma_k$ for some $c>0$ (absorbing constants), we obtain
\begin{equation} \label{eq:fast_error_final}
    \mathbb E\|e_{k+1}^z\|_{D_\mu}^2
\ \le\ \big(1 - c\gamma_k + C\gamma_k^2\big)\,\mathbb E\|e_k^z\|_{D_\mu}^2
\ +\ C\gamma_k^2 \ +\ C(\beta_k^2+\alpha_k^2).
\end{equation}

To apply Robbins--Siegmund to \eqref{eq:fast_error_final} we set
\[
V_k := \mathbb{E}\|e_k^z\|_{D_\mu}^2, 
\quad a_k := c\gamma_k - C\gamma_k^2, 
\quad b_k := C\gamma_k^2 + C(\beta_k^2+\alpha_k^2),
\]
so that
\[
V_{k+1} \;\le\; (1-a_k)V_k + b_k.
\]
Assumption \textbf{A5} (namely, $\sum_k \gamma_k = \infty$ and 
$\sum_k (\gamma_k^2 + \alpha_k^2 + \beta_k^2) < \infty$) implies 
$\sum_k a_k = \infty$ and $\sum_k b_k < \infty$, with $a_k \in [0,1]$ 
for large $k$. Robbins--Siegmund then yields that $V_k$ converges and 
$\sum_k a_k V_k < \infty$.

Since 
\[
a_k \;=\; c\gamma_k - C\gamma_k^2 
\;=\; c\gamma_k\Bigl(1 - \tfrac{C}{c}\gamma_k\Bigr),
\]
and by Assumption~\textbf{A5} we have $\gamma_k \to 0$, there exists $K$ such that 
$\gamma_k \le \tfrac{c}{2C}$ for all $k \ge K$. 
Thus, for $k \ge K$,
\[
1 - \tfrac{C}{c}\gamma_k \;\ge\; \tfrac{1}{2},
\qquad\text{so}\qquad
a_k \;\ge\; \tfrac{c}{2}\gamma_k.
\]

Since $a_k \ge \tfrac{c}{2}\gamma_k$ for all sufficiently large $k$ and 
$\sum_{k=0}^\infty \gamma_k = \infty$, it follows that
\[
\sum_{k=0}^\infty a_k = \infty.
\]
Suppose, for contradiction, that $\lim_{k\to\infty} V_k = L > 0$. 
Then there exists $K$ such that $V_k \ge \tfrac{L}{2}$ for all $k \ge K$. 
Hence
\[
\sum_{k=0}^\infty a_k V_k 
\;\ge\; \sum_{k=K}^\infty a_k \,\tfrac{L}{2}
\;=\; \tfrac{L}{2}\sum_{k=K}^\infty a_k 
\;=\; \infty,
\]
which contradicts the Robbins--Siegmund conclusion that 
$\sum_k a_k V_k < \infty$. 
Therefore $L=0$, i.e.\ $V_k \to 0$.
 Finally, the standard martingale argument (Doob + Borel–Cantelli) implies $\|e_k^z\|_{D_\mu}\to 0$ almost surely~\citep{duflo2013random}.

\end{proof}

\begin{lemma}[Medium tracking of the inner-level (in-sim)]
\label{lem:medium-tracking-3ts}
Under \textbf{A1}, \textbf{A2}, \textbf{A4}--\textbf{A7}, and Lemma~\ref{lem:fast-tracking-3ts}, we get:
\[
\sum_{k=0}^\infty \beta_k\,\mathbb E\|e_k^\phi\|^2<\infty
\quad\Longrightarrow\quad
e_k^\phi\to 0 \quad \text{a.s.}
\]
\end{lemma}

\begin{proof}

Work in the Euclidean norm on $\Phi$. Let
\[
e_k^\phi:=\phi_k-\phi^\star(\theta_k),\quad
h(\phi,\theta):=\hat\varphi(\phi,\theta;z^{\!}(\phi,\theta)).
\]
Define
\[
D_k^\phi:=h(\phi_k,\theta_k)-h^\star(\phi^\star(\theta_k),\theta_k),
\qquad
\delta_k^z:=\hat\varphi(\phi_k,\theta_k;z_k)-\hat\varphi(\phi_k,\theta_k;z_k^\star),
\qquad
\Delta_{k+1}^\phi:=\phi^\star(\theta_{k+1})-\phi^\star(\theta_k).
\]
From \textbf{A4} (inner-layer stability), there exists $m>0$ such that, for $k$ large enough (iterates in a compact neighborhood by projection \textbf{A7}),
\begin{equation}
\label{eq:inner-stability}
\langle e_k^\phi,\, D_k^\phi\rangle \le -\,m\,\|e_k^\phi\|^2,
\qquad
\|D_k^\phi\|\le L_h \|e_k^\phi\| \quad \text{(local Lipschitzness from \textbf{A2})}.
\end{equation}
By Lipschitzness of $\hat\varphi$ in $z$ on compacts (\textbf{A2}),
\begin{equation}
\label{eq:delta-z-bound}
\|\delta_k^z\| \le L_z \|z_k - z_k^\star\| = L_z \|e_k^z\|.
\end{equation}
By \textbf{A1} (IFT) $\phi^\star\in C^1$, thus Lipschitz on compact $\Theta$:
\begin{equation}
\label{eq:phi-star-motion}
\|\Delta_{k+1}^\phi\|
\le L_{\phi^\star}\,\|\theta_{k+1}-\theta_k\|.
\end{equation}

\medskip

Using nonexpansiveness of $\Pi_\Phi$ and $\phi^\star(\theta_{k+1})\in\Phi$,
\begin{align}
\|e_{k+1}^\phi\|
&= \big\|\Pi_\Phi\big(\phi_k+\beta_k[\hat\varphi(\phi_k,\theta_k;z_k)+M^{(\phi)}_{k+1}]\big)-\phi^\star(\theta_{k+1})\big\|
\nonumber\\
&\le \Big\| \phi_k + \beta_k\big(h^\star(\phi_k,\theta_k) + \delta_k^z + M^{(\phi)}_{k+1}\big) - \phi^\star(\theta_{k+1}) \Big\|
\nonumber\\
&= \Big\| e_k^\phi + \beta_k D_k^\phi + \beta_k \delta_k^z + \beta_k M^{(\phi)}_{k+1} - \Delta_{k+1}^\phi \Big\|.
\label{eq:med-one-step}
\end{align}

Squaring \eqref{eq:med-one-step} and taking $\mathbb E[\cdot\mid\mathcal F_k]$, the martingale-difference property (\textbf{A6}) gives
\[
\mathbb E\langle e_k^\phi + \beta_k D_k^\phi + \beta_k \delta_k^z - \Delta_{k+1}^\phi,\ M^{(\phi)}_{k+1}\mid \mathcal F_k\rangle = 0,
\quad
\mathbb E\|M^{(\phi)}_{k+1}\|^2\mid \mathcal F_k \le \sigma_\phi^2.
\]
Hence
\begin{align}
\mathbb E\!\left[\|e^\phi_{k+1}\|^2\mid\mathcal F_k\right]
&\le \|e_k^\phi+\beta_k D_k^\phi - \Delta_{k+1}^\phi\|^2
\ +\ \beta_k^2\,\mathbb E\|\delta_k^z\|^2
\ +\ \beta_k^2\,\sigma_\phi^2.
\label{eq:med-master-cond}
\end{align}

Expanding and applying \eqref{eq:inner-stability}:
\begin{align}
\|e_k^\phi+\beta_k D_k^\phi\|^2
= \|e_k^\phi\|^2 + 2\beta_k \langle e_k^\phi, D_k^\phi\rangle + \beta_k^2 \|D_k^\phi\|^2
\le \big(1 - 2m\beta_k + L_h^2 \beta_k^2\big)\,\|e_k^\phi\|^2.
\label{eq:med-drift-contract}
\end{align}

Using Young’s inequality as in the fast-layer proof (Step~4 there), for any $\varepsilon\in(0,1)$,
\[
-2\langle e_k^\phi+\beta_k D_k^\phi,\ \Delta_{k+1}^\phi\rangle
\ \le\ \varepsilon\,\|e_k^\phi+\beta_k D_k^\phi\|^2\ +\ \frac{1}{\varepsilon}\,\|\Delta_{k+1}^\phi\|^2.
\]
By \eqref{eq:phi-star-motion} and the projected slow step (\textbf{A7}) with bounded second moments (\textbf{A2}),
\begin{equation}
\label{eq:Delta-phi-second-moment}
\mathbb E\|\Delta_{k+1}^\phi\|^2
\le L_{\phi^\star}^2\,\mathbb E\|\theta_{k+1}-\theta_k\|^2
\le C\,\alpha_k^2.
\end{equation}
By \eqref{eq:delta-z-bound} and boundedness of $\|e_k^z\|$ on $\mathcal Z$ (projection),
\begin{equation}
\label{eq:delta-z-second-moment}
\mathbb E\|\delta_k^z\|^2 \le C\,\mathbb E\|e_k^z\|^2 \le C.
\end{equation}

Plugging \eqref{eq:med-drift-contract} and the Young bound into \eqref{eq:med-master-cond}, and choosing $\varepsilon>0$ small, we obtain
\[
\mathbb E\!\left[\|e^\phi_{k+1}\|^2\mid\mathcal F_k\right]
\ \le\ \big(1 - c\,\beta_k + C\beta_k^2\big)\,\|e_k^\phi\|^2
\ +\ C\,\beta_k^2\,\mathbb E\|\delta_k^z\|^2
\ +\ C\,\mathbb E\|\Delta_{k+1}^\phi\|^2
\ +\ C\,\beta_k^2,
\]
for some $c\in(0,1)$ and finite $C$ (uniform on the compact set).
Now apply \eqref{eq:delta-z-second-moment} and \eqref{eq:Delta-phi-second-moment}:
\[
\mathbb E\|e^\phi_{k+1}\|^2
\ \le\ (1 - c\,\beta_k + C\beta_k^2)\,\mathbb E\|e_k^\phi\|^2
\ +\ C\,\beta_k^2
\ +\ C\,\alpha_k^2
\ +\ C\,\beta_k^2\,\mathbb E\|e_k^z\|^2.
\]
Let $V_k:=\mathbb E\|e_k^\phi\|^2$, $a_k:=c\beta_k - C\beta_k^2$, and $b_k:=C\beta_k^2(1+\mathbb E\|e_k^z\|^2)+C\alpha_k^2$.
By \textbf{A5}, $\sum_k a_k=\infty$ and $\sum_k b_k<\infty$ (since $\sum \beta_k^2,\sum \alpha_k^2<\infty$ and $\|e_k^z\|$ is bounded by projection).
Robbins--Siegmund implies $V_k$ converges and $\sum_k a_k V_k<\infty$.
As $a_k\ge \tfrac{c}{2}\beta_k$ for large $k$ and $\sum_k \beta_k=\infty$, we get $\lim_{k \to \infty} V_k = 0$.

Finally, the assumption $\sum_k \beta_k\,\mathbb E\|e_k^\phi\|^2<\infty$ together with the boundedness of $\{e_k^\phi\}$ 
standardly yields $e_k^\phi\to 0$ almost surely (see the same martingale/Borel--Cantelli argument used after Lemma~\ref{lem:fast-tracking-3ts}).
\end{proof}

\begin{lemma}[Convergence of outer-level with slow timescale]
\label{lem:slow-3ts}
Define the drift term of the slow ODE
\[
g(\theta)\;:=\;\varphi\big(\phi^\star(\theta),\theta;\,z^{\!}(\phi^\star(\theta),\theta)\big)
\;=\;\nabla_\theta F(\theta),
\]
and the actual estimator
\[
\widehat g_k\;:=\;\varphi(\phi_k,\theta_k;z_k)\;+\;M^{(\theta)}_{k+1}
\;=\;g(\theta_k)\;+\;b_k\;+\;M^{(\theta)}_{k+1},
\]
with the bias
\[
b_k \;:=\; \varphi(\phi_k,\theta_k;z_k) - \varphi(\phi^\star(\theta_k),\theta_k;z^{\!}(\phi^\star(\theta_k),\theta_k)).
\]
Under \textbf{A1}--\textbf{A7} and Lemmas~\ref{lem:fast-tracking-3ts}--\ref{lem:medium-tracking-3ts}:
\begin{enumerate}
\item $\|b_k\|\le C\big(\|e^\phi_k\|+\|e^z_k\|\big)$ and hence $\displaystyle \sum_k \alpha_k \|b_k\|^2 < \infty$ a.s.;
\item the projected slow update $\theta_{k+1}=\Pi_\Theta\big(\theta_k+\alpha_k \widehat g_k\big)$ satisfies
\[
F(\theta_k)\ \text{converges a.s.},\qquad
\sum_{k=0}^\infty \alpha_k\,\|\nabla_\theta F(\theta_k)\|^2<\infty\ \text{a.s.},
\]
and every limit point of $\{\theta_k\}$ lies in $\mathcal E:=\{\theta:\nabla_\theta F(\theta)=0\}$.
\end{enumerate}
\end{lemma}

\begin{proof}
We work in the Euclidean norm on $\Theta$. On compact $\Theta$, $\nabla F$ is $L$-Lipschitz (by \textbf{A1}–\textbf{A2}).
$M^{(\theta)}_{k+1}$ is a martingale difference with $\mathbb E[\|M^{(\theta)}_{k+1}\|^2\mid\mathcal F_k]\le \sigma_\theta^2$ (\textbf{A6}).

\smallskip
Local Lipschitzness of $\varphi$ in $(\phi,z)$ on the compact $\Phi\times\Theta\times\mathcal Z$ (\textbf{A2}) gives
\[
\|b_k\|
\ \le\ L_\varphi\Big(\|\phi_k-\phi^\star(\theta_k)\| + \|z_k - z^{\!}(\phi^\star(\theta_k),\theta_k)\|\Big)
\ \le\ C\big(\|e^\phi_k\|+\|e^z_k\|\big).
\]
Therefore $\|b_k\|^2 \le C\big(\|e^\phi_k\|^2+\|e^z_k\|^2\big)$.

\smallskip
From Lemma~\ref{lem:medium-tracking-3ts}: $\sum_k \beta_k\,\mathbb E\|e^\phi_k\|^2<\infty$ a.s.  
From Lemma~\ref{lem:fast-tracking-3ts}: $\sum_k \gamma_k\,\mathbb E\|e^z_k\|^2<\infty$ a.s.
Since $\alpha_k/\beta_k\to 0$ and $\alpha_k/\gamma_k\to 0$ (\textbf{A5}), there exist $c_\phi,c_z$ and $K$ s.t. for all $k\ge K$,
$\alpha_k\le c_\phi \beta_k$ and $\alpha_k\le c_z \gamma_k$. Therefore, it proves (i).
\[
\sum_{k=0}^\infty \alpha_k\,\mathbb E\|b_k\|^2
\ \le\ C\sum_k \alpha_k\big(\mathbb E\|e^\phi_k\|^2+\mathbb E\|e^z_k\|^2\big)
\ <\ \infty \quad\text{a.s.}
\]

\smallskip
By projected smoothness (same inequality as used in Lemma~\ref{lem:medium-tracking-3ts}):
\[
F(\theta_{k+1})
\ \ge\ F(\theta_k) + \alpha_k\,\langle \nabla F(\theta_k),\, \widehat g_k\rangle
\ -\ \tfrac{L}{2}\,\alpha_k^2 \|\widehat g_k\|^2.
\]
Take $\mathbb E[\cdot\mid \mathcal F_k]$ and use $\mathbb E[M^{(\theta)}_{k+1}\mid\mathcal F_k]=0$:
\[
\mathbb E\!\left[F(\theta_{k+1})\mid\mathcal F_k\right]
\ \ge\ F(\theta_k)
+\alpha_k\,\langle \nabla F(\theta_k),\, g(\theta_k)\rangle
+\alpha_k\,\langle \nabla F(\theta_k),\, b_k\rangle
-\tfrac{L}{2}\alpha_k^2\,\mathbb E\|g(\theta_k)+b_k+M^{(\theta)}_{k+1}\|^2.
\]
But $g(\theta_k)=\nabla F(\theta_k)$, hence the main progress term is $\alpha_k\|\nabla F(\theta_k)\|^2$.
Also 
\[
\mathbb E\|g(\theta_k)+b_k+M^{(\theta)}_{k+1}\|^2 \le C\big(\|\nabla F(\theta_k)\|^2+\|b_k\|^2+\sigma_\theta^2\big)\,.
\]

\smallskip
By Young’s inequality (same as in the medium-layer proof),
for $\eta=\tfrac{1}{2}$,
\[
\alpha_k\langle \nabla F(\theta_k), b_k\rangle
\ \ge\ -\,\frac{\alpha_k}{4}\,\|\nabla F(\theta_k)\|^2 \ -\ \alpha_k\,\|b_k\|^2.
\]
Absorb $C\alpha_k^2\|\nabla F(\theta_k)\|^2$ into the main term for large $k$ (since $\alpha_k\to 0$).
Thus there exist $k_0$ and $c\in(0,1)$ such that for all $k\ge k_0$,
\[
\mathbb E\!\left[F(\theta_{k+1})\mid\mathcal F_k\right]
\ \ge\ F(\theta_k)
\ +\ c\,\alpha_k\,\|\nabla F(\theta_k)\|^2
\ -\ \alpha_k\,\|b_k\|^2
\ -\ C\,\alpha_k^2.
\]

\smallskip
Define the almost-supermartingale
\[
X_k\ :=\ F(\theta_k) - \sum_{i=0}^{k-1}\big(\alpha_i\|b_i\|^2 + C\alpha_i^2\big).
\]
Since $\sum_k \alpha_k\|b_k\|^2<\infty$ a.s., and $\sum_k \alpha_k^2<\infty$ (\textbf{A5}),
so $\sum_k ( \alpha_k\|b_k\|^2 + C\alpha_k^2)<\infty$ a.s.
Hence
\[
\mathbb E[X_{k+1}\mid\mathcal F_k] \ \ge\ X_k \ +\ c\,\alpha_k\,\|\nabla F(\theta_k)\|^2.
\]
Robbins--Siegmund implies: (i) $X_k$ converges a.s., so $F(\theta_k)$ converges a.s.;
(ii) $\sum_k \alpha_k \|\nabla F(\theta_k)\|^2<\infty$ a.s.  
Therefore $\liminf_k \|\nabla F(\theta_k)\|=0$, and any limit point $\theta^\star$ satisfies
$\nabla F(\theta^\star)=0$ by continuity of $\nabla F$. This proves (ii).
\end{proof}

\begin{theorem}[Three–timescale bi-level SPG: direct almost-sure convergence]
\label{thm:3ts-main}
Assume \textbf{A1}–\textbf{A7} and let the iterates $(z_k,\phi_k,\theta_k)$ be generated by
the three–timescale recursions in \eqref{eq:fast_layer}–\eqref{eq:slow_layer}. Define the fast- and
medium-layer errors
\[
e_k^z\ :=\ z_k - z^\star(\phi_k,\theta_k),
\qquad
e_k^\phi\ :=\ \phi_k - \phi^\star(\theta_k),
\]
and the slow-layer stationary set
\[
\mathcal E\ :=\ \bigl\{\theta\in\Theta:\ \nabla_\theta F(\theta)=0\bigr\},
\qquad
F(\theta)=J_{\rm real}\!\big(\pi_{\phi^\star(\theta)}\big).
\]
Then, almost surely,
\[
e_k^z \longrightarrow 0,
\qquad
e_k^\phi \longrightarrow 0,
\qquad
\mathrm{dist}\bigl(\theta_k,\mathcal E\bigr) \longrightarrow 0.
\]
Equivalently, every limit point of the triple $(z_k,\phi_k,\theta_k)$ belongs to
\[
\Bigl\{\bigl(z^\star(\phi^\star(\theta),\theta),\ \phi^\star(\theta),\ \theta\bigr)\ :\ \theta\in\mathcal E\Bigr\}.
\]
\end{theorem}

\begin{proof}
By Lemma~\ref{lem:fast-existence-stability}, the fast ODE has a unique, globally
asymptotically stable equilibrium $z^\star(\phi,\theta)$; Lemma~\ref{lem:fast-tracking-3ts}
implies $e_k^z\to 0$ a.s. Using this and the inner-layer stability in \textbf{A4},
Lemma~\ref{lem:medium-tracking-3ts} yields $e_k^\phi\to 0$ a.s. Finally, with
$z_k\approx z^\star(\phi_k,\theta_k)$ and $\phi_k\approx \phi^\star(\theta_k)$, the slow drift
equals $\nabla_\theta F(\theta_k)$ up to a summable bias, and
Lemma~\ref{lem:slow-3ts} gives $\mathrm{dist}(\theta_k,\mathcal E)\to 0$ a.s.
\end{proof}

\begin{corollary}[Two timescales (critic exact or faster)]
    If $z_k\equiv z_k^\star$ (exact) or the fast layer is strictly faster so that
$e_k^z\to 0$ with summable bias, Lemma~\ref{lem:fast-tracking-3ts} is obvious and
Theorem~\ref{thm:3ts-main} follows from Lemmas~\ref{lem:medium-tracking-3ts}--\ref{lem:slow-3ts}.
\end{corollary}

\begin{corollary}[Nested (inner exact at each outer step)]
    If $z_k=z^{\!}(\phi^\star(\theta_k),\theta_k)$ and $\phi_k=\phi^\star(\theta_k)$, then
$\theta_{k+1}=\Pi_\Theta(\theta_k+\alpha_k[\nabla_\theta F(\theta_k)+M^{(\theta)}_{k+1}])$,
and the standard single-level SPG convergence to $\mathcal E$ follows from
Lemma~\ref{lem:slow-3ts}.

\end{corollary}

\section{Efficient Adjoint Computation of the Outer Gradient}
\label{appendix:adjoint_ift}

This appendix provides an equivalent but more computationally efficient way to compute the outer gradient in the bi-level \ac{rl} formulation of Section~\ref{sec:bi_level_rl} (cf. Eq.~\eqref{eq:bi_level_rl_aprox}--Eq.~\eqref{eq:real_world_spg}). The key idea is to avoid forming the full Jacobian $\nabla_\theta \phi(\theta)$ (or any dense $d_\phi \times d_\theta$ object) and instead compute the required scalar outer gradient via a single \emph{adjoint} linear solve. This adjoint form is mathematically equivalent to the \acf{ift}-based sensitivity expression in~\eqref{eq:ift}.

Recall the in-sim \ac{spg} convergence (Eq.~\eqref{eq:inner_pg})
\begin{equation}
\hat{\varphi}(\phi,\theta)
:=\mathbb{E}_{s\sim \hat{\rho}_{\theta}^{\pi_{\phi}}}
\Bigl[\nabla_{\phi}\log\pi_{\phi}(a|s)\,Q_{\theta}^{\pi_{\phi}}(s,a)\Bigr],
\qquad
\hat{\varphi}(\phi,\theta)=0.
\label{eq:app_residual_def}
\end{equation}
Under the local regularity assumptions stated in Section~\ref{sec:bi_level_rl} (local isolation and nonsingularity of $\nabla_\phi \hat{\varphi}$), the stationarity condition implicitly defines a locally unique differentiable solution branch $\phi=\phi(\theta)$.

Let the outer objective be the real-world return~\eqref{eq:bi_level_rl_aprox}, written abstractly as a scalar function
\begin{equation}
\mathcal{L}(\theta)
:= \mathbb{E}_{\tau\sim \pi_{\phi(\theta)},\,\mathcal{P}}
\Bigl[\sum_{t=0}^{\infty}\gamma^t r(s_t,a_t)\Bigr].
\label{eq:app_outer_objective}
\end{equation}
This outer objective may, in general, have both (i) \emph{implicit dependence} through $\phi(\theta)$ and (ii) \emph{explicit dependence} on $\theta$ (e.g., if additional regularizers are added to the outer level). We keep both possibilities in the derivation.

\subsection*{Avoiding forming $\nabla_\theta \phi$ for scalar objectives}
Differentiate $\mathcal{L}(\theta)$ using the chain rule:
\begin{equation}
\nabla_\theta \mathcal{L}(\theta)
=
\underbrace{\partial_\theta \mathcal{L}(\phi,\theta)}_{c(\theta)}
+
\underbrace{\bigl(\nabla_\theta \phi(\theta)\bigr)^{\top}
\nabla_\phi \mathcal{L}(\phi,\theta)}_{(\nabla_\theta \phi)^\top g}.
\label{eq:app_chain_rule}
\end{equation}
Let's define the following objects evaluated at $(\phi,\theta)$:
\begin{equation}
A(\phi,\theta) := \nabla_\phi \hat{\varphi}(\phi,\theta)\in\mathbb{R}^{d_\phi\times d_\phi},
\qquad
B(\phi,\theta) := \nabla_\theta \hat{\varphi}(\phi,\theta)\in\mathbb{R}^{d_\phi\times d_\theta},
\label{eq:app_A_B_def}
\end{equation}
\begin{equation}
g(\phi,\theta):=\nabla_\phi \mathcal{L}(\phi,\theta)\in\mathbb{R}^{d_\phi},
\qquad
c(\phi,\theta):=\partial_\theta \mathcal{L}(\phi,\theta)\in\mathbb{R}^{d_\theta}.
\label{eq:app_g_c_def}
\end{equation}
From the \acf{ift}~\eqref{eq:ift}, we have
\begin{equation}
\nabla_\theta \phi(\theta) = -A(\phi,\theta)^{-1} B(\phi,\theta).
\label{eq:app_ift_basic}
\end{equation}
Substituting~\eqref{eq:app_ift_basic} into~\eqref{eq:app_chain_rule} yields
\begin{equation}
\nabla_\theta \mathcal{L}(\theta)
=
c(\phi,\theta)
-
B(\phi,\theta)^\top A(\phi,\theta)^{-\top} g(\phi,\theta).
\label{eq:app_outer_via_transpose_inverse}
\end{equation}
Therefore, to compute the scalar outer gradient, we do \emph{not} need the full matrix $\nabla_\theta \phi(\theta)$; we only need the vector $A^{-\top}g$.

\subsection*{Adjoint formulation}
Introduce the adjoint variable $\lambda\in\mathbb{R}^{d_\phi}$ defined as the solution of the linear system
\begin{equation}
A(\phi,\theta)^\top \lambda = g(\phi,\theta).
\label{eq:app_adjoint_system}
\end{equation}
Then~\eqref{eq:app_outer_via_transpose_inverse} becomes
\begin{equation}
\boxed{
\nabla_\theta \mathcal{L}(\theta)
=
c(\phi,\theta)
-
B(\phi,\theta)^\top \lambda.
}
\label{eq:app_adjoint_gradient}
\end{equation}
In the common case where the outer objective depends on $\theta$ only through the deployed policy $\pi_{\phi(\theta)}$ (so that $c(\phi,\theta)\approx 0$ by construction), we obtain the simplified expression
\begin{equation}
\boxed{
\nabla_\theta \mathcal{L}(\theta)
=
-
B(\phi,\theta)^\top \lambda,
\qquad
A(\phi,\theta)^\top \lambda = g(\phi,\theta).
}
\label{eq:app_adjoint_gradient_no_direct}
\end{equation}

\eqref{eq:app_adjoint_gradient} is the most efficient reverse-mode form of the bi-level gradient for scalar outer objectives: it avoids constructing any dense $d_\phi\times d_\theta$ sensitivity matrix and requires only (i) one linear solve in $d_\phi$ dimensions and (ii) one vector--Jacobian product to obtain $B^\top \lambda$.

\subsection*{Obtaining $A^\top v$ and $B^\top v$ without forming $A$ or $B$}
The adjoint system~\eqref{eq:app_adjoint_system} and the gradient formula~\eqref{eq:app_adjoint_gradient} only require the ability to apply the linear operators
\[
v \mapsto A(\phi,\theta)^\top v,
\qquad
v \mapsto B(\phi,\theta)^\top v.
\]
These can be computed without forming $A$ or $B$ explicitly using reverse-mode automatic differentiation as \emph{vector--Jacobian products (VJPs)} applied to the residual $\hat{\varphi}(\phi,\theta)$ in~\eqref{eq:app_residual_def}. 

Given a vector $v\in\mathbb{R}^{d_\phi}$:
\begin{equation}
A(\phi,\theta)^\top v
=
\nabla_\phi\Bigl(\langle \hat{\varphi}(\phi,\theta), v\rangle\Bigr),
\qquad
B(\phi,\theta)^\top v
=
\nabla_\theta\Bigl(\langle \hat{\varphi}(\phi,\theta), v\rangle\Bigr).
\label{eq:app_vjp_identities}
\end{equation}
Thus, iterative Krylov solvers (e.g., GMRES/BiCGSTAB) can be used to solve~\eqref{eq:app_adjoint_system} using only the operator $v\mapsto A^\top v$.

Additionally the operator $A=\nabla_\phi \hat{\varphi}$ need not be symmetric or positive definite. Therefore, a nonsymmetric Krylov method (e.g., GMRES or BiCGSTAB) is generally appropriate. Damping can be used for numerical stability by solving
\begin{equation}
\bigl(A(\phi,\theta)^\top + \lambda I\bigr)\lambda = g(\phi,\theta),
\qquad \lambda>0.
\label{eq:app_damped_adjoint}
\end{equation}

The adjoint form in~\eqref{eq:app_adjoint_gradient} is fully consistent with Theorem~\ref{Th:bi_level_spg} while avoiding the computational burden of forming $\nabla_\theta \phi$ explicitly. In particular,
\[
B(\phi,\theta)^\top \lambda
=
\Bigl(\nabla_\theta \hat{\varphi}(\phi,\theta)\Bigr)^\top \lambda
\]
 can be estimated using the same trajectory-based estimators used for Theorem~\ref{Th:bi_level_spg} (e.g., Lemma~\ref{lem:mc_sensitivity} and Lemma~\ref{lem:policy_eval_sensitivity}).

\begin{figure}[h]
  \centering
  \includegraphics[width=0.95\linewidth]{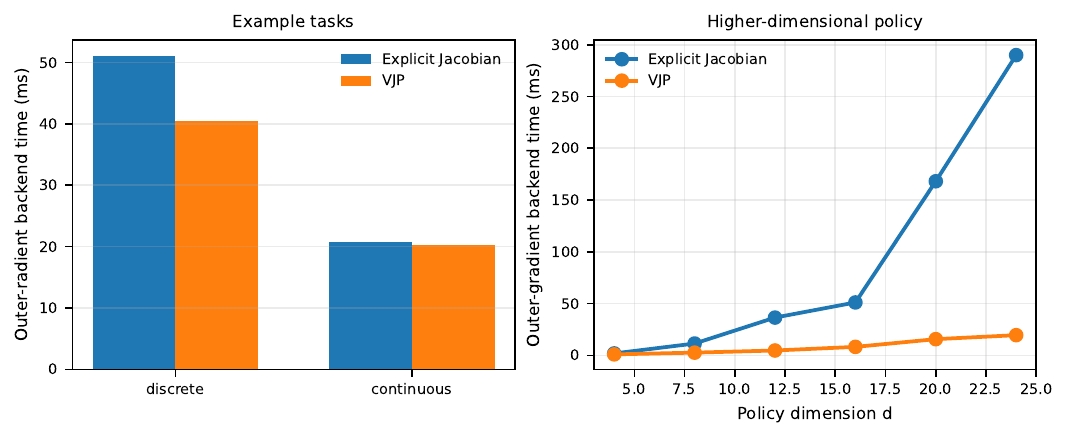}
\caption{Comparison of computational time between the proposed VJP-based method and explicit Jacobian computation. The left panel shows results for low-dimensional problems (a 3-state discrete MDP and a 1-dimensional continuous MDP), while the right panel reports results for a higher-dimensional \ac{nn} policy.}
  \label{fig:vjp-comparison}
\end{figure}
\subsection*{Evaluation}
A comparison between the VJP approach and explicitly forming the Jacobian $\nabla_\theta \phi$ using a PyTorch implementation is shown in Fig.~\ref{fig:vjp-comparison}. For the low-dimensional discrete task and the continuous MDP with a \ac{nn} policy containing six parameters used in Section~\ref{sec:examples}, the difference in computational time is not substantial due to the simplicity of the policies. However, as the size of the policy network increases, the computational advantage becomes significant. For \ac{nn} policies with more than 20 parameters, the VJP-based method is more than two orders of magnitude faster than explicitly computing the Jacobian. The reported timings focus only on the outer-gradient computation after trajectory batches have already been generated. Since the rollout data are shared between both methods, excluding rollout time isolates the computational cost of the gradient backend and highlights the efficiency gain from avoiding explicit Jacobian construction.

% For scalar outer objectives, the most efficient \acf{ift}-based bi-level gradient computation is obtained by the adjoint system~\eqref{eq:app_adjoint_system} and the gradient formula~\eqref{eq:app_adjoint_gradient}. This avoids forming $\nabla_\theta \phi$ or any dense sensitivity matrices and fits naturally with reverse-mode automatic differentiation via the VJP identities in Eq.~\eqref{eq:app_vjp_identities}.

\section{Details on Illustrative Examples}\label{appendix:experiments}
\subsection{Example 1: Discrete MDP}

We considered a finite MDP with state space \(\mathcal S=\{0,1,2\}\) and action space \(\mathcal A=\{0,1\}\). The real and model MDP parameters have the following form:

{Transition logits}
\[
\theta_\text{trans}[s,a] \;=\; \bigl[\theta_\text{trans}[s,a,0],\,\theta_\text{trans}[s,a,1],\,\theta_\text{trans}[s,a,2]\bigr]
\]
so that 
\[
P(s'\!\mid s,a)=\mathrm{softmax}\bigl(\theta_\text{trans}[s,a]\bigr)_{s'}.
\]
The real‐MDP transition logits are
\[
\theta_\text{real}^f = 
\bigl\{\,
\underbrace{[\,0.5,2.0,0.5],\,[\,1.0,1.5,0.5]}_{s=0},\;
\underbrace{[\,1.0,1.0,1.0],\,[\,1.5,1.0,0.5]}_{s=1},\;
\underbrace{[\,0.5,1.0,0.1],\,[\,1.0,0.5,1.0]}_{s=2}
\,\bigr\}.
\]

\[
R(s,a)=\theta^R[s,a],
\]
with
\[
\theta_\text{real}^R = 
\begin{bmatrix}
1.0 & 0.5\\
0.0 & 3.0\\
0.01& 2.0
\end{bmatrix}.
\]
A similar structure holds for the simulation MDP, where the simulation model and reward parameters for the trials shown in Fig.~\ref{fig:results} were initialized at random in $[0,5]$.

The in-sim policy is solved using \ac{dp} with a convergence threshold of $1\times 10^{-2}$, which provides a deterministic policy, which is then converted into a stochastic policy. After solving the discrete MDP by soft value iteration, we obtain a state–action value table
\[
Q^\star(s,a)
\;=\;
R(s,a) \;+\;\gamma\,\sum_{s'}P(s'\mid s,a)\,V^\star(s')\,, 
\]
where \(P(s'\!\mid s,a)\) and \(R(s,a)\) come from the MDP parameters.  A  greedy policy would be
\[
\pi_{\mathrm{det}}(s) = \arg\max_{a} Q^\star(s,a)\,,
\]
but this is non-differentiable.  Instead, we define a \emph{soft} (stochastic) policy
\[
\pi(a \mid s)
\;=\;
\frac{\exp\bigl(Q^\star(s,a)/\tau\bigr)}
     {\sum_{b}\exp\bigl(Q^\star(s,b)/\tau\bigr)}\,,
\]
where \(\tau>0\) is a temperature parameter controlling exploration, which we set as 2.0.
Because every entry of \(\pi(\cdot\mid s)\) is a smooth function of the underlying \(Q^\star\), and \(Q^\star\) itself depends differentiably on the simulation parameters \(\theta\), this construction makes the whole inner-level fully differentiable.

Once \(Q^\star\) is computed, we set the policy logits as:
\[
\text{logits}(s,a) \;=\;\log\bigl(\pi(a\mid s)\bigr)
\quad\Longrightarrow\quad
\pi(a\mid s)=\mathrm{softmax}\bigl(\text{logits}(s,\cdot)\bigr)\,.
\]

Using an on-policy, in-sim trajectory of length 1000 drawn from the \ac{dp} solution, we compute the Markov-chain sensitivities (Eq.~\ref{eq:mc_pg_phi_sensitivity_expected_value_form} and~\ref{eq:mc_pg_theta_sensitivity_expected_value_form}) and the critic sensitivities (Eq.~\ref{eq:critic_sensitivity_theta} and~\ref{eq:critic_sensitivity_phi}). Each sensitivity table is represented in tabular form and updated iteratively until convergence with a threshold of $1\times 10^{-2}$.
For each outer-level iteration, we use a trajectory of length 1000 generated by applying the in-sim policy to the real \ac{mdp} to update the model and reward parameters using \ac{spg} steps according to Eq.~\eqref{eq:real_world_spg} with a learning rate of 0.1.

\subsection{Example 2: Continuous MDP}
We use a one-dimensional linear system with additive Gaussian noise and an exponential reward:
\begin{align}
s_{t+1} &= \theta_s\,s_t \;+\;\theta_a\,a_t \;+\;\epsilon_t,\quad \epsilon_t\sim\mathcal{N}(0,\sigma^2),\\
R_\theta(s_t,a_t) &= \exp\!\bigl(-\lambda\,[\,\theta_q\,s_t^2 + \theta_r\,a_t^2\,]\bigr).
\end{align}
The true system uses 
\(\theta_s=\theta_a=\theta_q=\theta_r=1\), 
while all five simulation parameters 
\(\{\theta_s,\theta_a,\theta_q,\theta_r\}\)
are initialized uniformly at random in \([0,1]\).  
 
By taking logarithms of the stage reward, we obtain an equivalent quadratic stage cost
\(\,c(s,a)=\lambda\,(q\,s^2 + r\,a^2)\), 
with \(q=\theta_q\), \(r=\theta_r\).  In the infinite-horizon, \(\gamma\)-discounted setting, the optimal control law is linear,
\[
a_t^\star = -K\,s_t,
\]
where the gain \(K\) and value-function curvature \(P\) solve the discrete Algebraic Riccati Equation (DARE)
\[
P = \lambda q + \gamma\,( \theta_s - \theta_a\,K )^2\,P,
\qquad
K = \frac{\theta_a\,P\,\theta_s}{r + \theta_a^2\,P}.
\]
Once \(P\) is found, the optimal value function is \(V^\star(s)=P\,s^2\) and the greedy policy is \(a=-K\,s\).

To test the robustness of our sensitivity estimates, we fit two small feed-forward \ac{nn} of one hidden layer of size 6 for the policy and 64 for the value function:
\[
V_\psi(s)\approx V^\star(s), 
\qquad
\hat\pi_\phi(a\mid s)\approx \delta(a+K\,s),
\]
where \(\hat\pi_\phi\) is trained to match the exact linear feedback as a stochastic Gaussian policy.  The learned \(V\) and \(\hat\pi\) are then used in place of the true LQR quantities in the sensitivity computations inner level.

Using an on-policy, in-sim trajectory of length 1000 drawn from the approximated \ac{nn} policy, we compute the Markov-chain sensitivities (Eq.~\ref{eq:mc_pg_phi_sensitivity_expected_value_form} and~\ref{eq:mc_pg_theta_sensitivity_expected_value_form}) and the critic sensitivities (Eq.~\ref{eq:critic_sensitivity_theta} and~\ref{eq:critic_sensitivity_phi}). Sensitivities are computed for each sample until convergence of $1\times 10^{-2}$.
For each outer-level iteration, we use 20 trajectories of length 200 generated by applying the in-sim \ac{nn} approximated policy to the real \ac{mdp} to update the model and reward parameters using \ac{spg} steps according to Eq.~\eqref{eq:real_world_spg} with a learning rate of 0.1.

\subsection{Example 3: Quadrotor}\label{app:quad}
To evaluate the proposed bi-level \ac{ppo} scheme in a nonlinear continuous-control setting, we consider the planar quadrotor (Quad2D) environment implemented in \texttt{safe-control-gym} \citep{yuan2022safe}. 
The objective is to stabilize the quadrotor at a desired reference position as quickly as possible while minimizing control effort.

The system state is defined as
\[
s = [x, \dot{x}, z, \dot{z}, \phi, \dot{\phi}]^T,
\]
where $(x,z)$ denote the planar position of the vehicle and $\phi$ is the pitch angle. 
The control inputs correspond to the thrusts generated by the two rotors,
\(
a = [u_1, u_2]^T.
\)

The simplified dynamics of the system are given by
\begin{align}
m\ddot{z} &= (u_1 + u_2)\cos\phi - mg, \\
m\ddot{x} &= -(u_1 + u_2)\sin\phi, \\
I_{yy}\ddot{\phi} &= l(u_2 - u_1),
\end{align}
where $m$ denotes the quadrotor mass, $l$ the arm length, and $I_{yy}$ the moment of inertia around the pitch axis. 
These dynamics capture the key nonlinear coupling between translational motion and attitude.

The reward function follows an exponential quadratic form similar to Example~2:
\begin{equation}
r_t = \exp\!\left(
- (s_t - s_{ref,t})^T W_s (s_t - s_{ref,t})
- (a_t - a_{ref,t})^T W_a (a_t - a_{ref,t})
\right),
\end{equation}
where $s_{ref,t}$ denotes the desired reference state, $a_{ref,t}$ is the hover thrust, and $W_s,\, W_a$ are the cost weights. 
This formulation encourages accurate tracking while penalizing excessive control effort.

For the outer task, the quadrotor mass and arm length are $m_0=0.033\,kg, l_0=0.039\,m$ and the cost weights are $W_{s}=diag([5,0.1,5,0.1,0.1,0.001]), W_{a} = diag([0.1, 0.1])$. 
For the in-sim task, the tunable simulation parameter is $\theta = [\theta_m, \theta_l, \theta_{ws}, \theta_{wa}]^T$ such that quadrotor mass $m = \theta_m m_0$, the arm length $l = \theta_l l_0$, and the cost weights $W_s = diag(\theta_{ws}), W_a=diag(\theta_{wa})$. The in-sim parameter $\theta$ is initialized as $\theta=[0.5, 1.0, 5,0.1,5,0.1,0.1,0.001, 0.1, 0.1]$. 

\subsubsection*{Bi-level PPO details}
Since no closed-form optimal policy exists for this nonlinear system, both the policy and value function are approximated using \acp{nn} trained with \ac{ppo}. 
The actor is implemented as a multi-layer perceptron with two hidden layers of $64$ units each, while the critic uses two hidden layers of $128$ units. 
Both networks use \emph{Tanh} activations.

The in-sim \ac{rl} performs standard \ac{ppo} updates in simulation to improve the policy under the current simulator parameter $\theta$. 
Each \ac{ppo} update uses $10$ optimization epochs with a learning rate of $3\times10^{-4}$. 
Additional hyperparameters include a discount factor $\gamma = 0.99$, GAE parameter $\lambda = 0.95$, and entropy coefficient $0.01$.

To estimate the Markov-chain sensitivities required for the outer-level updates, $40$ trajectories are sampled from the simulator using the current policy. 
These samples are used to compute the sensitivities according to Eq.~\ref{eq:nabla_theta_phi}, Eq.~\ref{eq:mc_pg_phi_sensitivity_expected_value_form}, and Eq.~\ref{eq:mc_pg_theta_sensitivity_expected_value_form}.
Note that a first-order approximation of Eq.~\ref{eq:nabla_theta_phi} is used in the quadrotor example, wherein the Hessian term is neglected for computational reasons. 

For each outer-level iteration, $n$ in-sim \ac{ppo} updates are carried out before applying a simulator-parameter update using the stochastic policy-gradient step in Eq.~\eqref{eq:real_world_spg}. 
Results are reported for $n=20$. The outer-loop learning rate is set to $10^{-3}$.

\section{Related Works}\label{appendix:related_works}
\subsection*{Sim-to-real gap}
The problem of the sim-to-real gap has been commonly addressed by either enriching the simulator fidelity or exposing agents to diverse simulated scenarios. Domain randomization is a common approach that perturbs the simulation parameters during training to improve robustness, such that the resulting in-sim policy generalizes better to the real-world~\citep{tobin2017domain,duan2024learning}. \emph{Domain adaptation} approaches instead aligns feature or parameter distributions between the simulator and real world, commonly via adversarial training on observations~\citep{chen2020adversarial,long2018conditional,pei2018multi,hofer2021sim2real,eysenbach2020off}. In \emph{meta \ac{rl}}, the agent is trained on a distribution of environments with the goal that the agent can then quickly adapt to the real world \citep{finn2017model}.  Other approaches include grounded action methods to adjust simulator dynamics to match observed real trajectories~\citep{hanna2017grounded,desai2020stochastic}, and distributionally robust \ac{rl} formulates the problem as maximizing worst‐case returns under bounded transition shifts~\citep{liu2024distributionally}. Most of these approaches are robustness‐focused rather than performance-oriented; they seek to reduce the sim-to-real gap by optimizing for proxy objectives (e.g., simulator accuracy, variability) rather than the real-world performance directly~\citep{da2025survey}. As a result, they can improve stability under distribution shift but do not offer a guarantee on optimal performance. Few works have explored embedding task performance in domain adaptation. For example, RL-CycleGAN fine-tunes scene parameters by minimizing an \ac{rl}-consistency loss on a learned Q-function~\citep{rao2020rl}, yet this is an indirect surrogate for real-world performance. In contrast, the sensitivity analysis provided in this paper enables the simulation adaptation directly with the real-world performance of the in-sim policy. 
\end{document}

%% file: acronyms.tex
\begin{acronym}
    \acro{sdm}[SDM]{Sequential Decision Making}
   \acro{mdp}[MDP]{Markov Decision Process}
   \acro{rl}[RL]{Reinforcement Learning}
   \acro{sl}[SL]{Supervised Learning}
   \acro{ssl}[SSL]{Self-Supervised Learning}
   \acro{dp}[DP]{Dynamic Programming}
   \acro{adp}[ADP]{Approximate Dynamic Programming}
   \acro{sp}[SP]{Stochastic Programming}
   \acro{lp}[LP]{Linear Programming}
   \acro{mpc}[MPC]{Model Predictive Control}
   \acro{empc}[EMPC]{Economic Model Predictive Control}
   \acro{td}[TD]{Temporal Difference}
   \acro{dpg}[DPG]{Deterministic Policy Gradient}
   \acro{spg}[SPG]{Stochastic Policy Gradient}
   \acro{mbrl}[MBRL]{Model-based Reinforcement Learning}
   \acro{dt}[DT]{Digital Twin}
   \acro{ml}[ML]{Machine Learning}
   \acro{lstm}[LSTM]{Long Short-Term Memory}
   \acro{rnn}[RNN]{Recurrent Neural Network }
   \acro{dnn}[DNN]{Deep Neural Network }
   \acro{dnns}[DNNs]{Deep Neural Networks}
   \acro{gp}[GP]{Gaussian Processes}
   \acro{ai}[AI]{Artificial Intelligence}
   \acro{mle}[MLE]{Maximum Likelihood Estimation}
   \acro{lqr}[LQR]{Linear Quadratic Regulator}
   \acro{gpu}[GPU]{Graphics processing unit}
   \acro{tpu}[TPU]{Tensor processing unit}
   \acro{gans}[GANs]{Generative Adversarial Networks}
   \acro{cnns}[CNNs]{Convolutional Neural Networks}
   \acro{ls}[LS]{Least Squares}
   \acro{mse}[MSE]{Mean Squared Error}
    \acro{map}[MAP]{Maximum A Posteriori}
    \acro{gmm}[GMM]{Gaussian Mixture Model}
    \acro{hmm}[HMM]{Hidden Markov Model}
    \acro{elbo}[ELBO]{Evidence Lower Bound}
    \acro{vae}[VAE]{Variational Autoencoder}
    \acro{gan}[GAN]{Generative Adversarial Network}
    \acro{ode}[ODE]{Ordinary Differential Equations}
    \acro{ude}[UDE]{Universal Differential Equations} 
    \acro{ham}[HAM]{Hybrid Analysis and Modeling} 
    \acro{vi}[VI]{Variational Inference} 
    \acro{em}[EM]{Expectation-Maximization} 
    \acro{lstdq}[LSTDQ]{Least Square Temporal Difference Q-learning} 
    \acro{rldp}[RLDP]{Reinforcement Learning of Decision Processes}
    \acro{or}[OR]{Operations Research}
    \acro{mppi}[MPPI]{Model Predictive Path Integral}
    \acro{cem}[CEM]{Cross Entropy Method}
    \acro{ift}[IFT]{Implicit Function Theorem}
    \acro{ppo}[PPO]{Proximal Policy Optimization}
    \acro{sac}[SAC]{Soft Actor Critic}
    \acro{llm}[LLM]{Large Language Model}
    \acro{nn}[NN]{Neural Network}
    \acro{rp}[RP]{Reparameterization Gradient}
    \acro{lr}[LR]{Likelihood-ratio Gradient}
    
\end{acronym}

%% file: figures/framework.pdf_tex
%% Creator: Inkscape 1.4 (1:1.4+202410161351+e7c3feb100), www.inkscape.org
%% Accompanies image file 'RLfig.pdf' (pdf, eps, ps)
\begingroup%
  \makeatletter%
  \providecommand\color[2][]{%
    \errmessage{(Inkscape) Color is used for the text in Inkscape, but the package 'color.sty' is not loaded}%
    \renewcommand\color[2][]{}%
  }%
  \providecommand\transparent[1]{%
    \errmessage{(Inkscape) Transparency is used (non-zero) for the text in Inkscape, but the package 'transparent.sty' is not loaded}%
    \renewcommand\transparent[1]{}%
  }%
  \providecommand\rotatebox[2]{#2}%
  \newcommand*\fsize{\dimexpr\f@size pt\relax}%
  \newcommand*\lineheight[1]{\fontsize{\fsize}{#1\fsize}\selectfont}%
  \ifx\svgwidth\undefined%
    \setlength{\unitlength}{518.27635277bp}%
    \ifx\svgscale\undefined%
      \relax%
    \else%
      \setlength{\unitlength}{\unitlength * \real{\svgscale}}%
    \fi%
  \else%
    \setlength{\unitlength}{\svgwidth}%
  \fi%
  \global\let\svgwidth\undefined%
  \global\let\svgscale\undefined%
  \makeatother%
  \begin{picture}(1,0.25110531)%
    \lineheight{1}%
    \setlength\tabcolsep{0pt}%
    \put(0,0){\includegraphics[width=\unitlength,page=1]{figures/framework.pdf}}%
    \put(0.16184688,0.20905721){\color[rgb]{0.10196078,0.10196078,0.10196078}\makebox(0,0)[lt]{\lineheight{1.25}\smash{\begin{tabular}[t]{l}$a$\end{tabular}}}}%
    \put(0.31199202,0.19446747){\color[rgb]{0.10196078,0.10196078,0.10196078}\makebox(0,0)[lt]{\lineheight{1.25}\smash{\begin{tabular}[t]{l}Real-world\end{tabular}}}}%
    \put(0.16419903,0.01558035){\color[rgb]{0.10196078,0.10196078,0.10196078}\makebox(0,0)[lt]{\lineheight{1.25}\smash{\begin{tabular}[t]{l}Sim-to-real RL\end{tabular}}}}%
    \put(0.12281138,0.04175839){\color[rgb]{0.10196078,0.10196078,0.10196078}\makebox(0,0)[lt]{\lineheight{1.25}\smash{\begin{tabular}[t]{l}$\mathcal{P}_\theta$\end{tabular}}}}%
    \put(0.02002733,0.04815159){\color[rgb]{0.10196078,0.10196078,0.10196078}\makebox(0,0)[lt]{\lineheight{1.25}\smash{\begin{tabular}[t]{l}$R_\theta, \hat{s}_+$\end{tabular}}}}%
    \put(0.07226644,0.13581332){\color[rgb]{0.10196078,0.10196078,0.10196078}\makebox(0,0)[lt]{\lineheight{1.25}\smash{\begin{tabular}[t]{l}In-sim RL\end{tabular}}}}%
    \put(0.05781322,0.11694563){\color[rgb]{0.10196078,0.10196078,0.10196078}\makebox(0,0)[lt]{\lineheight{1.25}\smash{\begin{tabular}[t]{l}optimizing $\pi_\phi$\end{tabular}}}}%
    \put(0.08634826,0.17360229){\color[rgb]{0.10196078,0.10196078,0.10196078}\makebox(0,0)[lt]{\lineheight{1.25}\smash{\begin{tabular}[t]{l}$\pi_\phi$\end{tabular}}}}%
    \put(0.23796957,0.14598119){\color[rgb]{0.10196078,0.10196078,0.10196078}\makebox(0,0)[lt]{\lineheight{1.25}\smash{\begin{tabular}[t]{l}$\pi_\phi$\end{tabular}}}}%
    \put(0.62146803,0.20766722){\color[rgb]{0.10196078,0.10196078,0.10196078}\makebox(0,0)[lt]{\lineheight{1.25}\smash{\begin{tabular}[t]{l}$a$\end{tabular}}}}%
    \put(0.75207292,0.1948085){\color[rgb]{0.10196078,0.10196078,0.10196078}\makebox(0,0)[lt]{\lineheight{1.25}\smash{\begin{tabular}[t]{l}$a$\end{tabular}}}}%
    \put(0.74109026,0.07979534){\color[rgb]{0.10196078,0.10196078,0.10196078}\makebox(0,0)[lt]{\lineheight{1.25}\smash{\begin{tabular}[t]{l}$r, s_+$\end{tabular}}}}%
    \put(0.88040566,0.19454644){\color[rgb]{0.10196078,0.10196078,0.10196078}\makebox(0,0)[lt]{\lineheight{1.25}\smash{\begin{tabular}[t]{l}Real-world\end{tabular}}}}%
    \put(0.64694981,0.01557787){\color[rgb]{0.10196078,0.10196078,0.10196078}\makebox(0,0)[lt]{\lineheight{1.25}\smash{\begin{tabular}[t]{l}Bi-level sim-to-real RL\end{tabular}}}}%
    \put(0.58243253,0.04036839){\color[rgb]{0.10196078,0.10196078,0.10196078}\makebox(0,0)[lt]{\lineheight{1.25}\smash{\begin{tabular}[t]{l}$\mathcal{P}_\theta$\end{tabular}}}}%
    \put(0.47964848,0.04676159){\color[rgb]{0.10196078,0.10196078,0.10196078}\makebox(0,0)[lt]{\lineheight{1.25}\smash{\begin{tabular}[t]{l}$R_\theta, \hat{s}_+$\end{tabular}}}}%
    \put(0.53188759,0.13442333){\color[rgb]{0.10196078,0.10196078,0.10196078}\makebox(0,0)[lt]{\lineheight{1.25}\smash{\begin{tabular}[t]{l}In-sim RL\end{tabular}}}}%
    \put(0.51743436,0.11555564){\color[rgb]{0.10196078,0.10196078,0.10196078}\makebox(0,0)[lt]{\lineheight{1.25}\smash{\begin{tabular}[t]{l}optimizing $\pi_\phi$\end{tabular}}}}%
    \put(0.54596941,0.17221229){\color[rgb]{0.10196078,0.10196078,0.10196078}\makebox(0,0)[lt]{\lineheight{1.25}\smash{\begin{tabular}[t]{l}$\pi_\phi$\end{tabular}}}}%
    \put(0.69708006,0.14884628){\color[rgb]{0.10196078,0.10196078,0.10196078}\makebox(0,0)[lt]{\lineheight{1.25}\smash{\begin{tabular}[t]{l}Outer-level RL\end{tabular}}}}%
    \put(0.68955024,0.1283126){\color[rgb]{0.10196078,0.10196078,0.10196078}\makebox(0,0)[lt]{\lineheight{1.25}\smash{\begin{tabular}[t]{l}optimizing $R_\theta,\mathcal{P}_\theta$\end{tabular}}}}%
  \end{picture}%
\endgroup%